# Harnessing the Intrinsic Knowledge of Pretrained Language Models for Challenging Text Classification Settings

by

Lingyu Gao

A thesis submitted
in partial fulfillment of the requirements for
the degree of

Doctor of Philosophy in Computer Science

at the

Toyota Technological Institute at Chicago
Chicago, Illinois

August, 2024

Thesis Committee:

Professor Kevin Gimpel (Thesis Advisor)
Professor Karen Livescu
Dr. Debanjan Ghosh

Lingyu Gao
lygao@ttic.edu

# ABSTRACT

Text classification, a classic task in natural language processing (NLP), involves assigning predefined categories to textual data and is crucial for applications ranging from sentiment analysis to spam detection. This thesis advances text classification by harnessing the intrinsic knowledge of Pretrained Language Models (PLMs) to address three challenging scenarios: distractor selection for multiple-choice cloze questions, improving robustness for prompt-based zero-shot text classification, and demonstration selection for retrieval-based in-context learning.

Firstly, we focus on selecting distractors for multiple-choice cloze questions, ensuring that they are misleading yet incorrect. We assess the relationship between human experts' annotations (accept/reject) and various features, including context-free features (e.g., word frequency) and context-sensitive features (e.g., conditional probabilities of fill-in-the-blank words). We utilize pretrained embeddings and follow annotation instructions for context-free feature design, and we find that using contextualized word representations from PLMs as features drastically improves performance over traditional feature-based models, even rivaling human performance (Chapter 3).

Secondly, prompt-based zero-shot approaches are highly sensitive to the choice of prompts, even with task descriptions. We propose to exploit the intrinsic knowledge of the model by providing domain-independent label descriptions. We craft small datasets that describe task labels with related terms, short templates, dictionary definitions, and more. This approach achieves an average improvement of 17-19% in accuracy over traditional zero-shot methods across multiple datasets. It is robust to variations in patterns and verbalizers and proves effective across different text domains, even outperforming few-shot out-of-domain learning in multiple settings (Chapter 4).

Lastly, we consider PLMs' existing knowledge of the task-specific label space of both in-context learning demonstrations and test inputs. We find that using demonstrations that are misclassified by the models, particularly those that lie near the decision boundary of the test examples, leads to better performance. Additionally, considering output label space is more important than semantic similarity, and our methods help reduce model confusion. Extensive experiments on fine-grained classification tasks show that our method improves F1 macro scores by up to 2.6% over traditional retriever-based ap-proaches (Chapter 5).

In conclusion, by leveraging contextualized word representations for distractor selection, and focusing on zero-shot and few-shot tasks that emphasize strategic demonstration selection, this thesis demonstrates the effective use of PLMs to enhance performance and robustness in text classification.

# ACKNOWLEDGMENTS

# ACKNOWLEDGMENTS

First and foremost, I would like to thank my advisor, Kevin Gimpel. He is incredibly knowledgeable and has a deep understanding of many areas beyond natural language processing. Kevin has always been kind, patient, and encouraging, consistently providing great insights. I am fortunate to have been his student, and I genuinely believe my path would have been very different had I not come to TTIC. While I may never reach even half of the rigor, wisdom, and modesty that Kevin embodies, I aspire to follow his example throughout my life.

I would also like to thank my exceptional committee members, Debanjan Ghosh and Karen Livescu. Debanjan, one of my mentors during my internship at Educational Testing Service (ETS), provided invaluable guidance during my first NLP internship. Since then, we have continued to collaborate, and I have greatly benefited from his suggestions. I am also grateful to Karen for her constructive feedback and detailed comments. Additionally, I would like to extend my appreciation to other faculty members and administrative staff at TTIC, particularly Allyson Ettinger, Jiawei Zhou, Greg Shakhnarovich, Madhur Tulsiani, and Adam Bohlander, for their inspiring and helpful conversations.

My heartfelt thanks go to my collaborators and fellow lab members: Xiaomeng Ma, Qihui Xu, Bowen Shi, Freda Shi, Shubham Toshniwal, Arnar Jensson, Xiaoan Ding, Mingda Chen, Lifu Tu, Qingming Tang, Xiao Luo, Ankita Pasad, Davis Yoshida, Shashank Srivastava, Sudarshan Babu, Zewei Chu, Hai Wang, and many more students at TTIC. Special thanks to Xiaomeng Ma for all the delightful academic and personal discussions.

During my PhD, I was fortunate to intern at ETS, TikTok, and Google Research. I am grateful to Swapna Somasundaran, Hillary Molloy, and Mengxuan Zhao from ETS; Mengyin Lu and Shizhu Liu from TikTok; and Aditi Chaudhary, Krishna Srinivasan, Kazuma Hashimoto, Karthik Raman, and Michael Bendersky from Google for their mentorship and support. I also met many wonderful fellow interns, whose presence made my internship experience both enjoyable and rewarding.

Thanks to my friends Lingyun Li, Anna Ren, Rui Xu, Wanfang Guo, Silu Ou, Haotian Zhu, Henry Sui, Songhao Jiang, Yujian Xu, Jiabin Lin, and many others for the virtual study sessions, joyful gaming times, and all the shared jokes and laughter, especially during the challenging times of the pandemic. My thanks also go to Bo Chen, Yanzhi Xin, Yu Wu, Nan Shi, and all other Circle Cat members; it has been a pleasure working with you all. Special thanks to Ye Jiang for supporting me during my darkest moments. It has been more than 15 years since we met, and I look forward to many more years of friendship ahead.

Finally, I want to express my deepest gratitude to my family for their unconditional love, patience, understanding, and unwavering support. They have been my rock, providing strength throughout this journey.

# TABLE OF CONTENTS

# LIST OF FIGURES

# LIST OF TABLES

# CHAPTER 1

# Introduction

Consider a news article that begins with *"Climate scientists tell a conference that greater efforts should be made to pull $CO_2$ from the atmosphere."* [1] Where would it be categorized? Most likely, under Tech News rather than Sports News. Similarly, the title of an email might indicate whether it is a scam, and the correctness of a math problem solution can be judged given the question and answer.

These scenarios highlight the essence of **text classification**—a fundamental task in natural language processing (NLP) where the objective is to assign predefined categories to a piece of text, whether it be a phrase, a sentence, a paragraph, or a series of long documents [2]. Text classification is ubiquitous in daily life, underpinning a broad range of applications from sentiment analysis to toxic text filtering. At the same time, it could be challenging due to the complex dependencies and inherent ambiguity of natural language.

Traditional text classification models often require extensive labeled datasets and manual feature engineering. To classify a series of textual inputs, we first map them to real-valued vectors, a process known as **feature extraction** [2]. Early models heavily relied on manually designed features, and often benefited from traditional pretrained word representations such as Word2Vec [3, 4] and GloVe [5]. These static word embeddings map the discrete words to continuous vector space, and maintain intrinsic pretrained knowledge, such as word analogies (e.g., "king" - "man" + "woman" ≈ "queen"). However, these static word embeddings fail to capture the context-dependent nature of polysemous words, and the small-scale models trained from scratch often struggle with generalization and require large, domain-specific labeled datasets to achieve satisfactory performance, especially on complex tasks.

---

[1]This sentence is taken from AGNews dataset [1].

Recent advancements in deep learning, particularly in transformer architecture [6] and large-scale pretraining, have achieved inspiring success in NLP fields. **Pretrained language models (PLMs)**,[2] such as ELMo [8], GPT-2 [9] and BERT [10], have demonstrated the ability to capture intricate patterns within large corpora and retain vast amounts of knowledge during training [11, 12]. They can be used directly or adapted as needed to enhance text classification tasks, and their encoded intrinsic knowledge facilitates strong performance in **zero-shot** scenarios, where we don't have available training data.

In this thesis, we explore three challenging settings in text classification, focusing particularly on harnessing PLMs and leveraging their intrinsic knowledge for the task. These settings are:

- **Distractor Analysis and Selection for Multiple-Choice Cloze Questions** (Section 1.1). To tackle the challenge of generating *misleading yet incorrect* distractors for cloze questions, we develop models that utilize features designed with internal word representations derived from PLMs.

- **Label-Description Training for Zero-Shot Text Classification** (Section 1.2). To address the difficulty of *generalizing to unseen labels*, we craft small finetuning datasets that describe task labels. This approach significantly improves model robustness and performance by exploiting the models' intrinsic knowledge.

- **Ambiguity-Aware In-Context Learning with Large Language Models** (Section 1.3). To deal with the *sensitivity of PLMs to prompts* in in-context learning, we select effective demonstrations by considering misclassified demonstrations and resolving model

---

[2]Here, we consider language models as models that assign probabilities to sequences of words [7].

ambiguity about test example labels.

Detailed discussions of these settings and our contributions are provided in the subsequent sections.

## 1.1 Feature engineering with PLMs for distractor selection

**Distractor selection** involves determining whether an incorrect answer (distractor) is plausible enough to challenge the test-taker without making the question unanswerable. Consider this cloze question: "The bank will _____ its customers of the new policy." with its correct answer being "notify". We need to decide whether "collaborate" is a good distractor here. Compared to predicting correct answers, designing questions with appropriate distractors is more complex. While we can retrieve questions and correct answers from plain text, selecting optimal distractors—those that are similar to the correct answers but still incorrect [13]—is challenging.

In real-world scenarios, the annotators who select the distractors are often domain experts following specific instructions that could be transformed into features. In practice, a feature-based lightweight model is sometimes preferred. But how do we design a feature-based lightweight model while using the PLMs' knowledge at the same time? How much gain would the PLMs bring?

**Feature engineering** is the practice of constructing or selecting suitable features with domain knowledge to improve model performance [14]. For this cloze question task, the features could be one or more words before or after the blank, part-of-speech (POS) tags of the previous word, frequency of the candidate words, etc. While we could directly input word frequency to the model, the text needs to be mapped to appropriate representations.

In contrast to static word embeddings, **contextualized word representations** derived from PLMs are functions of the entire textual input [11], making them context-sensitive and potentially better for feature design. For instance, the polysemous word "bank" could refer to either a river bank or a financial bank by itself, which could be disambiguated from the context in the example mentioned above.

To achieve a deeper understanding of context, PLMs typically share the common objective of predicting or reconstructing tokens based on contextual information during pretraining tasks. Encoder-only models, such as BERT and its variants, utilize a **"Masked Language Model (MLM)"** pretraining objective. It involves randomly masking some input tokens and predicting them based on the surrounding context, similar to an open cloze-style question. For the cloze question example in this section, we could make predictions with BERT by computing the conditional probabilities of the candidate "collaborate" in the given contexts. The conditional probabilities can be seen as features and combined across multiple models for the final prediction.

Past work often lacks direct supervision for training, making it challenging to develop and evaluate automatic methods for distractor selection. In this thesis, we experiment on two datasets of multiple-choice cloze questions (MCQs) for second-language learners, where the distractor selections are manually annotated by human experts. As it is a binary classification task, it could be turned into a ranking problem and auto-suggests candidates for human experts.

We assess the relationship between annotators' choices and features based on distractors and the correct answers, both with and without the surrounding passage context in the cloze questions. We find that simple features of the distractor and correct answer correlate with the annotations, though using PLMs to measure the fit of the distractor in the context additionally offers substantial benefits. Based on these analyses, we also propose and train models to automatically select distractors and quantitatively measure the importance of model components. Our contributions are:

- We design a range of features, both context-free and context-sensitive, and find that they weakly correlate with human annotations.

- We develop and train models by combining simple features with advanced contextu-

alized word representations from PLMs. Our strongest models are able to approach or even exceed human performance.

- We provide a detailed quantitative analysis of the importance of various model components, offering insights into how different features contribute to performance.

## 1.2 Improving robustness for prompt-based zero-shot classification

The emergence of PLMs has given rise to a pretrain-and-finetune paradigm [15], which achieves impressive performance but typically requires labeled data from downstream tasks. In **zero-shot text classification**, where such datasets are unavailable, it becomes challenging for models to generalize to new, unseen labels during training.

One approach to address this challenge is to provide the model with task descriptions, exploiting the intrinsic knowledge of PLMs to solve zero-shot tasks without supervision [12, 16, 17, 18, 19]. The core idea is transforming text classification into language modeling, i.e., prompt-based classification.

Among the various prompt-based methods, the pattern-verbalizer approach [18] (detailed in Section 2.2.2) is a notable example. It converts the task into a cloze question to match the pretraining task format. In this method, a pattern constructs the prompt from the textual input with a single mask token, and the verbalizer maps each label to a word from the model's vocabulary. For instance, to classify the restaurant review "Overpriced, salty and overrated!", a pattern like "the restaurant is [MASK]" is appended to the review. The model then predicts the most probable verbalizer (e.g., "good" for positive sentiment and "bad" for negative) for the [MASK] position. While this approach is commonly associated with encoder-based models like RoBERTa [20], autoregressive models can generate the next word or phrase based on the prompt, adhering to the same underlying idea of prompt-based classification.

While effective, this approach is highly sensitive to the choice of patterns and verbalizers, and minor changes in the wording of the prompt can lead to significant variations in model performance [21, 22, 23, 24]. This sensitivity has led to the development of *prompt engineering* to find the most appropriate prompt for better performance, however, the best practices vary by task [25]. Additionally, despite advancements in understanding task descriptions, models still face challenges with the representation of labels in text classification. To avoid irrelevant answers, researchers often make predictions by comparing the conditional probabilities of pre-defined strings. However, this approach may suffer from various biases, including **surface form competition** [26], where probability is distributed among various valid strings, including those that differ trivially, such as by capitalization.

To mitigate the sensitivity of the models, we curate small finetuning datasets intended to describe the labels for a task. Unlike typical finetuning data, which has texts annotated with labels, our data describes the labels in language, e.g., using a few related terms, dictionary or encyclopedia entries, and short templates. Our method works for both MLM-style models and autoregressive models, as the data can be used for finetuning as well as in-context learning, where it is included in the textual prompts. This approach is domain-independent, easily adaptable to most use cases, and improves model robustness and performance. Our contributions are:

- Across a range of topic and sentiment datasets, our method is more accurate than zero-shot by 17-19% absolute.

- It is more robust to choices required for zero-shot classification, such as patterns for prompting the model to classify and mappings from labels to tokens in the model's vocabulary.

- Since our data merely describes the labels but does not use input texts, finetuning on it yields a model that performs strongly on multiple text domains for a given label set, even improving over few-shot out-of-domain classification in multiple settings.

## 1.3 Demonstration selection for in-context learning

**In-context learning (ICL)** is a tuning-free approach where the input-output examples (known as **demonstrations**) are concatenated with the textual input [17]. ICL preserves the generality of the LLMs as it doesn't change the model parameters [27]. However, the length of the input prompt is usually limited, and only a few demonstrations could be included. Since PLMs are sensitive to the prompts, selecting good demonstrations becomes a crucial research question.

One effective strategy is leveraging semantic similarity between the ICL demonstrations and test examples with a text retriever [28]. The retriever can either be an off-the-shelf one such as [29, 30, 31, 32], or a retriever trained specifically for that task [28, 33]. Compared to a static set of demonstrations, this dynamic and context-sensitive approach leads to substantial improvements and makes PLMs less sensitive to factors such as demonstration ordering [34].

However, Lyu et al. [35] indicates that there is a **copy effect** where the language model's predictions are significantly influenced by demonstration inputs that closely resemble the test input. This suggests that the retrieval-based approach depends heavily on the retriever. Off-the-shelf retrievers may not be ideal for some tasks, and tuning the retriever involves a finetuning process similar to traditional finetuning, which undermines the tuning-free benefit of ICL. Additionally, this approach can be sub-optimal without considering the PLM's existing knowledge about the task, especially with respect to the output label space.

Motivated by **uncertainty sampling**—a technique in active learning where the model selects the data points it is most uncertain about—we aim to resolve model ambiguity about test example labels in this thesis by conducting zero-shot experiments in advance.

Through extensive experimentation on three text classification tasks, we find that including demonstrations that the LLM previously mis-classified and also fall near the test example's decision boundary, brings the most performance gain. Our contributions are:

- We develop an ICL method that considers model ambiguity regarding both demonstration and test example labels.

- We add constraints incrementally in our experiments on fine-grained topic and sentiment classification tasks, showing that our method outperforms the retrieval-based ICL on two different model sizes.

- We observe that semantically similar demonstrations tend to share the same gold label as the test example, and filtering them with the set of the top two most likely labels offers a more accurate approximation. This insight sheds light on retrieval-based ICL's effectiveness and contributes to the success of our proposed method.

## 1.4 Organization of the Thesis

The thesis is organized as follows:

- Chapter 2: Background of pretrained language models' architectures and text classification approaches with PLMs that are adopted in this thesis, including the pattern-verbalizer approach and in-context learning.

- Chapter 3: Distractor Analysis and Selection for Multiple-Choice Cloze Questions for Second-Language Learners. It presents the challenges of selecting effective distractors and details the method and the performance gain of utilizing contextualized word representations from PLMs for features. This chapter is based on [36].

- Chapter 4: Label-Description Training for Zero-Shot Text Classification. It explains the creation of small finetuning datasets that describe task labels for topic and sentiment classification. This chapter is based on [37].

- Chapter 5: Ambiguity-Aware In-Context Learning with Large Language Models. It shows that using PLM's existing knowledge, such as the model prediction of demonstrations and test examples, is important to improve model performance and resolve model ambiguity. This chapter is based on [38].

- Chapter 6: Summary of the thesis, including a synthesis of the contributions and potential future work.

# CHAPTER 2

# Text Classification with Pretrained Language Models

This chapter provides an overview of the architectures of pretrained language models (PLMs) and existing approaches for text classification tasks using PLMs.

## 2.1 Pretrained Language Models

**Pretrained language models** are language models that have been trained on large-scale corpora using self-supervised learning techniques [15]. While there is a rich history of work in using large-scale language models [39, 40] and pretraining [41, 42], the widespread adoption and use of PLMs as the default tool for NLP began with the introduction of ELMo [8]. Subsequently, it was further popularized by a series of work such as GPT [43] and BERT [44]. The primary training target of these models involves predicting or reconstructing tokens based on contextual information. This approach aligns with two crucial aspects of language use from a psycholinguistics perspective: comprehension (the ability to understand) and production (the ability to generate).

## 2.1.1 ELMo

ELMo (**E**mbeddings from **L**anguage **Mo**dels) was introduced in 2018 as a novel type of deep contextualized word representation rather than a model for finetuning. However, it improved the state of the art on several NLP benchmarks by integrating deep contextual word representations with existing task-specific architectures.

ELMo is constructed using a bidirectional language model (biLM) and a task-specific layer. The biLM is constructed by two LSTM (Long Short-Term Memory) networks [45]: one is in the forward direction and one in the backward direction. Assume we are given a series of textual inputs $\{x_1, \cdots, x_N\}$, these LSTM networks predict the probability of a token $x_t$ given its history and future context, respectively, as shown below:

$$p(x_1, x_2, \cdots, x_N) = \sum_{t=1}^{N} p(x_t | x_1, x_2, \cdots, x_{t-1})$$

$$p(x_1, x_2, \cdots, x_N) = \sum_{t=1}^{N} p(x_t | x_{t+1}, x_{t+2}, \cdots, x_N)$$

In the pretraining process, the token representations[1] and softmax layer parameters of the two LSTMs are tied, and the objective is to maximize the joint log-likelihood of both the forward and backward directions. When used for a downstream task, the contextualized word representation can be obtained through a learned linear combination of all the layer representations of the word. As ELMo adopts two LSTM layers, the first layer is more suitable for part-of-speech (POS) tagging, while the second layer is better for word sense prediction.

This makes ELMo a feature-based approach, as the ELMo representations are typically used as additional input features for other models. However, the authors of ELMo also mention that finetuning the biLM on domain-specific data improves model performance in some cases.

---

[1] ELMo uses a character-level convolutional neural network (character CNN) [46] to encode each word.

## 2.1.2 GPT and BERT

Feature engineering played a crucial role in early NLP tasks, leading researchers to initially use pretrained language models' (PLMs) contextualized embeddings for feature design. However, as PLMs have become increasingly powerful, the necessity for extensive feature engineering has diminished.

**GPT.** GPT (**G**enerative **P**re-trained **T**ransformer) is a unidirectional (left-to-right) decoder-based model, making it particularly well-suited for natural language generation tasks. GPT uses token and absolute positional embeddings to map input text into a vector space,[2] and the embeddings are directly input to the decoder without the encoder structure. It is trained autoregressively to predict the next token given a sequence of textual inputs, but its transformer layers can also serve as contextualized representations.

Unlike ELMo, which applies task-specific layers on top of the pretrained representations, GPT aims to learn a universal representation and requires minimal changes to the model architecture when transferring to new tasks. For the classification task given a series of textual inputs $\{x_1, \cdots, x_N\}$, and the corresponding labels $\{y_1, \cdots, y_c\} \in \mathcal{L}$, where $\mathcal{L}$ is the set of all possible labels, the following loss, which includes a weight $\lambda$, is applied:[3]

$$L = \sum_{(x,y)} \log P(y|x_1, \cdots, x_N) + \lambda \sum_t \log P(x_t|x_1, \cdots, x_{t-1})$$

After proposing GPT, OpenAI scaled up the model parameter size, used more data for pretraining, and introduced GPT-2 [9] and GPT-3 [47] with a few modifications. They emphasized the importance of unsupervised multitask learning in Radford et al. [9] and introduced "in-context learning" in Brown et al. [47].

---

[2]Note BERT also uses learned absolute position embeddings.
[3]However, the sequence needs to be truncated if its length exceeds a pre-defined context window.

**BERT.** BERT (**B**idirectional **E**ncoder **R**epresentations from **T**ransformers), on the other hand, is an encoder-based model. It is bi-directional, with all layers conditioned on both left and right context. Compared to GPT, BERT has an advantage in tasks that require incorporating context from both directions.

BERT is pretrained on two unsupervised tasks: Masked LM and binarized next sentence prediction (NSP).[4] For the Masked LM (MLM) task (a cloze-style training objective that is crucial for training this bidirectional language model), 15% of the tokens are randomly sampled for prediction. To mitigate possible mismatch between pretraining and finetuning data, the sampled token in the input can be:

- Replaced by a special token [MASK] (80%)

- Replaced by a random token (10%)

- Left unchanged (10%)

As BERT doesn't include the transformer decoder, it uses an MLM head[5] to predict the masked token. Regarding special tokens aside from [MASK], BERT inserts a [CLS] token at the beginning of every input example, which can be used for sentence-level classification with a classification head. It also uses a [SEP] token as a separator to distinguish between text segments, such as sentences.

BERT can be used in a feature-based manner without finetuning, as each transformer layer in BERT provides a contextualized representation of each token. A common approach is to combine these layers with a weighted sum.

**Comparison to ELMo.** Unlike ELMo, which follows a feature-based approach, GPT and BERT belong to the finetuning approach. Another key difference lies in their model architectures: GPT and BERT are built on the transformer architecture [6], whereas ELMo adopts LSTM.

---

[4]RoBERTa [20], a follow-up paper of BERT, removed the NSP task as they show that "removing the NSP loss matches or slightly improves downstream task performance." This improvement could be due to RoBERTa's use of more data and a more challenging Masked LM task. There might also be differences in how BERT and RoBERTa handle the ablation studies regarding the NSP task.

[5]The term "head" refers to the additional neural circuitry added on top of the basic transformer architecture to enable a specific task, and we use a language modeling head for language modeling [7].

The original transformer architecture consists of an encoder and a decoder, where the encoder extracts features from the textual input, and the decoder uses these features to produce the output. GPT and BERT utilize different parts of the transformer architecture: GPT is decoder-based, while BERT is encoder-based.

### 2.1.3   Other Recent PLMs

With the field's rapid evolution, many PLMs have been introduced for both general usage and specific domains, such as finance [48] and medicine [49]. Due to their scaled-up parameter sizes, these models are often referred to as large language models (LLMs) rather than PLMs.

Regarding model architectures, a few modern variants of BERT, such as RoBERTa and DeBERTa [50], are commonly used as encoder-based models for tasks such as text classification and natural language inference. Encoder-decoder models, such as T5 [51] and BART [52], are suitable for sequence-to-sequence (seq2seq) tasks, such as machine translation and text summarization. Many recent LLMs are decoder-based, e.g., GPT-3.5 [53], the Llama series [54, 55, 56], and Google's PaLM and PaLM 2 [19, 57]. Reasons for the popularity of decoder-only models at large scale may include their simpler architecture, strong zero-shot generalization after self-supervised training [58], and ease of use for general-purpose generation tasks [59].

## 2.2   Text Classification with PLMs

Text classification involves assigning predefined categories to textual data. Finetuning on in-domain data generally achieves good performance [60]. However, the increasing size of PLMs makes finetuning challenging. Additionally, the lack of sufficient data in specific domains for finetuning has prompted research into data-efficient methods [61, 62], addressing zero-shot or few-shot scenarios. **Zero-shot** refers to situations where the model is tested on new classes or tasks it hasn't seen during training, and **few-shot** refers to scenarios where only a few examples are available for the task (or class) of interest.

### 2.2.1 Finetuning

The finetuning approach involves adjusting the model's parameters or implementing techniques like prompt tuning or parameter-efficient methods (e.g., adapters [63], LoRA [64], and QLoRA [65]) to minimize parameter changes. This approach is also used to calibrate pretrained models to reduce biases [66]. However, finetuning potentially makes the model to become less generalizable. For example, a question-answering model may not achieve high performance on a classification task [67].

One method to address this is **instruction-tuning** [67, 68, 69], where the model is finetuned on multiple tasks and datasets to learn to follow instructions, thereby enhancing cross-task generalization. This method is different from multi-task finetuning, as ablation studies show that natural instructions are crucial [68]. Prepending inputs with natural language instructions yields better zero-shot results compared to using the task and dataset names; moreover, using inputs without any templates leads to the worst performance [68].

Improving the model's ability to follow instructions will also help it adapt to the user's needs to perform specific tasks, especially for those unlikely to appear naturally in the unsupervised pre-training data. In Wei et al. [68], they phrase the natural language inference (NLI) task as a more natural question, which achieves better performance. Another common approach to aligning models with human preferences is **reinforcement learning with human feedback (RLHF)** [70], which is often conducted after instruction tuning. Recently, some work has adopted high-quality synthetic feedback data generated by LLMs [71].

While these finetuning methods have achieved great success, Zhou et al. [72] argue that only a limited amount of high-quality instruction tuning data is necessary to align models with human preferences and end tasks, as LLMs acquire vast knowledge during the pretraining process.

### 2.2.2 Prompting

**Prompting** involves providing a language model with a textual input (prompt) in inference time to perform tasks. The content of a prompt depends on the use case and can include task descriptions/instructions and input-output examples. This approach leverages the pretrained capabilities of the model to handle various tasks without gradient updates. However, the model is sensitive to the input prompts [23, 24, 73], which can be challenging for practitioners to design effectively in true zero-shot settings.

**Pattern-verbalizer Approach.** The pattern-verbalizer approach [18] is a prompt-based method suitable for zero-shot and few-shot scenarios. However, its main advantage lies in data efficiency.[6]

This approach transforms text classification into a language modeling task, utilizing the pretrained capabilities of models like BERT. For example, given a restaurant review, a prompt might be "[CLS] Overpriced, salty and overrated! The restaurant is [MASK]. [SEP]" The model predicts the masked word based on the context, mapping it to a predefined label using a verbalizer. These verbalizers, such as "great" for positive reviews or "awful" for negative reviews, should be semantically related to the corresponding labels.

It is known that this approach is sensitive to the pattern and verbalizer choices. When Schick and Schütze [18] focuses more on combining different prompt patterns, Gao et al. [74] explores automatically generating prompts and selecting the verbalizer. They also consider including demonstrations (input-output pairs) in the prompt when finetuning over a small number of examples. For instance, consider the following: "[CLS] Overpriced,

---

[6]We introduce this approach under the prompting section because it emphasizes the use of patterns and verbalizers, is suitable for zero-shot experiments, and typically requires less data for fine-tuning.

salty and overrated! It was [MASK]. [SEP] A beautiful park. It was great. [SEP] No reason to watch. It was awful. [SEP]"

**In-Context Learning Approach.** **In-context learning** (ICL) is a tuning-free approach where the demonstrations are concatenated with the textual input [17]. It is similar to the pattern-verbalizer approach, but it focuses on leveraging the model's ability to generalize from a few examples without updating model parameters.[7] LLMs can better follow instructions, and finetuning them is often expensive or impossible due to limited access to model parameters. This makes ICL an effective and flexible approach.

Recent studies show that ICL demonstrations (input-output pairs) are primarily used for specifying the domain and format [35, 75, 76], and ICL demonstrations selection and ordering both influenced the effectiveness [29, 73, 77]. However, retrieval-based ICL, which selects a set of demonstrations for each test example using a retriever, has shown advantages in both robustness and performance [27].

The retriever's objectives often focus on either similarity (mainly semantic similarity) [31, 33] or diversity [78, 79]. Selecting demonstrations based on semantic similarity (or term matching) ensures that the examples are relevant and contextually appropriate, and there are also works based on structural similarity, such as [80]. On the other hand, emphasizing diversity in demonstration selection helps in exposing the model to varied examples, potentially enhancing its robustness and adaptability.

## 2.3 Summary

In this chapter, we give an overview of PLMs' architectures used in this thesis, and existing approaches (both finetuning and prompting) for text classification tasks with PLMs. In Chapter 3, we address the challenges of selecting distractors for cloze questions using contextualized word representations derived from PLMs. In Chapter 4, we tackle the issue of model sensitivity to prompts and propose the use of small finetuning datasets. In Chapter 5, we focus on improving the selection of demonstrations in prompting, proposing a solution to model ambiguity by considering model predictions.

---

[7]Based on our definition, the pattern-verbalizer approach with demonstrations can be seen as ICL without finetuning.

# CHAPTER 3

# Distractor Analysis and Selection for Multiple-Choice Cloze Questions for Second-Language Learners

In this chapter, we focus on selecting distractors for multiple-choice cloze questions, i.e., deciding whether a candidate is selected or not with a binary classifier. This task is challenging because the distractors should be attractive enough to mislead test-takers, yet still be incorrect in terms of the knowledge being tested. In our case, it is a mixture of vocabulary knowledge and contextual understanding, and the distractors could be either semantically or syntactically inappropriate, contributing to the difficulty. Moreover, annotated data has inherent limitations because there is no single right choice; instead, many choices are possible. This variability makes traditional supervised learning challenging. However, pretrained language models could naturally pick up on signals from their training corpora that correlate with distractor quality. We can then leverage this pretrained knowledge with a small amount of supervised data.

Given the complexity of the selection rules, we design a range of features, both context-free and context-sensitive, including contextualized word representations from PLMs. Remarkably, our strongest model matches human performance.

This chapter is based on [36].

## 3.1 Introduction

Multiple-choice cloze questions (MCQs) are widely used in examinations and exercises for language learners [81]. The quality of MCQs depends not only on the question and choice of blank, but also on the choice of **distractors**, i.e., incorrect answers. While great improvements are achieved in question answering and reading comprehension, selecting good distractors is still a problem. Different from selecting the best ones in most of the NLP tasks, distractors, which could be phrases or single words, are incorrect answers that distract students from the correct ones.

According to Pho et al. [82], distractors tend to be syntactically and semantically homogeneous with respect to the correct answers. Distractor selection may be done manually through expert curation or automatically using simple methods based on similarity and dissimilarity to the correct answer [83, 84]. Intuitively, optimal distractors should be sufficiently similar to the correct answers in order to challenge students, but not so similar as to make the question unanswerable [85]. However, past work usually lacks direct supervision for training, making it difficult to develop and evaluate automatic methods. To overcome this challenge, Liang et al. [81] sample distractors as negative samples for the candidate pool in the training process, and Chen et al. [86] sample questions and use manual annotation for evaluation.

In this thesis, we experiment on two datasets of MCQs for second-language learners with distractor selections annotated manually by human experts. Both datasets consist of instances with a sentence, a blank, the correct answer that fills the blank, and a set of candidate distractors. Each candidate distractor has a label indicating whether a human annotator selected it as a distractor for the instance. The first dataset, which we call MCDSENT, contains solely the sentence without any additional context, and the sentences are written such that they are understandable as standalone sentences. The second dataset, MCDPARA, contains sentences drawn from an existing passage and therefore also supplies the passage context.

| dataset | context with **correct answer** | distractor | label |
|---------|--------------------------------|------------|-------|
| MCDSENT | How many people are planning to **attend** the party? | contribute | T |
|         | The large automobile **manufacturer** has a factory near here. | beer | F |
|         | The large automobile **manufacturer** has a factory near here. | corporation | F |
|         | The large automobile **manufacturer** has a factory near here. | apartment | T |
| MCDPARA | Stem cells are special cells that can divide to produce many different kinds of cells. When they divide, the new cells may be the same type of **cell** as the original cell.... | plastic | F |
|         | ...These circumstances made it virtually impossible for salmon to mate. Therefore, the number of **salmon** declined dramatically. | thousands | T |

Table 3.1: Example instances from MCDSENT and MCDPARA. Contexts are shown and correct answers are bold and underlined. Part of the paragraph contexts are replaced by ellipses.

To analyze the datasets, we design context-free features of the distractor and the correct answer, including length difference, embedding similarities, frequencies, and frequency rank differences. We also explore context-sensitive features, such as probabilities from large-scale pretrained models like BERT [44]. In looking at the annotations, we found that distractors are unchosen when they are either too easy or too hard (i.e., too good of a fit in the context). Consider the examples in Table 3.1. For the sentence "The large automobile **manufacturer** has a factory near here.", "beer" is too easy and "corporation" is too good of a fit, so both are rejected by annotators. We find that the BERT probabilities capture this tendency; that is, there is a nonlinear relationship between the distractor probability under BERT and the likelihood of annotator selection.

We develop and train models for automatic distractor selection that combine simple features with representations from pretrained models like BERT and ELMo [8]. Our results show that the pretrained models improve performance drastically over the feature-based models, leading to performance rivaling that of humans asked to perform the same task. By analyzing the models, we find that the pretrained models tend to give higher score to grammatically-correct distractors that are similar in terms of morphology and length to the correct answer, while differing sufficiently in semantics so as to avoid unanswerability.

## 3.2 Related Work

Existing approaches to distractor selection use WordNet [87] metrics [86, 88], word embedding similarities [89], thesauruses [90, 91], and phonetic and morphological similarities [92]. Other approaches consider grammatical correctness, and introduce structural similarities in an ontology [93], and syntactic similarities [94]. When using broader context, bigram or $n$-gram co-occurrence [95, 96], context similarity [83], and context sensitive inference [97] have also been applied to distractor selection.

Based on these heuristic features, Liang et al. [81] assemble these features and apply neural networks, training the model to predict the answers within a lot of candidates. Yeung et al. [85] further applies BERT for ranking distractors by masking the target word. As we have two manually annotated datasets that have different lengths of contexts, we adopt both word pair features and the context-specific distractor probabilities to build our feature-based models. Moreover, we build both ELMo-based and BERT-based models, combining them with our features and measuring the impact of these choices on performance.

## 3.3 Datasets

We define an **instance** as a tuple $\langle x, c, d, y \rangle$ where $x$ is the **context**, a sentence or paragraph containing a blank; $c$ is the **correct answer**, the word/phrase that correctly fills the blank; $d$ is the **distractor candidate**, the distractor word/phrase being considered to fill the blank; and $y$ is the **label**, a true/false value indicating whether a human annotator selected the distractor candidate.[1] We use the term **question** to refer to a set of instances with the same values for $x$ and $c$.

---

[1] Each instance contains only a single distractor candidate because this matches our annotation collection scenario. Annotators were shown one distractor candidate at a time. The collection of simultaneous annotations of multiple distractor candidates is left to future work.

### 3.3.1 Data Collection

We build two datasets with different lengths of context. The first, which we call MCDSENT ("Multiple Choice Distractors with SENTence context"), uses only a single sentence of context. The second, MCDPARA ("Multiple Choice Distractors with PARAgraph context"), has longer contexts (roughly one paragraph).

Our target audience is Japanese business people with TOEIC level 300-800, which translates to pre-intermediate to upper-intermediate level. Therefore, words from two frequency-based word lists, the New General Service List (NGSL; [98]) and the TOEIC Service List (TSL; [99]), were used as a base for selecting words to serve as correct answers in instances. A proprietary procedure was used to create the sentences for both MCDSENT and MCDPARA tasks, and the paragraphs in MCDPARA are excerpted from stories written to highlight the target words chosen as correct answers. The sentences are created following the rules below:

- A sentence must have a particular minimum and maximum number of characters.

- The other words in the sentence should be at an equal or easier NGSL frequency level compared with the correct answer.

- The sentence theme should be business-like.

All the MCDSENT and MCDPARA materials were created in-house by native speakers of English, most of whom hold a degree in Teaching English to Speakers of Other Languages (TESOL).

### 3.3.2 Distractor Annotation

We now describe the procedure used to propose distractors for each instance and collect annotations regarding their selection.

A software tool with a user interface was created to allow annotators to accept or reject distractor candidates in MCDSENT and MCDPARA. Distractor candidates are sorted automatically for presentation to annotators in order to favor those most likely to be selected. The distractor candidates are drawn from a proprietary dictionary, and those with the same part-of-speech (POS) as the correct answers (if POS data is available) are preferred. Moreover, the candidates that have greater similarity to the correct answers are preferred, such as being part of the same word learning section in the language learning course and the same NGSL word frequency bucket. There is also preference for candidates that have not yet been selected as distractors for other questions in the same task type and the same course unit.[2]

After the headwords are decided through this procedure, a morphological analyzer is used to generate multiple inflected forms for each headword, which are provided to the annotators for annotation. Both the headwords and inflected forms are available when computing features and for use by our models.

Six annotators were involved in the annotation, all of whom are native speakers of English. Out of the six, four hold a degree in TESOL. Selecting distractors involved two-step human selection. An annotator would approve or reject distractor candidates suggested by the tool, and a different annotator, usually more senior, would review their selections. The annotation guidelines for MCDSENT and MCDPARA follow the same criteria. The annotators are asked to select distractors that are grammatically plausible, semantically implausible, and not obviously wrong based on the context. Annotators also must accept a minimum number of distractors depending on the number of times the correct answer appears in the course. Table 3.1 shows examples from MCDSENT and MCDPARA along with annotations.

---

[2]More specific details about this process are included in the supplementary material (Section A.1).

| # annotators | MCDSENT | | | MCDPARA | | |
|---|---|---|---|---|---|---|
| | agree | disagree | total | agree | disagree | total |
| 1 | - | - | 232256 | - | - | 734063 |
| 2 | 2553 | 122 | 2675 | 9680 | 152 | 9841 |
| 3 | 121 | 2 | 123 | 493 | 3 | 496 |
| 4 | 17 | 0 | 17 | 62 | 0 | 62 |
| 5 | 10 | 0 | 10 | 12 | 0 | 12 |
| 6 | 0 | 0 | 0 | 2 | 0 | 2 |

Table 3.2: Numbers of samples for which annotators agree or disagree.

### 3.3.3 Annotator Agreement

Some instances in the datasets have multiple annotations, allowing us to assess annotator agreement. We use the term "sample" to refer to a set of instances with the same $x$, $c$, and $d$. Table 3.2 shows the number of samples with agreement and disagreement for both datasets.[3] Samples with only one annotation dominate the data. Of the samples with multiple annotations, nearly all show agreement.

### 3.3.4 Distractor Phrases

While most distractors are words, some are phrases, including 16% in MCDSENT and 13% in MCDPARA. In most cases, the phrases are constructed by a determiner or adverb ("more", "most", etc.) and another word, such as "most pleasant", "more recently", and 'More Utility'. However, some candidates show other patterns, such as noun phrases "South Pole", erroneously-inflected forms "come ed" and other phrases (e.g. "Promises Of", "No one").

### 3.3.5 Dataset Preparation

We randomly divided each dataset into train, development, and test sets. We remind the reader that we define a "question" as a set of instances with the same values for the context $x$ and correct answer $c$, and in splitting the data we ensure that for a given question, all of its instances are placed into the same set. The dataset statistics are shown in Table 3.3. False labels are much more frequent than true labels, especially for MCDPARA.

---

[3]We are unable to compute traditional inter-annotator agreement metrics like Cohen's kappa since we lack information about annotator identity for each annotation.

| dataset | type | $y$ | train | dev | test | total |
|---|---|---|---|---|---|---|
| MCDSENT | questions | - | 2,713 | 200 | 200 | 3,113 |
| | instances | T | 30,737 | 1,169 | 1,046 | 32,952 |
| | | F | 191,908 | 6,420 | 6,813 | 205,141 |
| | | total | 222,645 | 7,589 | 7,859 | 238,093 |
| MCDPARA | questions | - | 14,999 | 1,000 | 1,000 | 16,999 |
| | instances | T | 49,575 | 597 | 593 | 50,765 |
| | | F | 688,804 | 7,620 | 8,364 | 704,788 |
| | | total | 738,379 | 8,217 | 8,957 | 755,553 |

Table 3.3: Dataset sizes in numbers of questions (a "question" is a set of instances with the same $x$ and $c$) and instances, broken down by label ($y$) and data split.

## 3.4 Features and Analysis

We now analyse the data by designing features and studying their relationships with the annotations.

### 3.4.1 Features

We now describe our features. The dataset contains both the headwords and inflected forms of both the correct answer $c$ and each distractor candidate $d$. In defining the features below based on $c$ and $d$ for an instance, we consider separate features for two types of word pairs:

- headword pair: correct answer headwords and candidate headwords

- inflected form pair: correct answer and candidate

| feature | MCDSENT | | MCDPARA | |
| --- | --- | --- | --- | --- |
|  | head | infl | head | infl |
| length difference | -0.116 | -0.171 | -0.145 | -0.173 |
| embedding similarity | -0.018 | 0.026 | -0.014 | 0.016 |
| candidate frequency | -0.057 | 0.113 | -0.062 | 0.028 |
| freq. rank difference | -0.048 | -0.161 | -0.033 | -0.091 |

Table 3.4: Spearman correlations with T/F choices, where "head" denotes headword pairs, and "infl" denotes inflected form pairs.

For features that require embedding words, we use the 300-dimensional GloVe word embeddings [100] pretrained on the 42 billion token Common Crawl corpus. The GloVe embeddings are provided in decreasing order by frequency, and some features below use the line numbers of words in the GloVe embeddings, which correspond to frequency ranks. For words that are not in the GloVe vocabulary, their frequency ranks are $|N|+1$, where $N$ is the size of the GloVe vocabulary. We use the four features listed below:

- **length difference**: absolute value of length difference (in characters, including whitespace) between $c$ and $d$.

- **embedding similarity**: cosine similarity of the embeddings of $c$ and $d$. For phrases, we average the embeddings of the words in the phrase.

- **distractor frequency**: negative log frequency rank of $d$. For phrases, we take the max rank of the words (i.e., the rarest word is chosen).

- **freq. rank difference**: feature capturing frequency difference between $c$ and $d$, i.e., $\log(1 + |r_c - r_d|)$ where $r_w$ is the frequency rank of $w$.

## 3.4.2 Correlations of Features and Annotations

The Spearman correlations between feature values and labels are presented in Table 3.4. We compute Spearman correlations between features and the T/F annotations, mapping the T/F labels to 1/0 for computing correlations. The overall correlations are mostly close to zero, so we explore how the relationships vary for different ranges of feature values below. Nonetheless, we can make certain observations about the correlations:

- Length difference has a weak negative correlation with annotations, which implies that the probability of a candidate being selected decreases when the absolute value of word length difference between the candidate and correct answer increases. The same conclusion can be drawn with headword pairs although the correlation is weaker.

- Embedding similarity has a very weak correlation (even perhaps none) with the annotations. However, the correlation for headwords is slightly negative while that for inflected forms is slightly positive, suggesting that annotators tend to select distractors with different lemmas than the correct answer, but similar inflected forms (e.g., for the instance "I ⎯⎯ many cakes to find a good one." where the correct answer is "tasted", "taste" and "tastes" are both rejected, while "borrowed", "inspired" and "hired" are selected).

- Candidate frequency also has a very weak correlation with annotations (negative for headwords and positive for inflected forms). Since the feature is the negative log frequency rank, a distractor with a rare headword but more common inflected form is more likely to be selected, at least for MCDSENT.

- Frequency rank difference has a weak negative correlation with annotations, and this trend is more significant with the inflected form pair. This implies that annotators tend to select distractors in the same frequency range as the correct answers.

The correlations are not very large in absolute terms, however we found that there were stronger relationships for particular ranges of these feature values and we explore this in the next section.

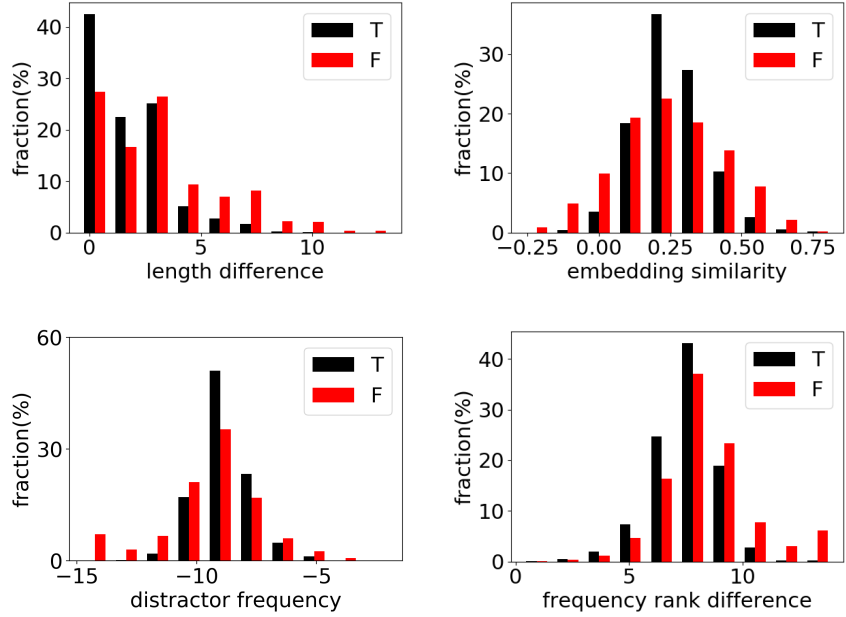

Figure 3.1: Label-specific feature histograms for MCDSENT (inflected form pairs).

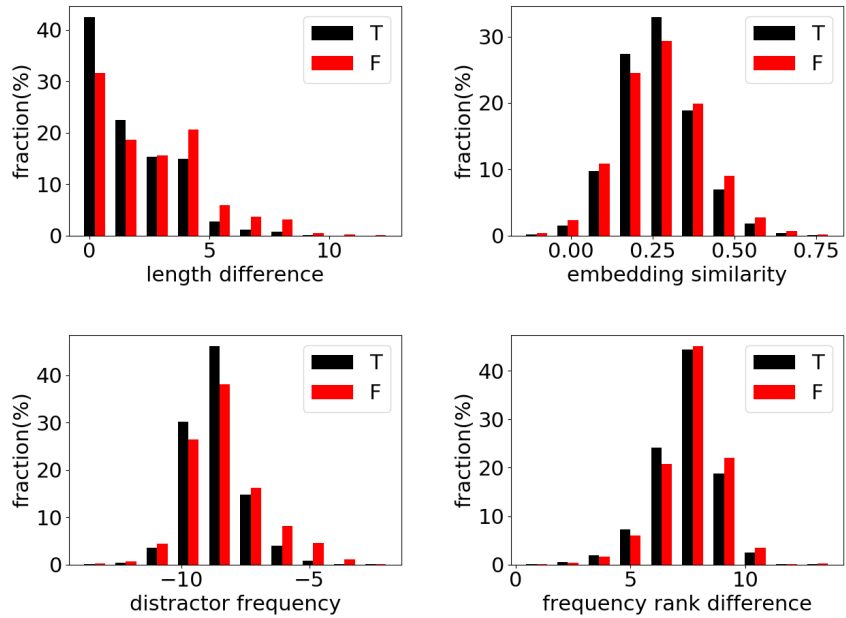

Figure 3.2: Label-specific feature histograms for MCDSENT (headword pairs).

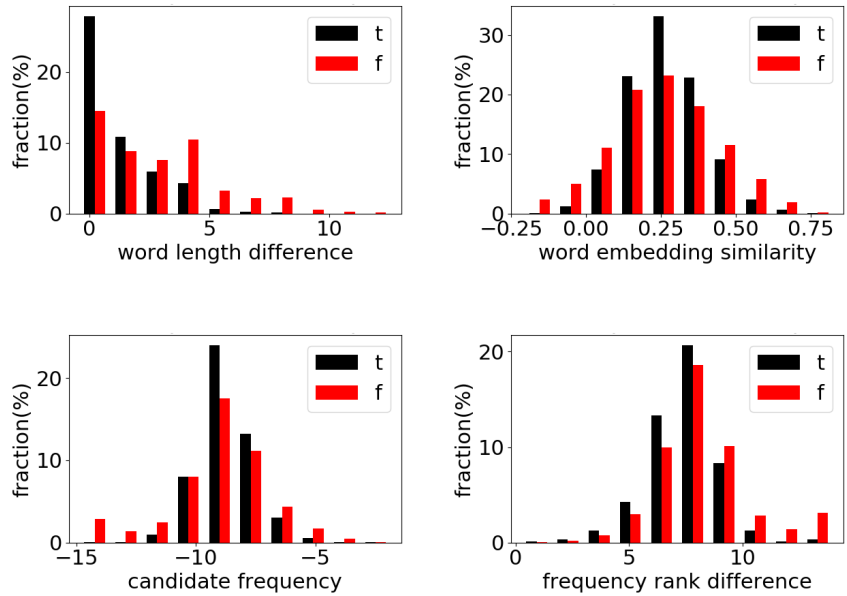

Figure 3.3: Label-specific feature histograms for MCDPARA (inflected form pairs).

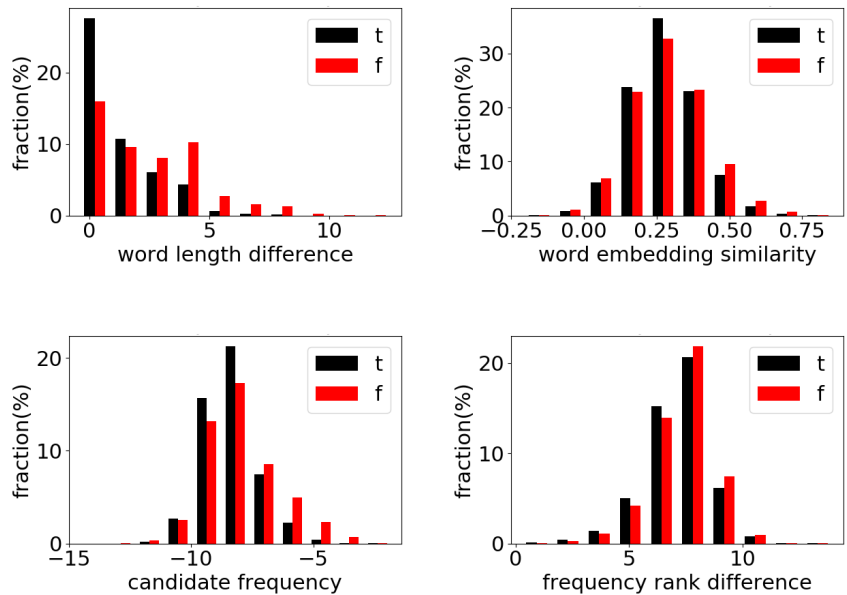

Figure 3.4: Label-specific feature histograms for MCDPARA (headword pairs).

### 3.4.3 Label-Specific Feature Histograms

Figure 3.1 - 3.4 shows histograms of feature values for each label on inflected form pairs and headword pairs for MCDSENT and MCDPARA. Since the data is unbalanced, the histograms are "label-normalized", i.e., normalized so that the sum of heights for each label is 1. So, we can view each bar as the fraction of that label's instances with feature values in the given range. Given that the figures exhibit common trends, we will discuss them below, using Figure 3.1 and 3.2 as an example. We make several observations:

- The annotators favor candidates that have approximately the same length as the correct answers (Fig. 3.1, plot 1), as the true bars are much higher in the first bin (length difference 0 or 1), which accords with the correlations in the previous section.

- Selected distractors have moderate embedding similarity to the correct answers (Fig. 3.1, plot 2). If cosine similarity is very high or very low, then those distractors are much less likely to be selected. Such distractors are presumably too difficult or too easy, respectively. These trends are much clearer for the inflected forms.

- Selected distractors are moderately frequent (Fig. 3.1, plot 3). Very frequent and very infrequent distractors are less likely to be selected. More common words are to the right of the plot. Compared to candidate headwords, there are more rare words among candidates (the heights of the bars rise to the far left under the inflected form setting), which are mainly annotated as false. These tend to be erroneously-inflected and correctly-inflected-but-extremely-rare forms, which annotators do not select.

- Distractors with small frequency rank differences (those on the left of plot 4) are more likely to be chosen (Fig. 3.1, plot 4). Large frequency differences tend to be found with very rare distractors, some of which may be erroneously-inflected forms.

### 3.4.4 Probabilities of Distractors in Context

We use BERT [44] to compute probabilities of distractors and correct answers in the given contexts in MCDSENT. We insert a mask symbol in the blank position and compute the probability of the distractor or correct answer at that position.[4] Figure 3.5 shows histograms for correct answers and distractors (normalized by label). The correct answers have very high probabilities. The distractor probabilities are more variable and the shapes of the histograms are roughly similar for the true and false labels. Interestingly, however, when the probability is very high or very low, the distractors tend to not be selected. The selected distractors tend to be located at the middle of the probability range. This pattern shows that BERT's distributions capture (at least partially) the nonlinear relationship between goodness of fit and suitability as distractors.

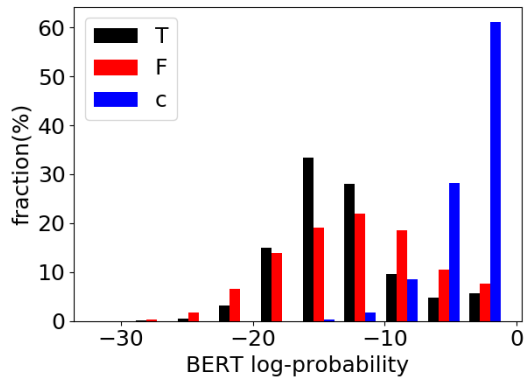

Figure 3.5: Histograms of BERT log-probabilities of selected distractors ("T"), unselected distractors ("F"), and correct answers ("$c$") in MCDSENT.

## 3.5 Models

Since the number of distractors selected for each instance is uncertain, our datasets could be naturally treated as a binary classification task for each distractor candidate. We now present models for the task of automatically predicting whether a distractor will be selected by an annotator. We approach the task as defining a predictor that produces a scalar score for a given distractor candidate. This score can be used for ranking distractors for a given question, and can also be turned into a binary classification using a threshold. We define

---

[4]For distractors with multiple tokens, we mask each position in turn and use the average of the probabilities.

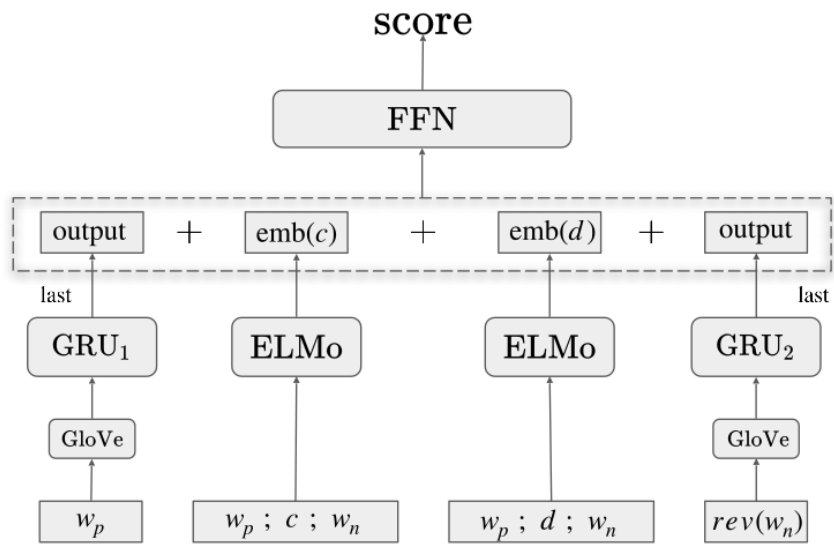

Figure 3.6: Illustration of the ELMo-based model $M_{ELMo}$, where semicolon refers to vector concatenation.

three types of models, described in the subsections below.

### 3.5.1 Feature-Based Models

Using the features described in Section 3.4, we build a simple feed-forward neural network classifier that outputs a scalar score for classification. Only inflected forms of words are used for features without contexts, and all features are concatenated and used as the input of the classifier. For features that use BERT, we compute the log-probability of the distractor and the log of its rank in the distribution. For distractors that consist of multiple subword units, we mask each individually to compute the above features for each subword unit, then use the concatenation of mean, min, and max pooling of the features over the subword units. We refer to this model as $M_{feat}$.

### 3.5.2 ELMo-Based Models

We now describe models that are based on ELMo [8] which we denote $M_{ELMo}$. Since MCDPARA instances contain paragraph context, which usually includes more than one sentence, we denote the model that uses the full context by $M_{ELMo}(\ell)$. By contrast, $M_{ELMo}$ uses only a single sentence context for both MCDSENT and MCDPARA. We denote the correct answer by $c$, distractor candidate by $d$, the word sequence before the blank by $w_p$, and the word sequence after the blank by $w_n$, using the notation $rev(w_n)$ to indicate the reverse of the sequence $w_n$.

We use GloVe [100] to obtain pretrained word embeddings for context words, then use two separate RNNs with gated recurrent units (GRUs; [101]) to output hidden vectors to represent $w_p$ and $w_n$. We reverse $w_n$ before passing it to its GRU, and we use the last hidden states of the GRUs as part of the classifier input. We also use ELMo to obtain contextualized word embeddings for correct answers and distractors in the given context, and concatenate them to the input. An illustration of this model is presented in Figure 3.6.

A feed-forward network (FFN) with 1 ReLU hidden layer is set on top of these features to get the score for classification:

$$FFN(z) = \max(0, zW_1 + b_1)W_2 + b_2$$

where $z$ is a row vector representing the inputs shown in Figure 3.6. We train the model as a binary classifier by using a logistic sigmoid function on the output of $FFN(z)$ to compute the probability of the true label. Based on the model in Figure 3.6, we further experiment with the following variations of this model:

- Concatenate the features from Section 3.4 with $z$.

- Concatenate the correct answer to the input of the GRUs on both sides (denoted *gru+c*).

- Concatenate the GloVe embeddings of the correct answers and distractors with $z$. We combine this with *gru+c*, denoting the combination *all*.

### 3.5.3 BERT-Based Models

Our final model type uses a structure similar to $M_{ELMo}$ but using BERT in place of ELMo when producing contextualized embeddings, which we denote by $M_{BERT}$ and $M_{BERT}(\ell)$ given different types of context. We also consider the variation of concatenating the features to the input to the classifier, i.e., the first variation described in Section 3.5.2. We omit the *gru+c* and *all* variations here because the BERT-based models are more computationally expensive than those that use ELMo.

## 3.6 Experiments

We now report the results of experiments with training models to select distractor candidates.

### 3.6.1 Evaluation Metrics

We use precision, recall, and F1 score as evaluation metrics. These require choosing a threshold for the score produced by our predictors. We also report the area under the precision-recall curve (AUPR), which is a single-number summary that does not require choosing a threshold.

### 3.6.2 Baselines

As the datasets are unbalanced (most distractor candidates are not selected), we report the results of baselines that always return "True" in the "baseline" rows of Tables 3.6 and 3.7. MCDSENT has a higher percentage of true labels than MCDPARA.

| dataset | precision A | precision B | recall A | recall B | F1 A | F1 B |
|---|---|---|---|---|---|---|
| MCDSENT | 62.9 | 48.5 | 59.5 | 43.2 | 61.1 | 45.7 |
| MCDPARA | 32.1 | 25.0 | 36.0 | 24.0 | 34.0 | 24.5 |

Table 3.5: Results of human performance on distractor selection for two human judges labeled A and B.

### 3.6.3 Estimates of Human Performance

We estimated human performance on the distractor selection task by obtaining annotations from NLP researchers who were not involved in the original data collection effort. We performed three rounds among two annotators, training them with some number of questions per round, showing the annotators the results after each round to let them calibrate their assessments, and then testing them using a final set of 30 questions, each of which has at most 10 distractors.

Human performance improved across rounds of training, leading to F1 scores in the range of 45-61% for MCDSENT and 25-34% for MCDPARA (Table 3.5). Some instances were very easy to reject, typically those that were erroneous word forms resulting from incorrect morphological inflection from the word form generator or those that were extremely similar in meaning to the correct answer. But distractors that were at neither extreme were very difficult to predict, as there is a certain amount of variability in the annotation of such cases. Nonetheless, we believe that the data has sufficient signal to train models to provide a score indicating suitability of candidates to serve as distractors.

### 3.6.4 Modeling and Training Settings

All models have one hidden layer for the feed-forward classifier. The $M_{feat}$ classifier has 50 hidden units, and we train it for at most 30 epochs using Adam [102] with learning rate 1e−3. We stop training if AUPR keeps decreasing for 5 epochs.[5] Although our primary metric of interest is AUPR, we also report optimal-threshold F1 scores on dev and test, tuning the threshold on the given set (so, on the test sets, the F1 scores we report are oracle F1 scores). The threshold is tuned within the range of 0.1 to 0.9 by step size 0.1.

For $M_{ELMo}$ and $M_{ELMo}(\ell)$, we use ELMo (Original[6]) for the model, and BERT-large-cased to compute the BERT features from Section 3.4 (only applies to rows with "features = yes" in the tables). We increase the number of classifier hidden units to 1000 and run 20 epochs at most, also using Adam with learning rate 1e−3. We stop training if AUPR does not improve for 3 epochs.

For $M_{BERT}$ and $M_{BERT}(\ell)$, we applied the same training settings as $M_{ELMo}$ and $M_{ELMo}(\ell)$. We compare the BERT-base-cased and BERT-large-cased variants of BERT. When doing so, the BERT features from Section 3.4 use the same BERT variant as that used for contextualized word embeddings.

For all models based on pretrained models, we keep the parameters of the pretrained models fixed. However, we do a weighted summation of the 3 layers of ELMo, and all layers of BERT except for the first layer, where the weights are trained during the training process.

### 3.6.5 Results

We present our main results for MCDSENT in Table 3.6 and for MCDPARA in Table 3.7.

**Feature-based models.** The feature-based model, shown as $M_{feat}$ in the upper parts of the tables, is much better than the trivial baseline. Including the BERT features in $M_{feat}$ improves performance greatly (10 points in AUPR for MCDSENT), showing the value of using the context effectively with a powerful pretrained model. There is not a large

---
[5]We also tune by F1 score as another set of settings with similar trends, which are included in the supplementary material.
[6]https://allennlp.org/elmo

| model variant | | development set | | | | test set | | | | BERT | features | best epoch | threshold |
|---|---|---|---|---|---|---|---|---|---|---|---|---|---|
| | | precision | recall | F1 | AUPR | precision | recall | F1 | AUPR | | | | |
| baseline | | 15.4 | 100 | 26.7 | - | 13.3 | 100 | 23.5 | - | - | - | - | - |
| $M_{feat}$ | | 33.6 | 62.9 | 43.8 | 36.5 | 23.7 | 55.4 | 33.2 | 24.6 | none | yes | 28 | 0.2 |
| | | 44.5 | 57.1 | **50.0** | 46.1 | 28.2 | 70.9 | 40.3 | 32.4 | base | yes | 25 | 0.2 (0.3) |
| | | 36.4 | 77.8 | 49.6 | **47.0** | 30.0 | 71.3 | **42.2** | **34.5** | large | yes | 22 | 0.2 |
| $M_{ELMo}$ | none | 43.2 | 87.5 | 57.8 | 59.0 | 41.4 | 88.0 | 56.3 | 54.6 | - | no | 2 | 0.3 |
| | gru+c | 44.8 | 84.4 | 58.5 | 57.4 | 47.6 | 68.4 | 56.1 | 54.1 | - | no | 2 | 0.3 (0.4) |
| | all | 47.2 | 88.9 | 61.7 | 61.2 | 48.3 | 75.0 | 58.7 | 55.8 | - | no | 2 | 0.3 (0.4) |
| | none | 51.7 | 77.8 | 62.1 | 64.6 | 50.4 | 76.5 | 60.8 | 57.2 | large | yes | 3 | 0.3 |
| | gru+c | 55.7 | 73.3 | 63.3 | 65.3 | 49.1 | 82.3 | 61.5 | **63.1** | large | yes | 5 | 0.4 (0.3) |
| | all | 56.2 | 74.4 | **64.0** | **66.5** | 49.8 | 80.8 | **61.6** | 58.8 | large | yes | 5 | 0.4 (0.3) |
| $M_{BERT}$ | | 47.9 | 78.1 | 59.4 | 60.8 | 44.8 | 81.0 | 57.7 | 55.7 | base | no | 1 | 0.3 |
| | | 49.6 | 79.3 | 61.0 | 64.1 | 45.3 | 80.2 | 57.9 | 53.4 | large | no | 1 | 0.3 |
| | | 50.6 | 83.9 | **63.2** | 65.3 | 44.8 | 78.5 | 57.0 | 53.8 | base | yes | 12 | 0.1 |
| | | 53.8 | 73.1 | 62.0 | **66.5** | 49.7 | 73.9 | **59.4** | **56.3** | large | yes | 2 | 0.4 |

Table 3.6: Results for MCDSENT. Boldface indicates the best F1/AUPR on dev/test for each model type. We include the threshold tuned on the test set in parentheses when it differs from the threshold tuned on dev.

difference between using BERT-base and BERT-large when computing these features.

**ELMo-based models.** Even without features, $M_{ELMo}$ outperforms $M_{feat}$ by a wide margin. Adding features to $M_{ELMo}$ further improves F1 by 2-5% for MCDSENT and 5-6% for MCDPARA. The F1 score for $M_{ELMo}$ on MCDSENT is close to human performance, and on MCDPARA the F1 score outperforms humans (see Table 3.5). For MCDSENT, we also experiment with using the correct answer as input to the context GRUs (*gru+c*), and additionally concatenating the GloVe embeddings of the correct answers and distractors to the input of the classifier (*all*). Both changes improve F1 on dev, but on test the results are more mixed.

**BERT-based models.** For $M_{BERT}$, using BERT-base is sufficient to obtain strong results on this task and is also cheaper computationally than BERT-large. Although $M_{BERT}$ with BERT-base has higher AUPR on dev, its test performance is close to $M_{ELMo}$. Adding features improves performance for MCDPARA (3-5% F1), but less than the improvement found for $M_{ELMo}$. While $M_{feat}$ is aided greatly when including BERT features, the features have limited impact on $M_{BERT}$, presumably because it already incorporates BERT in its model.

| model | development set | | | | test set | | | | BERT features | best epoch | threshold |
|---|---|---|---|---|---|---|---|---|---|---|---|
| | precision | recall | F1 | AUPR | precision | recall | F1 | AUPR | | | |
| baseline | 7.3 | 100 | 13.5 | - | 6.6 | 100 | 12.4 | - | - | - | - | - |
| $M_{feat}$ | 15.3 | 63.1 | 24.6 | 17.3 | 14.5 | 63.6 | 23.6 | 15.5 | - | yes | 23 | 0.1 |
| | 18.2 | 69.2 | 28.9 | 21.6 | 16.3 | 65.6 | 26.1 | **19.1** | base | yes | 27 | 0.1 |
| | 19.8 | 64.0 | **30.2** | **22.3** | 16.9 | 64.2 | **26.8** | 18.8 | large | yes | 22 | 0.1 |
| $M_{ELMo}$ | 35.4 | 47.7 | 40.7 | 38.4 | 26.1 | 75.6 | 38.8 | 30.4 | - | no | 5 | 0.3 (0.2) |
| | 37.9 | 61.3 | **46.9** | **46.8** | 34.6 | 63.9 | 44.9 | 37.6 | large | yes | 7 | 0.3 |
| $M_{ELMo}(\ell)$ | 30.5 | 61.1 | 40.7 | 36.6 | 29.1 | 61.6 | 39.5 | 33.2 | - | no | 5 | 0.3 |
| | 37.1 | 62.7 | 46.6 | 43.7 | 34.4 | 65.1 | **45.0** | **40.1** | large | yes | 6 | 0.3 |
| $M_{BERT}$ | 35.4 | 61.6 | 45.0 | 40.9 | 29.2 | 58.7 | 39.0 | 30.1 | base | no | 2 | 0.2 |
| | 33.0 | 63.7 | 43.5 | 40.9 | 29.1 | 65.1 | 40.2 | 32.4 | large | no | 2 | 0.2 |
| | 44.3 | 55.4 | **49.3** | **47.3** | 31.5 | 73.2 | **44.0** | 36.7 | base | yes | 2 | 0.3 (0.2) |
| | 35.6 | 66.0 | 46.2 | 45.0 | 35.5 | 54.5 | 43.0 | 36.6 | large | yes | 2 | 0.2 (0.3) |
| $M_{BERT}(\ell)$ | 33.1 | 65.3 | 43.9 | 39.7 | 28.8 | 66.4 | 40.2 | 29.8 | base | no | 2 | 0.2 |
| | 37.4 | 67.3 | 48.1 | 46.0 | 31.3 | 69.1 | 43.1 | **37.0** | base | yes | 2 | 0.2 |

Table 3.7: Results for MCDPARA (tuned based on AUPR).

**Long-context models.** We now discuss results for the models that use the full context in MCDPARA, i.e., $M_{ELMo}(\ell)$ and $M_{BERT}(\ell)$. On dev, $M_{ELMo}$ and $M_{BERT}$ outperform $M_{ELMo}(\ell)$ and $M_{BERT}(\ell)$ respectively, which suggests that the extra context for MCDPARA is not helpful. However, the test AUPR results are better when using the longer context, suggesting that the extra context may be helpful for generalization. Nonetheless, the overall differences are small, suggesting that either the longer context is not important for this task or that our way of encoding the context is not helpful. The judges in our manual study (Sec. 3.6.3) rarely found the longer context helpful for the task, pointing toward the former possibility.

### 3.6.6 Statistical Significance Tests

For better comparison of these models' performances, a paired bootstrap resampling method is applied [103]. We repeatedly sample with replacement 1000 times from the original test set with sample size equal to the corresponding test set size, and compare the F1 scores of two models. We use the thresholds tuned by the development set for F1 score computations, and assume significance at a $p$ value of 0.05.

- For $M_{ELMo}$, $M_{ELMo}(\ell)$, $M_{BERT}$ and $M_{BERT}(\ell)$, the models with features are significantly better than their feature-less counterparts ($p < 0.01$).[7]

- When both models use features, $M_{ELMo}(\ell)$ is almost the same as $M_{ELMo}$ ($p = 0.477$). However, when both do not use features, $M_{ELMo}(\ell)$ is significantly better ($p < 0.01$).

- When using BERT-base-cased, $M_{BERT}(\ell)$ is better than $M_{BERT}$, but not significantly so ($p = 0.4$ with features and $0.173$ without features).

- On MCDPARA, switching from BERT-base to BERT-large does not lead to a significant difference for $M_{BERT}$ without features (BERT-large is better with $p = 0.194$) or $M_{BERT}$ with features (BERT-base is better with $p = 0.504$). For MCDSENT, $M_{BERT}$ with BERT-large is better both with and without features ($p < 0.2$).

- On MCDPARA, $M_{BERT}(\ell)$ outperforms $M_{ELMo}(\ell)$ without features but not significantly. With features, $M_{ELMo}(\ell)$ is better with $p = 0.052$.

- On MCDSENT, $M_{BERT}$ without features (BERT-large-cased) is better than $M_{ELMo}$ without features, but not significantly so ($p = 0.386$). However, if we add features or use $M_{BERT}$ with BERT-base-cased, $M_{ELMo}$ is significantly better ($p < 0.01$).

- On MCDPARA, $M_{ELMo}$ is nearly significantly better than $M_{BERT}$ when both use features ($p = 0.062$). However, dropping the features for both models makes $M_{BERT}$ significantly outperform $M_{ELMo}$ ($p = 0.044$).

---
[7]We only use BERT-base-cased for $M_{BERT}(\ell)$ due to computational considerations.

|  | annotations | feature | ELMo | ELMo with feature | BERT | BERT with feature |
|---|---|---|---|---|---|---|
| spell | F | 7 | 2 | 1 | 8 | 3 |
| subscribe | T | 5 | 1 | 2 | 7 | 10 |
| overcome | T | 9 | 6 | 3 | 2 | 1 |
| consult | F | 3 | 3 | 4 | 1 | 6 |
| quit | F | 2 | 8 | 5 | 10 | 4 |
| collaborate | T | 10 | 5 | 6 | 4 | 9 |
| violate | T | 1 | 11 | 7 | 3 | 5 |
| collaborating | F | 33 | 30 | 39 | 45 | 48 |
| chatted | F | 39 | 46 | 40 | 38 | 38 |
| customizing | F | 34 | 42 | 41 | 16 | 27 |
| subscribed | F | 14 | 35 | 42 | 26 | 35 |
| chatting | F | 30 | 47 | 43 | 47 | 43 |
| subscribing | F | 21 | 40 | 44 | 24 | 28 |
| overcoming | F | 17 | 39 | 45 | 21 | 37 |

Figure 3.7: Ranks of distractors for question "The bank will **notify** its customers of the new policy." The colors represent the normalized scores of the models and the numbers in the cells are the ranks of the candidates.

### 3.6.7 Examples

Figure 3.7 shows an example question from MCDSENT, i.e., "The bank will **notify** its customers of the new policy", and two subsets of its distractors. The first subset consists of the top seven distractors using scores from $M_{ELMo}$ with features, and the second contains distractors further down in the ranked list. For each model, we normalize its distractor scores with min-max normalization.[8]

Overall, model rankings are similar across models, with all distractors in the first set ranked higher than those in the second set. The high-ranking but unselected distractors ("spell", "consult", and "quit") are likely to be reasonable distractors for second-language learners, even though they were not selected by annotators.

We could observe the clustering of distractor ranks with similar morphological inflected form in some cases, which may indicate that the model makes use of the grammatical knowledge of pretrained models.

---
[8]Given original data $x$, we use $(x - \min(x))/(\max(x) - \min(x))$ to normalize it.
41

## 3.7 Conclusion

We described two datasets with annotations of distractor selection for multiple-choice cloze questions for second-language learners. We designed features and developed models based on pretrained language models. Our results show that the task is challenging for humans and that the strongest models are able to approach or exceed human performance. The rankings of distractors provided by our models appear reasonable and can reduce a great deal of human burden in distractor selection. Future work will use our models to collect additional training data which can then be refined in a second pass by limited human annotation. Other future work can explore the utility of features derived from pretrained question answering models in scoring distractors.

# CHAPTER 4

# The Benefits of Label-Description Training for Zero-Shot Text Classification

In this chapter, we examine the challenge of zero-shot text classification. This task requires the model to generalize from its existing knowledge without any labeled data for the new classes. Recent approaches transform text classification into a language modeling task [18]. However, this method is highly sensitive to the prompt design and the specific words or phrases used to represent the labels.

Since the model must have a deep understanding of both the text and the class labels to achieve good performance, we propose to finetune the pretrained language model with a small curated dataset of label descriptions, which improves both model performance and robustness.

This chapter is based on [37].

## 4.1 Introduction

Pretrained language models (PLMs) [10, 17, 20, 104, 105] have produced strong results in zero-shot text classification for a range of topic and sentiment tasks, often using a pattern-verbalizer approach [18]. With this approach, to classify the restaurant review "Overpriced, salty and overrated!", a pattern like "the restaurant is [MASK]" is appended to the review and verbalizers are chosen for each label (e.g., "good" for positive sentiment and "bad" for

| Label | Input |
| --- | --- |
| Business | business<br>finance<br>Business is the activity of making one's living or making money by producing or buying and selling products... |
| Sports | sports<br>racing<br>An athletic activity requiring skill or physical prowess and often of a competitive nature, as racing, baseball, tennis, golf,... |

(a) Topic classification

| Label | Input |
| --- | --- |
| Very Negative | awful<br>It was *terrible*.<br>A *horrendous* experience. |
| Very Positive | great<br>Just *fantastic*.<br>Overall, it was *outstanding*. |

(b) Sentiment classification

Table 4.1: A few examples of LABELDESC training data for topic and sentiment classification.

negative). The text is classified by the pretrained masked language modeling (MLM) head to choose the most probable verbalizer for the [MASK] position.[1] Although effective, the approach is sensitive to the choice of specific pattern/verbalizer pairs, with subtle changes in the pattern, the verbalizer, or both, often having a large impact on performance [23, 24].

To alleviate these issues, we propose a simple alternative approach of training on small curated datasets intended to describe the labels for a task. Unlike typical training datasets, which consist of input texts annotated by hand with labels, our data contains only the *descriptions* of the labels. We refer to this data as LABELDESC data and show a few examples for topic and sentiment classification in Table 4.1. For topic classification, we include a few terms related to the label (e.g., "finance" for "Business", "racing" for "Sports"), a definition of the label from dictionary.com (e.g., "An athletic activity ..." for "Sports"), and a sentence from the opening paragraph of the label's Wikipedia article (e.g., "Business is the activity of ..." for "Business"). For sentiment classification, we simply use related terms that capture the specific sentiment (e.g., "terrible" for "Very Negative") as well as a few hand-crafted templates (e.g., "It was $t$." where $t$ is a related term).

---
[1]Please refer to Schick and Schütze [18] for more details on the pattern-verbalizer approach.

Next, we finetune pretrained models using the pattern-verbalizer approach on LABELDESC data and evaluate them for text classification. For topic classification, we use patterns and verbalizers from Schick and Schütze [106] to train on our LABELDESC examples by finetuning the model as well as the MLM head (see Section 4.4 for details). We refer to training on LABELDESC data as LABELDESCTRAINING. In experiments, we show that LABELDESCTRAINING consistently improves accuracy (average improvement of 17-19%) over zero-shot classification across multiple topic and sentiment datasets (Table 4.9). We also show that LABELDESCTRAINING can decrease accuracy variance across patterns compared to zero-shot classification (Table 4.10), thus being less sensitive to the choice of pattern.

We then conduct additional experiments to reveal the value of LABELDESCTRAINING under various circumstances. To study the impact of verbalizer choice, we experiment with uninformative (randomly initialized) and adversarial (intentionally mismatched) verbalizers (Section 4.6.1). While accuracy drops slightly, both settings are still much more accurate than zero-shot classification with its original verbalizers. That is, LABELDESCTRAINING is able to compensate for knowledge-free or even adversarial verbalizer choice. We also compare to finetuning a randomly initialized classifier head without any patterns or verbalizers, again finding accuracy to be higher than zero-shot (Section 4.6.2). Collectively, our results demonstrate that LABELDESCTRAINING leads to strong performance that is less sensitive than zero-shot classification in terms of pattern/verbalizer choice, while also not requiring a pretrained MLM head.

Since LABELDESC data focuses entirely on the labels without seeking to capture the input text distribution, we would hope that it would exhibit stable performance across datasets with the same labels. So, we compare LABELDESCTRAINING to the approach of training on a small supervised training set from one domain and testing on another (Section 4.6.4). In multiple cases, LABELDESCTRAINING actually attains higher accuracy than few-shot supervised learning tested on out-of-domain test sets, even when hundreds of manually labeled training examples are used (albeit from a different input domain).

In summary, this paper shows several benefits of LABELDESCTRAINING. First, once a practitioner identifies a label set of interest for zero-shot classification, it only requires a few minutes to collect the kind of LABELDESC data shown in Table 4.1, and training on this data improves over zero-shot by 17-19% absolute. Second, LABELDESCTRAINING leads to greater robustness to pattern/verbalizer choice than zero-shot. Third, LABELDESC data are domain independent with regard to the distribution of the inputs; a single LABELDESC training set can be used for any text classification task as long as it contains the same labels. Our experiments show that this independence to input distribution leads to stable accuracy across domains, even attaining higher accuracy than out-of-domain few-shot learning on a few cases.[2]

## 4.2 Related Work

One common approach in zero-shot text classification is to transfer knowledge from seen labels [107], which requires observed labels and a notion of label similarity. Some sources of semantic knowledge used for this purpose include multiple modalities [108], label relationships in knowledge graphs [109], and word representations [110, 111].

There are several other approaches to zero-shot classification. To classify documents, Chang et al. [112] used knowledge-based text representations derived from Wikipedia, and Barak et al. [113] used both Wikipedia and WordNet. However, they both require a large-scale knowledge base. Zhang et al. [114] combined label descriptions with a label hierarchy and word-to-label paths in ConceptNet, with data augmentation strategies. Yin et al. [115] used a textual entailment approach with label definitions from WordNet. Another approach that has gained popularity is self-training given label names and acquiring knowledge by mining an unlabeled dataset [116, 117]. van de Kar et al. [24] extend the mining-based approach by selecting unsupervised examples (via patterns) for training. Basile et al. [118] select label descriptions by aggregation. Meng et al. [119] use language models to generate new training examples. On the contrary, we train on a small set of domain-independent label descriptions. Our setup is influenced by Schick and Schütze [18, 106], although, instead of finetuning on training examples, we only use our LABELDESC data.

---

[2]Data and code are available at https://github.com/lingyugao/LabelDescTraining.

Autoregressive language models have also been used for zero-shot text classification; we report zero-shot and ICL results with LABELDESC data using GPT-3.5 [53]. Zhao et al. [120] found it beneficial to "calibrate" such models for this setting; this idea is not immediately applicable here due to our use of encoder-only models like RoBERTa. Calibration could be extended to encoder-only models, which we plan to explore in future work. Our work is closely related to dataless classification [112] which involves building classifiers by designing or learning a generic function that scores the compatibility of a document and label defined in natural language. We compared empirically to the dataless classification approaches of Chu et al. [121] and Chu et al. [122] who used pretrained models, naturally annotated data like that from Wikipedia categories, and unsupervised clustering techniques. There is a wealth of prior work in semi-supervised text classification [123, 124, 125]. There is also related work on generating label names [126] or label descriptions [127, 128] but for supervised text classification.

## 4.3 Tasks and LABELDESC Datasets

We evaluate on two types of tasks: *topic classification* on AGNews, Yahoo Answers, and DBPedia [1] and *sentiment classification* on the Stanford Sentiment Treebank (SST) [129], Yelp Reviews [1], IMDB [130], and Amazon Reviews Polarity [1]. We consider both binary and 5-way classification for SST and Yelp datasets (denoted as SST-2, SST-5, Yelp-2, and Yelp-5 henceforth) and only binary for IMDB and Amazon (denoted as IMDB and Amz-2 henceforth).[3] Dataset statistics are in Table 4.2. Below we describe how we construct LABELDESC data for each label set.

| dataset | #label | LD | dev | test |
|---|---|---|---|---|
| 20NG | 4 | 24 | 3389 | - |
| AGNews | 4 | 24 | 2,000 | 7,600 |
| Yahoo$_{AG}$ | | | 3,000 | 36,000 |
| Yahoo | 10 | 60 | - | 60,000 |
| DBPedia | 14 | 84 | - | 70,000 |
| Yelp-5 | 5 | 125 | 2,500 | 50,000 |
| SST-5 | | | 1,101 | 2,210 |
| Yelp-2 | 2 | 100 | 2,000 | 38,000 |
| SST-2 | | | 872 | 1,821 |
| Amz-2 | | | - | 400,000 |
| IMDB | | | - | 25,000 |

Table 4.2: Statistics of datasets we used, with '#' denoting the number of labels, LD refers to LABELDESC data.

### 4.3.1 Topic Classification

Since labels in topic classification represent general concepts, we use both subjective descriptors of the labels (e.g., related terms) and objective sources of information (e.g., dictionary definition and Wikipedia sentences) when selecting LABELDESC data. In particular, we create LABELDESC examples for the label term itself, three related terms, a selected definition from dictionary.com, and the leading sentence from the label's Wikipedia article. As there are typically multiple dictionary.com definitions for our labels, we select a single definition that best aligns with our understanding of the concept underlying the label. We use the leading Wikipedia sentence because it is typically a brief overview/definition of the concept. Most labels in the Yahoo dataset consist of two keywords (e.g., Society & Culture). For these, we use both label terms, definitions for each, and the leading Wikipedia sentences for each. As an example, the LABELDESC data for AGNews is shown in Table 4.3.

---
[3]Our method could be adopted for other tasks like natural language inference (NLI) using templates similar to how we approached sentiment classification. We leave a full exploration to future work.

| Label | Type | Training Data |
|---|---|---|
| World | terms | world | country | international | politics |
| | Wiki. | In its most general sense, the term "world" refers to the totality of entities, to the whole of reality or to everything that is. |
| | dict. | humankind; the human race; humanity |
| Sports | terms | sport | sports | racing | baseball |
| | Wiki. | Sport pertains to any form of competitive physical activity or game that aims to use, maintain or improve physical ability and skills while providing enjoyment to participants and, in some cases, entertainment to spectators. |
| | dict. | an athletic activity requiring skill or physical prowess and often of a competitive nature, as racing, baseball, tennis, golf, bowling, wrestling, boxing, hunting, fishing, etc. |
| Business | terms | business | finance | money | trade |
| | Wiki. | Business is the activity of making one's living or making money by producing or buying and selling products (such as goods and services). |
| | dict. | the purchase and sale of goods in an attempt to make a profit. |
| Sci/Tech | terms | technology | science | computer | biology |
| | Wiki. | Technology is the continually developing result of accumulated knowledge and application in all techniques, skills, methods, and processes used in industrial production and scientific research. |
| | dict. | the branch of knowledge that deals with the creation and use of technical means and their interrelation with life, society, and the environment, drawing upon such subjects as industrial arts, engineering, applied science, and pure science. |

Table 4.3: LABELDESC data for AGNews (and Yahoo$_{\text{AG}}$).

We did not tune any of these decisions experimentally, so these choices in defining LABELDESC data are almost certainly suboptimal. This suboptimality is especially likely for the "World" label in the AGNews label set. This label reflects international news, but the dictionary definition and Wikipedia article for the term "World" do not capture that sense of the word. Nonetheless, we did not change our procedure for this label because we wanted our results to reflect a real-world implementation of the idea, complete with its limitations for certain labels.

The LABELDESC instances we are using do not contain exhaustive information. We could easily extend the lists of related terms for each topic or use WordNet or other semantic knowledge resources [114]. However, one of the goals of this research is to demonstrate how simple it is to choose LABELDESC examples to improve zero-shot classification in very little time.

### 4.3.2 Sentiment Classification

We use a slightly different procedure for sentiment classification. For 5-way sentiment, we use the label verbalizer itself and four synonym terms. In addition, we write four simple templates: "It was $t$.", "A(n) $t$ experience.", "Just $t$.", and "Overall, it was $t$.", where $t$ is the label verbalizer or a synonym. For binary sentiment, we remove the neutral instances, combine the two positive labels ("Very Positive" and "Positive") into one, and combine the two negative labels ("Very Negative" and "Negative") into one. This procedure produces a total of 25 examples per label (5 terms + 5 terms $\times$ 4 templates) for 5-way sentiment and 50 examples per label for binary sentiment. Since these LABELDESC instances are domain-independent, we use the same data for both for 5-way sentiment (Yelp-5 and SST-5) and for binary sentiment (Yelp-2, SST-2, IMDB-2, Amz-2). As an example, the LABELDESC data for Yelp-5 and SST-5 is shown in Table 4.4.

| Label | Type | Training Data |
|---|---|---|
| Very Negative | terms | awful \| terrible \| horrendous \| horrible \| dreadful |
|  | sent. | It was $t$. \| A(n) $t$ experience. \| Just $t$. \| Overall, it was $t$. |
| Negative | terms | bad \| unpleasant \| unsatisfying \| lousy \| subpar |
|  | sent. | It was $t$. \| A(n) $t$ experience. \| Just $t$. \| Overall, it was $t$. |
| Neutral | terms | okay \| mediocre \| decent \| average \| alright |
|  | sent. | It was $t$. \| A(n) $t$ experience. \| Just $t$. \| Overall, it was $t$. |
| Positive | terms | good \| nice \| fine \| pleasant \| neat |
|  | sent. | It was $t$. \| A(n) $t$ experience. \| Just $t$. \| Overall, it was $t$. |
| Very Positive | terms | great \| amazing \| excellent \| fantastic \| outstanding |
|  | sent. | It was $t$. \| A(n) $t$ experience. \| Just $t$. \| Overall, it was $t$. |

Table 4.4: LABELDESC data for Yelp-5 and SST-5. "Sent." and "$t$" refer to hand-crafted sentence templates and terms, respectively.

### 4.3.3 Hyperparameter Tuning

We adhere to the "true" zero-shot setting where hyperparameters cannot be tuned on a development set for the task of interest [106]. Therefore, we use a separate dataset for hyperparameter tuning - the 20 Newsgroups (20NG, henceforth) [131] - a topic classification dataset with twenty labels. We select only four labels from 20NG for our purposes: *talk.religion.misc*, *rec.autos*, *sci.med*, and *talk.politics.guns*. We chose these four labels because they are sufficiently distinct that we expect tuning to be informative for other real-world classification datasets; many of the other 20NG labels are highly technical or similar to one other, e.g., the pair *comp.sys.ibm.pc.hardware* and *comp.sys.mac.hardware* as well as the pair *comp.os.ms-windows.misc* and *comp.windows.x*. We follow the same strategy as for topic classification above when constructing LABELDESC data for 20NG. The selected hyperparameters are used for both topic and sentiment classifications.

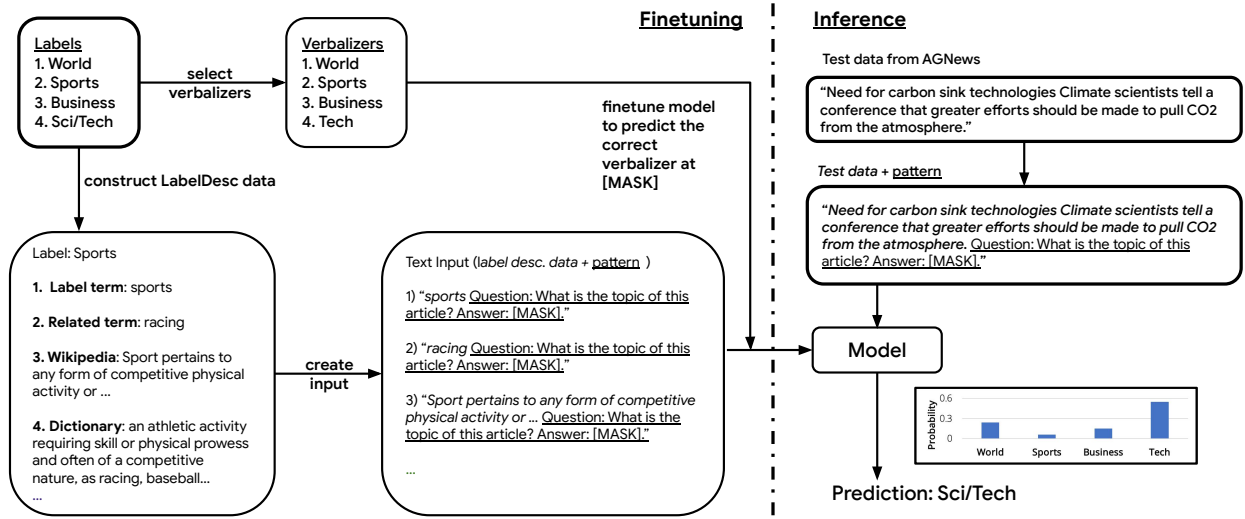

Figure 4.1: Overview of our proposed method, including the construction of LABELDESC data, the format of the text input, and the target used for both model finetuning and inference during test time. We present text inputs labeled as "Sports" from the topic classification task, and use one of our patterns (see Table 4.6) here as an illustration. Note that all our LABELDESC datasets are balanced, with each pattern being associated with a unique finetuned model checkpoint.

## 4.4 Experimental Settings

The following settings are used in our experiments. Unless stated otherwise, we use the pretrained RoBERTa-base ($b$) and RoBERTa-large ($l$) models [20] for all experiments since RoBERTa is the predominant choice in related zero-shot and dataless research [18, 24, 117]. Additionally, for every dataset, we use the entire available *test* sets for evaluation.

### 4.4.1 Patterns and Verbalizers

### 4.4.2 Zero-shot Classification Baseline

We use the standard "pattern-verbalizer" approach for topic and sentiment classification. The set of verbalizers used can be found in Table 4.5. For choosing verbalizers, we follow the choices of Schick and Schütze [18] for AGNews, Yahoo, Yelp-5, and SST-5. We follow van de Kar et al. [24] in choosing verbalizers for Yelp-2, SST-2, IMDB, and Amz-2 and we select verbalizers for DBPedia and 20NG ourselves.

| Dataset | Verbalizers |
|---|---|
| 20NG | talk.religion.misc ↦ religion, rec.autos ↦ automobile, sci.med ↦ medicine, talk.politics.guns ↦ gun |
| AGNews | World ↦ World, Sports ↦ Sports, Business ↦ Business, Sci/Tech ↦ Tech |
| Yahoo | Society & Culture ↦ Society, Science & Mathematics ↦ Science, Health ↦ Health, Education & Reference ↦ Education, Computers & Internet ↦ Computer, Sports ↦ Sports, Business & Finance ↦ Business, Entertainment & Music ↦ Entertainment, Family & Relationships ↦ Relationship, Politics & Government ↦ Politics |
| DBPedia | Company ↦ company, Educational institution ↦ school, Artist ↦ artist, Athlete ↦ sports, Office holder ↦ politics, Mean of transportation ↦ transportation, Building ↦ building, Natural place ↦ natural, Village ↦ village, Animal ↦ animal, Plant ↦ plant, Album ↦ album, Film ↦ film, Written work ↦ book |
| Yelp-5 SST-5 | Very Negative ↦ terrible, Negative ↦ bad, Neutral ↦ okay, Positive ↦ good, Very Positive ↦ great |
| Yelp-2 SST-2 IMDB Amz-2 | Negative ↦ awful, Positive ↦ great |

Table 4.5: Verbalizers selected for each dataset.

Each pattern comprises a prompt including a [MASK] symbol placed before or after the text input, and we aim to predict the masked token. For example, a prompt is added after the input $x$ to frame classification as a question answering task, e.g., "$x$ Question: What is the topic of this newsgroup? Answer: [MASK]." We use RoBERTa-base/large with its MLM head for zero-shot experiments. Although the model is able to predict any token within its vocabulary, we choose only among the set of verbalizers, which are designed to be semantically coherent with class labels and tokenized into a single token by the model's tokenizer.

For 20NG, we remove headers, quotes, and footers. For AGNews, we concatenate the headlines and the text body of the news articles. For Yahoo dataset, we concatenate the

| type | id | patterns |
|---|---|---|
| Q&A | 1 | $x$ Question: What is the topic of this article? Answer: [MASK]. |
| | 2 | $x$ Question: What is the category of this article? Answer: [MASK]. |
| | 3 | $x$ Question: What is the topic of this article? Answer: [MASK] |
| | 4 | $x$ Question: What is the category of this article? Answer: [MASK] |
| PROMPT | 1 | $x$ Category: [MASK]. |
| | 2 | $x$ Class: [MASK]. |
| | 3 | $x$ Topic: [MASK]. |
| | 4 | $x$ Theme: [MASK]. |
| | 5 | $x$ Category: [MASK] |
| | 6 | $x$ Class: [MASK] |
| | 7 | $x$ Topic: [MASK] |
| | 8 | $x$ Theme: [MASK] |
| | 9 | [MASK] News: $x$ |
| | 10 | [MASK] NEWS: $x$ |

(a) Patterns for AGNews

| type | id | patterns |
|---|---|---|
| Q&A | 1 | $x$ Question: What is the sentiment of this text? Answer: [MASK]. |
| | 2 | $x$ Question: What is the writer's opinion in this text? Answer: [MASK]. |
| | 3 | $x$ Question: What is the sentiment of this text? Answer: [MASK] |
| | 4 | $x$ Question: What is the writer's opinion in this text? Answer: [MASK] |
| PROMPT | 1 | $x$ Opinion: [MASK]. |
| | 2 | $x$ Feeling: [MASK]. |
| | 3 | $x$ Sentiment: [MASK]. |
| | 4 | $x$ Summary: [MASK]. |
| | 5 | $x$ Opinion: [MASK] |
| | 6 | $x$ Feeling: [MASK] |
| | 7 | $x$ Sentiment: [MASK] |
| | 8 | $x$ Summary: [MASK] |
| | 9 | [MASK] Sentiment: $x$ |
| | 10 | [MASK] SENTIMENT: $x$ |

(b) Patterns for sentiment classification

Table 4.6: Patterns where $x$ refers to the given text.

title, the question, and the top answer to it. And for IMDB and Amazon Reviews Polarity datasets, we concatenate the title and the content.

For topic classification tasks, we use the PROMPT and Q&A patterns from Schick and Schütze [106], which amounts to 14 patterns. For AGNews, we use "news/article" in the pattern templates, while for Yahoo we replace this with "question", and for 20NG we use "newsgroup". The patterns are shown in Table 4.6 (a). For the sentiment classification tasks, we create new Q&A patterns such as "$x$ Question: What is the sentiment of this text? Answer: [MASK]." and PROMPT patterns such as "$x$ Sentiment: [MASK]." where $x$ is the input text. There are 14 sentiment patterns in total shown in Table 4.6 (b).

|  |  |  | lr | steps |
|---|---|---|---|---|
| MLM | LDT | base | 5e-7 | 2160 |
|  |  | large | 5e-7 | 1920 |
|  | MISMATCHED | base | 5e-5 | 2160 |
|  |  | large | 5e-6 | 3000 |
|  | RANDOM | base | 5e-5 | 2160 |
|  |  | large | 5e-6 | 3240 |
| classifier |  | base | 1e-5 | 1920 |
|  |  | large | 1e-6 | 2280 |

Table 4.7: Hyperparameters (learning rate, training steps) selected by tuning on 20NG with RoBERTa.

|  |  |  | pattern | id |
|---|---|---|---|---|
| MLM | LDT | base | prompt | 9 |
|  |  | large | prompt | 7 |
|  | MISMATCHED | base | qa | 3 |
|  |  | large | qa | 1 |
|  | RANDOM | base | qa | 3 |
|  |  | large | prompt | 6 |

Table 4.8: Tuned pattern and pattern id for each model.

### 4.4.3 LABELDESCTRAINING

We use the same settings as the zero-shot baseline except that we finetune the models on LABELDESC data. We do not use any target task data for tuning or early stopping. Instead, we fix hyperparameter values, including number of training steps, by tuning on 20NG following the process described below.

We used LABELDESC data for the four selected 20NG labels as our training data and the original 20NG data (training and test sets) as our dev set, restricted to the four selected labels shown in Section 4.3. We preprocessed the data by removing headers, quotes, and footers. We used a batch size of 1 and tuned over a set of five learning rates ({5e-7, 1e-6, 5e-6, 1e-5, 5e-5}). Models were trained for 3500 training steps, evaluating on the dev set after each epoch, i.e., every 24 training steps since it's the size of LABELDESC dataset for 20NG.

After fine-tuning on 20NG, the hyperparameters are selected as shown in Table 4.7. The tuned patterns are listed in Table 4.8.[4] To our knowledge, this method works well when we adapt to other datasets. However, we also observe that there are fluctuations in the dev accuracy curve for 20NG during training, and we select the training steps in the middle of the flatter part of curves rather than the peak point for robustness. We suggest changing training steps or increasing batch size if this method doesn't work well. The tuned pattern is not necessarily the best pattern after adapting to other datasets, sometimes even a little lower than the average results over all 14 patterns.

Based on tuning accuracies, we chose learning rate 5e-7 and number of training steps 2160 for RoBERTa-base and 1920 for RoBERTa-large. Additionally, we explored variations of parameter freezing, such as freezing certain layers of RoBERTa. The best setting on 20NG was to freeze the lower half of the layers (excluding the embedding layer) during finetuning, so we used this for experiments reported below.

## 4.5 Results

Table 4.9 compares standard zero-shot classification and LABELDESCTRAINING. LABELDESCTRAINING has higher accuracy across all topic and sentiment classification datasets, outperforming zero-shot by about 17% on average when using RoBERTa-base and 19% with RoBERTa-large. The results demonstrate that we can greatly improve the performance of zero-shot models with just a few training examples that provide a richer characterization of the label but still without requiring any textual inputs from the task datasets.

Table 4.10 shows that accuracy variances across patterns using LABELDESCTRAINING are much lower than the zero-shot setting, which is known to be unstable [23]. Finetuning on LABELDESC data not only improves accuracy, but also mitigates sensitivity to pattern selection.

---

[4]MISMATCHED, RANDOM and classifier settings are introduced in Section 4.6.

|  |  | AGNews | Yahoo | DBPedia | Yelp-5 | SST-5 | Yelp-2 | SST-2 | Amz-2 | IMDB | Avg. |
|---|---|---|---|---|---|---|---|---|---|---|---|
| zero-shot | b | 62.7 | 41.5 | 54.6 | 38.0 | 35.6 | 63.6 | 62.6 | 64.0 | 69.9 | 54.7 |
|  | l | 68.0 | 47.7 | 63.9 | 38.7 | 35.0 | 70.6 | 63.7 | 67.5 | 74.1 | 58.8 |
| LABELDESCTRAINING | b | 77.4 | 58.8 | 79.5 | 43.6 | 42.0 | 88.3 | 84.5 | 88.6 | 86.9 | 72.2 |
|  | l | 79.4 | 60.8 | 86.6 | 51.3 | 49.2 | 94.6 | 91.3 | 94.1 | 92.1 | 77.7 |

Table 4.9: Test accuracy (%) comparison between zero-shot classification and LABELDESC-TRAINING, $b$ = RoBERTa-base, $l$ = RoBERTa-large. For zero-shot, each result is the average over 14 patterns; and for LABELDESCTRAINING, each result is the average over 14 patterns and three random seeds per pattern. The "Avg." column shows the average accuracies across columns.

|  |  | AGNews | Yahoo | DBPedia | Yelp-5 | SST-5 | Yelp-2 | SST-2 | Amz-2 | IMDB |
|---|---|---|---|---|---|---|---|---|---|---|
| zero-shot | b | 7.4 | 7.0 | 18.9 | 4.3 | 4.3 | 10.7 | 11.0 | 10.3 | 13.2 |
|  | l | 7.8 | 8.2 | 9.7 | 7.8 | 7.7 | 15.7 | 14.3 | 13.7 | 17.0 |
| LDT | b | 5.0, 5.1, 5.0 | 1.7, 1.6, 1.6 | 4.5, 4.5, 4.5 | 2.0, 2.1, 2.2 | 1.8, 1.4, 1.5 | 2.1, 2.8, 2.4 | 2.5, 2.3, 1.9 | 1.3, 1.2, 1.4 | 1.8, 2.3, 1.4 |
|  | l | 5.3, 6.4, 4.6 | 2.1, 2.0, 2.3 | 3.2, 2.9, 3.2 | 2.4, 2.5, 2.4 | 1.6, 1.2, 1.5 | 1.1, 2.5, 1.4 | 1.2, 2.8, 1.6 | 0.9, 1.9, 0.8 | 1.1, 1.4, 1.2 |

Table 4.10: Standard deviations of test accuracy (%) across 14 patterns for each test dataset. For LABELDESCTRAINING (LDT in the table), three random seeds were used so we show three standard deviations, one per random seed. All standard deviations over patterns are smaller for LDT than the corresponding values for zero-shot.

### 4.5.1 Comparisons to the State of the Art

We compare to state-of-the-art (SOTA) results from the literature in Table 4.11 (we show results using RoBERTa-base to better compare to other methods). For this comparison, we use only a single pattern with LABELDESCTRAINING, since doing so reflects more of a real-world use case than averaging over 14 patterns. We choose a single pattern for each of RoBERTa-base and large by tuning on 20NG as we did for other hyperparameters.[5] We use three random seeds and report average accuracies and standard deviations over seeds.

Chu et al. [121] and Chu et al. [122] are dataless classification approaches [112] that include single-encoder and dual-encoder methods; the latter include the idea of embedding documents and labels and performing classification via semantic retrieval; we report their non-ensemble results in Table 4.11.

---
[5]Please refer to Table 4.8 for details. We use the same setting for Table 4.12.

|  |  | AGNews | Yahoo | DBPedia | Yelp-5 | Yelp-2 | SST-5 | SST-2 | Amz-2 | IMDB |
|---|---|---|---|---|---|---|---|---|---|---|
| LABELDESCTRAINING | b | 84.6±0.3 | 59.9±0.3 | 82.4±1.2 | 42.0±0.4 | 84.8±0.6 | 44.3±0.1 | 88.2±0.2 | 89.6±0.4 | 83.4±0.4 |
|  | l | 85.1±1.0 | 61.2±0.3 | 88.5±0.4 | 52.5±1.2 | 95.3±0.4 | 49.4±1.1 | 91.4±0.8 | 94.5±0.3 | 92.9±0.1 |
| Chu et al. [121] | b | 68.8 | 57.8 | 81.9 | - | 67.3 | - | 65.0 | 66.8 | - |
| Chu et al. [122] | b | 75.1 | 60.0 | 88.6 | - | - | - | - | - | - |
| Schick and Schütze [106] | 10 | 79.5±2.2 | 58.4±2.7 | - | 44.3±2.5 | - | - | - | - | - |
|  | 100 | 87.5±0.8 | 65.3±1.0 | - | 54.8±1.5 | - | - | - | - | - |
| van de Kar et al. [24] | b | 79.2 | 56.1 | 80.4 | - | 92.0 | - | 85.6 | 92.0 | 86.7 |

Table 4.11: Test accuracy (%) comparison to state-of-the-art methods. 10/100 = # labeled examples used.

Schick and Schütze [106] use labeled training data (10 or 100 examples, see Table 4.11) for each task, which differs from the domain-independent LABELDESC examples which are agnostic to the domain of the textual inputs.[6] From van de Kar et al. [24], we include the highest accuracies.

The results of LABELDESCTRAINING are comparable to other methods across datasets. For sentiment classification, LABELDESCTRAINING performs better than dataless classification [121] by a large margin for all datasets and is competitive with van de Kar et al. [24] and Schick and Schütze [18]. Our method is better than that of van de Kar et al. on topic datasets (AGNews, Yahoo, and DBPedia) but not sentiment datasets except for SST-2. van de Kar et al. [24] search for naturally occurring data in large corpora; texts expressing sentiment are well-represented in corpora, while texts for topics in a fixed label set may be rarer. LABELDESCTRAINING trains on balanced data from a fixed label set, leveraging available knowledge resources to inform about topics.

Although van de Kar et al. [24] do not report 5-way classification results for Yelp or SST, we report results for both datasets (including base and large models) so that future work can compare to our results in this table. We recommend tuning zero-shot and few-shot methods on datasets that are excluded from the final comparison, like 20NG in this paper.

---

[6]We only include results with PROMPT and Q&A patterns (14 patterns for topic and 16 for sentiment) from Schick and Schütze [106], since those are similar to the pattern types we used for LABELDESCTRAINING.

|  |  | AGNews | Yahoo | DBPedia | Yelp-5 | Yelp-2 | SST-5 | SST-2 | Amz-2 | IMDB |
|---|---|---|---|---|---|---|---|---|---|---|
| LABELDESCTRAINING | b | 84.3±0.1 | 57.5±0.7 | 82.0±1.5 | 41.6±1.2 | 83.1±0.5 | 45.3±0.6 | 86.7±0.6 | 90.8±0.4 | 83.1±0.6 |
|  | l | 85.5±0.6 | 57.5±0.7 | 88.1±0.6 | 53.8±1.9 | 95.4±0.4 | 51.4±1.3 | 90.3±0.7 | 94.2±0.3 | 94.1±0.2 |
| `text-davinci-003` (zero-shot) | - | 80.2 | 58.5 | 70.1 | 47.2 | 92.3 | 49.3 | 89.3 | 93.3 | 78.9 |
| `text-davinci-003` (ICL) | - | 83.9 | 61.1 | 84.2 | 57.0 | 92.9 | 51.2 | 92.3 | 95.1 | 88.3 |

Table 4.12: Test accuracy (%) comparison to `text-davinci-003` on test set subsets.

### 4.5.2 Comparisons Involving GPT-3.5

Our method not only works for MLM-style models like RoBERTa, but also for autoregressive models. In Table 4.12, we show zero-shot and in-context learning (ICL), where we use the entire LABELDESC data for the task as ICL demonstrations, with `text-davinci-003` (GPT-3.5; 53). Due to our restricted budget, we decided to use only 1,000 test instances for each test dataset in GPT-3.5 experiments, while ensuring that the label distribution remains consistent with that of the full test dataset. It is well known that ICL is sensitive to a variety of design choices, including the order of the demonstrations [73, 132]. For ICL demonstrations, we included all LABELDESC data for a task to make predictions for each test instance. To avoid the "recency bias" (i.e., the tendency to predict labels that occur towards the end of the prompt; 66), we randomly shuffle the order of demonstrations. We left other parameters untouched. GPT-3.5 with ICL using LABELDESC data outperforms zero-shot GPT-3.5 on all datasets, showing the value of LABELDESC data even if in-domain inputs are unavailable. In comparison to GPT-3.5 flavors, LABELDESCTRAINING (RoBERTa-large) performs better on AGNews, DBPedia, Yelp-2, SST-5, and IMDB, and is competitive across other datasets.

## 4.6 Analysis and Discussion

One of the primary requirements of the zero-shot approach is the availability of pattern-verbalizer pairs [18, 106]. Here, we study several variations of LABELDESCTRAINING to investigate whether we can simplify or remove components of these pattern-verbalizer pairs. We first experiment with changing verbalizers to gauge the impact of verbalizer choice for LABELDESCTRAINING (Section 4.6.1). Next, we conduct classification experiments that do not use patterns or verbalizers at all (Section 4.6.2).

Furthermore, we include one more baseline, i.e., the model finetuned on the 20NG LABELDESC data and patterns to analyze the generalizability (Section 4.6.3). We also report additional experiments in which we measure the multi-domain robustness of LABELDESCTRAINING compared to a standard procedure of training on one domain and testing on an out-of-domain test set (Section 4.6.4). Finally, we take a closer look at label-wise performance to better understand how LABELDESCTRAINING outperforms zero-shot classification (Section 4.6.5).

### 4.6.1 Impact of Verbalizers

In this section we report experiments with LABELDESCTRAINING without meaningful verbalizers and even with adversarially chosen verbalizers. We explore two different verbalizer settings:

- RANDOM: We add $c$ new words, i.e., RANDOM1, RANDOM2, ..., RANDOM$c$, where $c$ is the number of dataset labels, to the model's vocabulary and randomly initialize their embeddings. This setting prevents the use of any prior knowledge in the verbalizer embeddings.

- MISMATCHED: We shuffle the original mapping of labels to verbalizers, ensuring that each verbalizer maps to a different label than in the original LABELDESCTRAINING setting. Since we are still finetuning the embeddings, finetuning can help the model recover from this mismatched initialization.

The results are shown in Table 4.13. Since we still use the MLM head for these results, we refer to them as "MLM, RANDOM" and "MLM, MISMATCHED". While LABELDESCTRAINING performs better than RANDOM, and RANDOM is better than MISMATCHED, both are better than zero-shot on average. These results suggest that LABELDESC data can partially compensate when the quality of the verbalizers is unknown or poor, at least to improve over zero-shot.

|  | | AGNews | Yahoo | DBPedia | Yelp-5 | SST-5 | Yelp-2 | SST-2 | Amz-2 | IMDB | Avg. |
|---|---|---|---|---|---|---|---|---|---|---|---|
| zero-shot | b | 62.7±7.4 | 41.5±7.0 | 54.6±18.9 | 38.0±4.3 | 35.6±4.3 | 63.6±10.7 | 62.6±11.0 | 64.0±10.3 | 69.9±13.2 | 54.7±9.7 |
|  | l | 68.0±7.8 | 47.7±8.2 | 63.9±9.7 | 38.7±7.8 | 35.0±7.7 | 70.6±15.7 | 63.7±14.3 | 67.5±13.7 | 74.1±17.0 | 58.8±11.3 |
| LDT$_{20\text{NG}}$ | b | 61.8±7.0 | 49.4±5.2 | 72.9±7.8 | 34.6±4.6 | 36.5±3.7 | 67.7±10.3 | 63.4±9.7 | 67.2±9.6 | 72.5±10.5 | 58.4±7.6 |
|  | l | 72.4±6.8 | 54.4±4.3 | 71.9±10.8 | 36.3±5.7 | 36.6±7.1 | 63.4±13.0 | 56.9±8.7 | 60.9±10.2 | 67.5±15.2 | 57.8±9.1 |
| LDT | b | 77.4±4.9 | 58.8±1.6 | 79.5±4.4 | 43.6±2.1 | 42.0±1.6 | 88.3±2.5 | 84.5±2.2 | 88.6±1.4 | 86.9±1.8 | 72.2±2.5 |
|  | l | 79.4±5.0 | 60.8±2.1 | 86.6±3.0 | 51.3±2.4 | 49.2±1.6 | 94.6±1.8 | 91.3±2.0 | 94.1±1.3 | 92.1±1.2 | 77.7±2.3 |
| MLM$_r$ | b | 77.3±4.0 | 54.3±3.9 | 81.3±7.3 | 38.1±3.8 | 37.0±3.2 | 78.4±10.0 | 73.3±7.9 | 80.0±9.9 | 73.8±9.6 | 65.9±6.6 |
|  | l | 75.2±5.0 | 58.0±3.0 | 85.4±13.0 | 46.4±3.3 | 43.4±2.9 | 90.8±7.6 | 84.1±6.8 | 90.2±7.1 | 87.4±6.2 | 73.4±6.1 |
| MLM$_m$ | b | 73.1±5.6 | 50.1±5.4 | 72.6±8.1 | 36.8±2.8 | 35.8±2.5 | 80.1±7.2 | 75.8±5.0 | 81.8±6.8 | 76.7±6.0 | 64.8±5.5 |
|  | l | 66.4±8.6 | 44.5±4.9 | 73.1±7.3 | 41.9±4.0 | 38.7±4.2 | 83.6±6.5 | 78.1±6.0 | 85.0±6.0 | 77.7±6.9 | 65.4±6.0 |
| classifier | b | 72.5±5.5 | 57.1±0.7 | 87.7±2.6 | 40.3±1.3 | 39.4±2.5 | 86.9±2.9 | 79.7±1.1 | 89.1±0.9 | 80.6±3.6 | 70.4±2.3 |
|  | l | 77.8±1.5 | 50.9±7.3 | 78.2±1.0 | 42.4±1.6 | 35.3±9.2 | 93.3±0.9 | 86.6±1.4 | 93.7±0.5 | 85.7±2.0 | 71.5±2.8 |

Table 4.13: Test accuracies (%) for several variations of LABELDESCTRAINING. The standard deviations are computed over 14 patterns for zero-shot; 3 random seeds for the classifier (no patterns); and both 14 patterns and 3 random seeds for LABELDESCTRAINING on 20NG, LABELDESCTRAINING, RANDOM, and MISMATCHED (LDT$_{20\text{NG}}$, LDT, MLM$_r$, and MLM$_m$ in Table).

### 4.6.2 Classifiers Without Patterns or Verbalizers

Since finetuning on LABELDESC data outperforms zero-shot results with RANDOM verbalizers, we also evaluate its performance without patterns, i.e., using a standard randomly initialized softmax classifier. The input is the original text without any patterns and we use a two-layer classification head on top of the [CLS] token representation of the pretrained models.

The bottom two rows of Table 4.13 show the results. The classifiers are close to that of the MLM/RANDOM setting and still much higher than zero-shot on average, suggesting that it is not necessary to use patterns, verbalizers, or even the pretrained MLM head in order to outperform zero-shot classifiers. If it is difficult to select verbalizers or design patterns for a particular classification task, using a classifier that has been finetuned on a small LABELDESC dataset may serve as a strong alternative to the pattern-verbalizer approach.

### 4.6.3 Cross-Task Generalizability

We report results on the model finetuned on the 20NG LABELDESC data and patterns, i.e., LABELDESCTRAINING on 20NG (LDT$_{20\text{NG}}$), in Table 4.13. While the patterns for the reported datasets are different from those used for 20NG, especially for sentiment datasets, they have similar structures (see Table 4.6). For RoBERTa-base, LDT$_{20\text{NG}}$ often outperforms zero-shot results, except for AGNews and Yelp-5. However, for RoBERTa-large, while LDT$_{20\text{NG}}$ outperforms the zero-shot results on all topic classification datasets, it's worse on sentiment classification except for SST-5.

### 4.6.4 Multi-Domain Evaluation

Since LABELDESC examples are domain-independent, they can be used for multiple datasets that have the *same* labels. To assess the multi-domain performance of LABELDESCTRAINING, we compare it to supervised few-shot learning in which a model is trained on data from one domain and then evaluated on a different domain with the same label set (i.e., training on SST-5 and evaluating on Yelp-5). To create multi-domain test sets for a single topic label set, we keep AGNews as it is and create a new subsampled version of Yahoo as follows: (1) "Politics & Government" and "Society & Culture" texts are assigned the label "World", (2) "Sports" texts are labeled "Sports", (3) "Business & Finance" texts are labeled "Business", and (4) "Science & Mathematics" and "Computers & Internet" texts are labeled "Sci/Tech". Other Yahoo texts are removed. We refer to this new version of the Yahoo dataset as Yahoo$_{\text{AG}}$. For sentiment classification, we choose two dataset pairs that share label sets, i.e., SST-5 and Yelp-5.

We do not change anything about the LABELDESCTRAINING configuration for these experiments. We simply evaluate the same model on multiple test sets, reporting average accuracies over patterns.

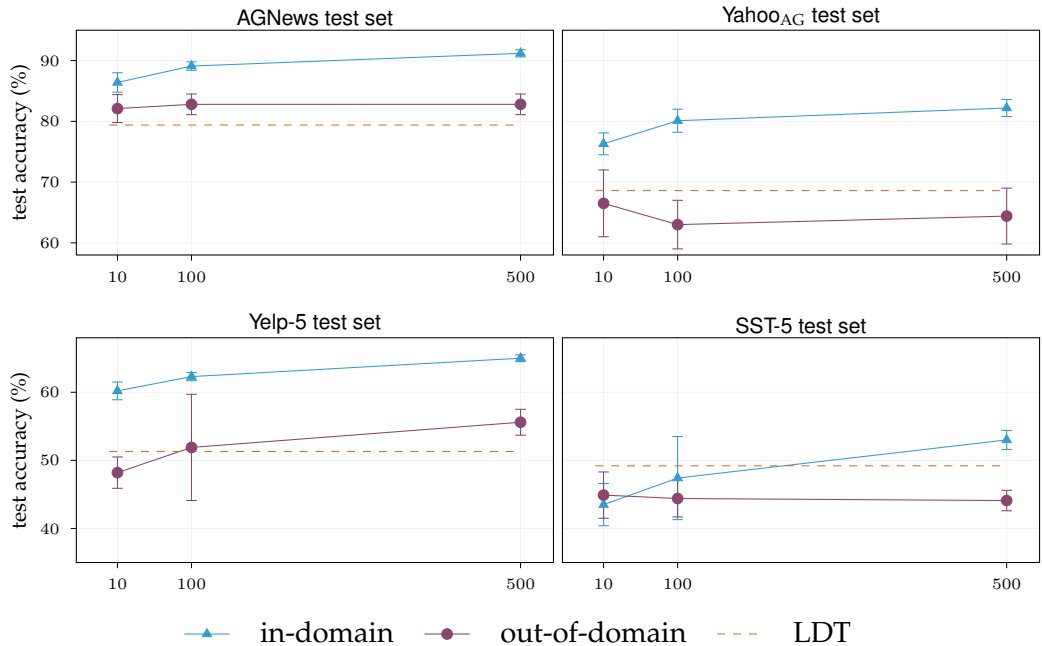

Figure 4.2: Domain transfer results, where the X-axis shows the number of training examples per label.

For few-shot setup, we create datasets with 10, 100, and 500 training examples per label. For *in-domain* experiments, train, dev, and test sets are drawn from the same domain/dataset, whereas for *out-of-domain* experiments, train and dev sets are drawn from one domain and the test set is drawn from another domain. We tune learning rates over the same ranges as mentioned earlier and use batch sizes 1, 2, and 4 for 10, 100, and 500 examples per label, respectively. We train for 15 epochs and select the checkpoint from the best epoch selected by the dev set.

The results using RoBERTa-large are shown in Figure 4.2. As we would expect, testing on out-of-domain data leads to accuracy drops but adding more out-of-domain training data reduces this gap. LABELDESCTRAINING, shown as an orange dotted line, outperforms supervised few-shot learning in some cases, such as training on AGNews and testing on Yahoo$_{\text{AG}}$, even with 500 examples per label (upper-right plot in Figure 4.2). We see the same trend when the supervised model is trained on Yelp-5 and tested on SST-5 (lower-right plot in Figure 4.2). In 3 out of 4 cases, LABELDESCTRAINING outperforms supervised few-shot out-of-domain learning with 10 examples per label, outperforming 100 in 2 out of 4 cases.

|              | zero-shot         | LABELDESCTRAINING |
|--------------|-------------------|-------------------|
| World        | 61.5$_{\pm 15.1}$ | 81.0$_{\pm 4.3}$  |
| Business     | 63.6$_{\pm 7.1}$  | 74.9$_{\pm 4.7}$  |
| Sports       | 88.2$_{\pm 3.9}$  | 92.7$_{\pm 4.5}$  |
| Sci/Tech     | 55.0$_{\pm 11.4}$ | 67.8$_{\pm 9.3}$  |

Table 4.14: AGNews label-wise F1 (RoBERTa-large).

|               | zero-shot         | LABELDESCTRAINING |
|---------------|-------------------|-------------------|
| Very Negative | 11.2$_{\pm 14.9}$ | 25.8$_{\pm 5.7}$  |
| Negative      | 37.6$_{\pm 21.2}$ | 62.5$_{\pm 2.0}$  |
| Neutral       | 1.2$_{\pm 2.9}$   | 10.8$_{\pm 5.5}$  |
| Positive      | 46.0$_{\pm 5.8}$  | 48.2$_{\pm 4.9}$  |
| Very Positive | 12.1$_{\pm 15.0}$ | 58.0$_{\pm 4.0}$  |

Table 4.15: SST-5 label-wise F1 (RoBERTa-large).

### 4.6.5 Label-wise Investigation

To better understand why LABELDESCTRAINING outperforms zero-shot, we report label-specific F1 scores in Tables 4.14 and 4.15.

For AGNews, the zero-shot classifiers have low F1 scores for the World label, probably because the verbalizer "World" is much less coherent and less representative of the actual label than others like "Sports." LABELDESCTRAINING improves F1 on the World label by roughly 20 points, while the improvement for Sports is only about 4 points. Likewise, the F1 scores for "Very Negative", "Very Positive", and "Neutral" are very low for the zero-shot models on SST-5, indicating that those labels are being largely ignored. Again, LABELDESCTRAINING shows large improvements in F1 for some of these labels, especially

| text ([headline][text body] for AGNews) | zero-shot | LABELDESCTRAINING |
|---|---|---|
| [Homeless families total 100,000][The figure for homeless families in England has topped 100,000 for the first time.] | Business | World |
| [Shifting signs in North Korea][Kim Jong Il dials back his personality cult as protest activities pick up.] | Sports | World |
| [GM, Daimler Go Green][Team-up will help the companies compete and fill gaps in both firms' portfolios.] | Sci/Tech | Business |
| (U)nrelentingly stupid. | Positive | Very Negative |
| Still, I'm not quite sure what the point is... | Positive | Negative |
| This 72-minute film does have some exciting scenes, but it's a tad slow. | Positive | Neutral |

Table 4.16: AGNews/SST-5 data that are correctly classified with LABELDESCTRAINING but not in zero-shot settings.

"Very Positive". These trends are likely due in part to the differences verbalizer probabilities, e.g., "good" and "bad" occur more frequently than "great" and "terrible". The LABELDESC data is balanced, which helps to mitigate the ignoring of labels, even though the task test sets are not all balanced. Table 4.16 shows examples that are incorrectly classified by zero-shot models but are correctly classified by the LABELDESCTRAINING models.

## 4.7 Conclusions

We presented LABELDESCTRAINING, a method for improving the accuracy of zero-shot classification by using small, curated datasets that simply describe the labels for a task in natural language. Our method is 17-19% more accurate than zero-shot on average across a range of datasets. LABELDESCTRAINING is also more robust to the choices required for zero-shot classification, such as patterns and verbalizers. Furthermore, LABELDESC data is domain agnostic and therefore can used for any text classification task as long as it contains the same set of labels. LABELDESCTRAINING can even outperform a supervised approach that uses training data from a different domain. One future direction would be to apply the idea to structured prediction, NLI, and natural language generation tasks. Another would be to investigate ways to reduce the dependence of pretrained models on patterns and verbalizers, such as directly calibrating the marginal probabilities of verbalizers with the goal of minimizing biases of pretrained models.

# CHAPTER 5

# Ambiguity-Aware In-Context Learning with Large Language Models

In-context learning has proven to be effective for recent powerful large language models [19, 47, 133]. Since these models are highly sensitive to prompts [29, 73, 77, 134], it is crucial to select better demonstrations (input-output pairs) to improve performance. However, this presents additional challenges because the underlying mechanisms are not yet fully understood.

In this chapter, we utilize retrieved demonstrations that are misclassified and fall near the decision boundary of the test example for in-context learning, which outperforms the retrieval-based baselines and resolves model ambiguity regarding the test example.

This chapter is based on [38].

## 5.1 Introduction

Leveraging LLMs [19, 47, 133] via *in-context learning* (ICL) is now a popular strategy for improving downstream task performance, wherein the model is able to perform a task by simply being conditioned on the task definition and/or few task *demonstrations* (input-output examples) [47, 135].

As ICL gets increasingly adopted, it has brought to light [29, 73, 77, 134] that LLMs are sensitive to the choice of prompts, making "prompt engineering" for different tasks challenging and time-consuming. However, prompt engineering does not have to be a complete guessing game; rather it can be governed by some data-derived signals. For example, selecting demonstrations that are semantically similar to a new input has been shown to be more effective than randomly sampled demonstrations [29, 31, 33]. In their approaches, a text retriever is used to select the top-$k$ training examples for each test example based on the *input text*. The motivation is that using information from existing similar situations will help solve a new problem [136].

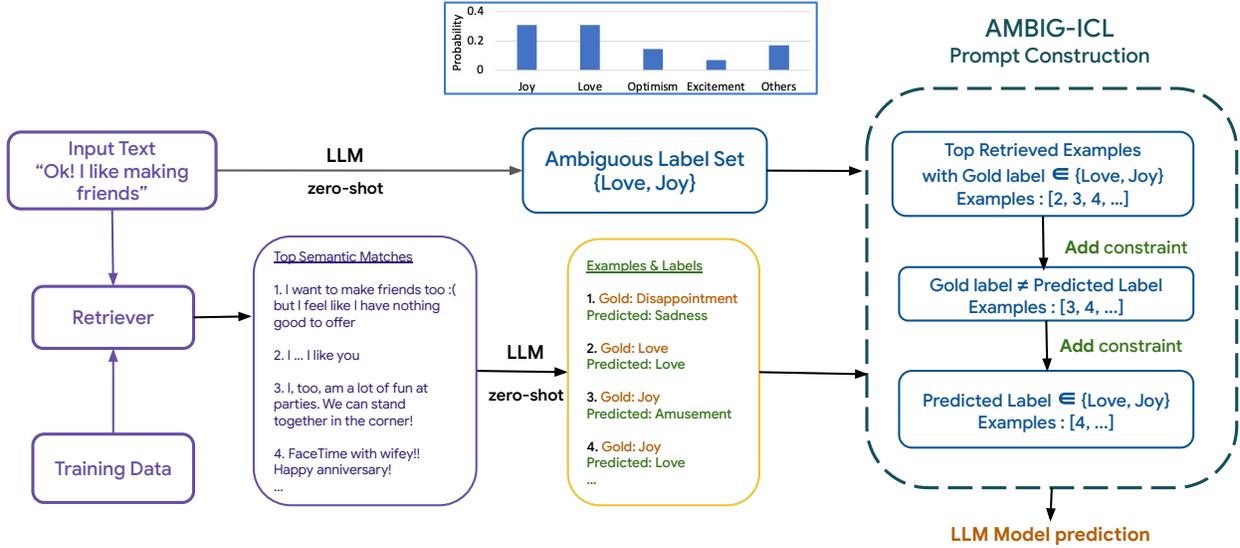

Figure 5.1: Overview of our proposed method for selecting ICL demonstrations: For each test example, we first use a retriever to rank training data by semantic similarity. At the same time, we identify the ambiguous label set for each test example and also obtain the output predictions on the retrieved training data. Next, we apply three constraints on the top-ranked demonstrations which are: 1) select those demonstrations whose gold label is in the ambiguous label set, 2) select those which are also mis-classified by the model, and 3) select those mis-classified examples whose predicted label is in the ambiguous label set. Finally, we construct prompts with selected ICL demonstrations to get the final model predictions.

However, the solely input-based selection does not explicitly capture the LLM's existing knowledge about the task-specific *label space* of both the ICL demonstration as well as the test input. For example, on a five-way sentiment classification task (SST [129]), we have observed that the Flan-PaLM 2 model (size L) [57] is confused between two specific

<figure>

| SST, zero-shot, Flan-PaLM 2 (L) | | | | | |
|---|---|---|---|---|---|
| Actual \ Predicted | VPos | Pos | Neu | Neg | VNeg |
| VPos | 340 | 59 | 0 | 0 | 0 |
| Pos | 236 | 252 | 20 | 2 | 0 |
| Neu | 42 | 123 | 108 | 100 | 16 |
| Neg | 1 | 41 | 81 | 439 | 71 |
| VNeg | 1 | 0 | 2 | 152 | 124 |

</figure>

Figure 5.2: Confusion Matrix of zero-shot experiments on SST with Flan-PaLM 2 (L). Labels: VPos (Very Positive), Pos (Positive), Neu (Neutral), Neg (Negative), VNeg (Very Negative).

labels, 'Very Negative' and 'Negative,' a lot more than say between 'Neutral' and 'Very Negative', as shown in Figure 5.2. This motivates us to investigate whether *the model's existing knowledge can also be leveraged to select even more effective demonstrations.*

Specifically, we derive signals from the underlying LLM about the output label space of both the new test example and the training data from which we select the demonstrations. As motivated above, the model's ambiguity around the new test example's output label will help us know *what the model is most confused about*, which in turn can be used to select those demonstrations that help reduce this confusion. For selecting such demonstrations from the training data, we propose to consider not only the ground truth labels paired with these demonstrations, but also the *usefulness* by looking at their model prediction. First, given a test example and pool of training data, for each test example we use an off-the-shelf retriever to retrieve top-$k$ examples that have similar input text. For each test example, we identify an *ambiguous label set* of two output labels that the model is most confused about. Next, we select top-ranked demonstrations such that their ground truth labels lie in the above label set. To further find *useful* demonstrations, we identify those which are mis-classified by the model; the intuition is that showing the model a previously mis-classified demonstration could force it to correct it [137, 138]. Finally, on top of the mis-classified demonstrations we add a constraint to select only those demonstrations whose model prediction falls within the ambiguous label set, i.e., near the test example's decision boundary.

To test our hypothesis, we focus on multi-class text classification tasks that have fine-grained nuance in the label space. We conduct extensive experimentation across three tasks, namely SST [129], GoEmotions [139], and EDOS (Task-B) [140], all of which have fine-grained label space, making the model more likely to be confused across labels. Our key observations are:

- Incrementally adding constraints, i.e., 1) considering *label ambiguity of test example*, 2) limiting ICL demonstrations to *mis-classified demonstrations*, and 3) considering *label ambiguity of training examples* leads to +1.5%, +2.2%, +2.6% improvement in F1 macro scores over the retrieval-based ICL, averaged across all datasets (Table 5.3).

- We find that adding such label-based constraints helps more on a smaller model, i.e., on Flan-PaLM 2 (M) (+3.9% gain) compared to +1.4% gain on Flan-PaLM 2 (L).

- We also attribute this success of our proposed methods to the observation that the *ambiguous label set acts as a good proxy to the gold test label*, and as noted by Min et al. [141], labels in the ICL demonstrations bias the model predictions the most. Therefore, showing the models the 'likely' gold label guides the model to make the correct prediction (Table 5.5).

## 5.2 Related Work

The performance of large language models (LLMs) is significantly influenced by the quality of ICL demonstrations, as demonstrated in multiple studies [29, 120, 134]. Consequently, the focus on retrieving superior demonstrations has increased. One prominent strategy is to finetune a retriever for specific tasks by similarity metrics [33, 142, 143] or by scores derived from language models [28, 144]. While some works introduce an unified retriever trained across various tasks [34, 145] for generalizabilty, another direction is to leverage off-the-shelf retrievers. Liu et al. [29] propose a KNN-based method to select ICL demonstrations based on semantic similarities; Margatina et al. [31] select ICL demonstrations with active learning algorithms based on uncertainty, diversity, and similarity, and show that selecting based on input text similarity consistently outperforms other methods; and Agrawal et al. [30] focus on selecting diverse demonstrations as well as promoting n-gram overlap between demonstrations and test examples. In our work, we adopt the off-the-shelf retriever approach as our focus is to show the generalizability of our approach across different classification tasks. However, we expect that our method will also benefit from a task-specific retriever. Additionally, to the best of our knowledge, we are the first ones to leverage the LLM's existing knowledge surrounding the test example for selecting demonstrations. Prior works have typically explored the LLM's existing knowledge, considering the model prediction for the training data.

Luo et al. [32] use the LLM prediction score on the training data to train a task-specific retriever, and also use Chain-of-Thought prompting [146] to improve model performance. Some works [73, 147] have found that ordering of the ICL demonstrations also affects the downstream performance, that is why in Table 5.2 we report the results across three shuffle orders. These works are orthogonal to our work but can be used in combination with our proposed methods.

## 5.3 Proposed Method

Typically, in an ICL regime, we assume access to training data $\mathcal{D}_{train} = \{(x_0, y_0), \cdots, (x_T, y_T)\}$ from which the goal is to select $d$ demonstrations to be used as the prompt. As motivated in the introduction, we follow a three-step approach for selecting demonstrations: For each test example, we 1) extract semantically similar examples from $\mathcal{D}_{train}$, 2) identify the ambiguous label-set and 3) extract model predictions for $\mathcal{D}_{train}$ to identify mis-classified examples. Below, we describe each step in more detail and how they are used together to select the "best" demonstrations.

**Extract Semantically Similar Demonstrations** Typically, in this approach, demonstrations are selected for each test example $x_t$ by finding those examples from the $\mathcal{D}_{train}$ that are semantically similar to the test input. The motivation is that observing demonstrations that are similar to the new input text will act as a hint for the model [31]. This requires the use of a retriever $R$, either an off-the-shelf one such as [29, 30, 31, 32] or a retriever trained specifically for that task [28, 33]. For each test example $x_t$, the retriever $R$ is used to rank examples from $\mathcal{D}_{train}$ based on semantic similarity of the text inputs. Top-$k$ input-output pairs are then selected from the ranked $\mathcal{D}_{train}$ to be used as ICL demonstrations.

**Identify Ambiguous Label-Set** As we can observe from the confusion matrix in Figure 5.2, the model is often confused between two labels. We hypothesize that in addition to semantic similarity, providing demonstrations that help the model resolve this ambiguity will help the model correct itself. Thus, as a next step, we construct a prompt $\theta$ for the test example $x_t$, and use the model log-likelihood to score each output label $l \in L$ given the prompt. Using this we identify top-2 labels that have the highest scores, which we refer to as the "**ambiguous label set**" of $x_t$, denoted as $\mathcal{L}_{ambig,t} = \{\hat{y}_t^{(1)}, \hat{y}_t^{(2)}\}$, where $\hat{y}_t^{(1)}$ and $\hat{y}_t^{(2)}$ are the first and second most likely labels, respectively.

**Extract Mis-classified Demonstrations**  The final component in our recipe is to consider the model prediction of the training data. While prior work [31, 141, 148] has looked at training data label-space from the lens of ground-truth labels, i.e., whether to retain them in the ICL or not, we aim to look at label-space from the perspective of model predictions. Specifically, we are interested in identifying "hard" demonstrations, i.e., examples on which the model makes mistakes. We hope that showing the model such examples with their ground truth labels will force the model to correct itself. Prior work has underscored the potential value of leveraging mis-classified examples from the training set to enhance model performance [137, 138], but it hasn't been tested for ICL demonstration selection on text classification. In addition to the mis-classified examples, we further constrain the model prediction of these mis-classified examples to be one of the ambiguous labels, identified in the above step. Given that we already know which output labels the model is confused between for the test examples, showing the model those demonstrations (with their ground truth labels) which fall near the decision boundary will likely guide the model to choose the correct label for the test input.

## 5.4 Experimental Setup

### 5.4.1 Model

We experiment with the Flan-PaLM 2 model, an instruction-tuned model which is fine-tuned on the Flan dataset [69, 149] based on PaLM-2 [57], a multilingual large language model pretrained on web documents, books, code, mathematics and conversational data. We chose these models as Luo et al. [32] find that retrieved demonstration for ICL works better with instruction-tuned models than general LLMs (e.g., GPT). In particular, we experiment with two variants of the model, namely Flan-PaLM-2 (M) and Flan-PaLM-2 (L), where the latter is a larger model.[1] The ICL demonstrations are selected using an off-the-shelf retriever which is finetuned on mT5-base [150] using the unsupervised objective proposed by Izacard et al. [151]. Since the order of demonstrations may impact the model performance [73, 147], we randomly shuffle the order of demonstrations for three random seeds and report the average results.

---

[1] Please refer to Anil et al. [57] for more details on the models.

|  | train | dev | test |
|---|---|---|---|
| EDOS | 3,398 | 486 | 970 |
| SST | 8,544 | 1,101 | 2,210 |
| GoEmotions | 23,485 | 2,952 | 2,978 |

Table 5.1: Number of examples in each dataset split.

### 5.4.2 Data

As mentioned above, the Flan-PaLM 2 models are finetuned on the Flan dataset which is a mixture of many supervised datasets. Specifically, we choose three text classification datasets that satisfy the following desiderata: 1) the output label space shows fine-grained nuance that spans multiple labels, and 2) these datasets are *not* part of the Flan mixture to avoid any inherent bias from the underlying model. We describe them below, with dataset statistics shown in Table 5.1. All datasets are in English.

**EDOS (Task-B):**   The Task B of Explainable Detection of Online Sexism [140], is a *topic classification* task where the sexist content is classified into four categories, i.e., 1) Threats, plans to harm & incitement, 2) Derogation, 3) Animosity, and 4) Prejudiced Discussion.

**SST:**   The Stanford Sentiment Treebank (SST, Socher et al. [129]) is a 5-way *sentiment classification* dataset for movie reviews with labels: Very Negative, Negative, Neutral, Positive, and Very Positive.

**GoEmotions:**   The GoEmotions [139] is a multi-class sentiment classification dataset with "neutral" and 27 emotional classes, e.g., "admiration" and "fear", collected from Reddit comments. As the label space is very large and given that we have limited sequence length, it becomes even more crucial to select a concise but effective prompt. [2]

### 5.4.3 Prompt Construction

We show our templates in Table C.1 in Appendix (we use 4-shot as an example for few-shot). Task definitions are listed below, denoted by $x_{defn}$:

- **EDOS:** Given a text input, the task is to classify the input as being a Threat, Prejudiced, Animosity, or Derogation category of sexism. Threat refers to language where an individual expresses intent and/or encourages others to take action against women which inflicts or incites serious harm and violence against them. It includes threats of physical, sexual or privacy harm. Prejudiced refers to language which denies the existence of discrimination, and justifies sexist treatment. It includes denial and justification of gender inequality, excusing women's mistreatment, and the ideology of male victimhood. Animosity refers to language which expresses implicit or subtle sexism, stereotypes or descriptive statements. It includes benevolent sexism, i.e., framed as a compliment. Derogation refers to language which explicitly derogates, dehumanises, demeans or insults women. It includes negative descriptions and stereotypes about women, objectification of their bodies, strong negative emotive statements, and dehumanising comparisons. It covers negative statements directed at a specific woman and women in general.

- **SST:** Given sentences from movie reviews, the task is to classify the sentences as being a Great, Good, Okay, Bad, or Terrible category of sentiment. Great refers to language that expresses extremely positive sentiment. Good refers to language that expresses positive sentiment, but not to the extreme. Okay refers to language that is neutral, i.e., neither expresses clear positive nor negative sentiments. Bad refers to language that expresses negative sentiment, but not to the extreme. Terrible refers to language that expresses extremely negative sentiment.

---
[2] We exclude 24,848 examples (19,925 from training set, 2,474 and 2,449 from dev and test set, respectively) that have multiple labels annotated for a single input, for a simpler experimental setting. We refer the reader to Demszky et al. [139] for more information on the single-label setting.

- **GoEmotions:** Given sentences from Reddit comments, the task is to classify the sentences as being an Admiration, Approval, Annoyance, Gratitude, Disapproval, Amusement, Curiosity, Love, Optimism, Disappointment, Joy, Realization, Anger, Sadness, Confusion, Caring, Excitement, Surprise, Disgust, Desire, Fear, Remorse, Embarrassment, Nervousness, Pride, Relief, or Grief category of emotions.

### 5.4.4 Baselines

We compare our proposed method against the following baselines:

**Frequent Label (FREQ).** Select the most frequent label as the model prediction for all test examples.

**Zero-shot ICL (ZERO).** For each test example $x_t$, we prepend the task definition to each test input and prompt the models.[3] To obtain the model prediction, we use the model log-likelihood to score each output label $l \in L$, given the prompt. Then, we select the label with the highest score. $y_t = \mathrm{argmax}_L \mathrm{score}(l, \theta)$ where $\theta$ refers to the prompt specifically used for this setting, and *score* refers to the model's log-likelihood.

**Static N-shot ICL (STATIC-$N$).** We select a fix set of $N$ demonstrations from $\mathcal{D}_{train}$, one for each of the $N$ output labels ($N = |\mathcal{L}|$). Note that these demonstrations are static for all test examples. Thus, we concatenate the task definition, $N$ demonstrations, and test example $x_t$ as the prompt for ICL and use the log-likelihood scores, as described above, to get the model prediction.

---
[3]Please refer to Section 5.4.3 for the exact prompt and prompt template used in this setting, as well as for few shot settings such as the subsequent STATIC-$N$ and RETR.

**Retrieval-based ICL (RETR).** Unlike above, where we used the same prompt for all test inputs, in this baseline, we retrieve demonstrations for each test input $x_t$. We use an off-the-shelf retriever $R$ (Section 5.4.1) to retrieve $k$ nearest neighbors $\{x_{1,t}, \cdots, x_{k,t}\}$ from $\mathcal{D}_{train}$, similar to Das et al. [33]. We encode the input text of the training set and the test example, and rank the training data by the inner product of the vectors. Of these $k$ examples, we select $n = 4, 8$ as ICL demonstrations.[4]

### 5.4.5 Proposed Method: AMBIG-ICL

As described in Section 5.3, our proposed method considers both semantic similarity and the label ambiguity for selecting demonstrations. Below, we summarize our proposed model variants. For each setting, we first retrieve the top-$k$ most similar examples from the training data $\mathcal{D}_{train}$ for each test example $x_t$. We denote these candidates by $R(x_t) = \{(x_{0,t}, y_{0,t}), \cdots, (x_{k,t}, y_{k,t})\}$. At the same time, for each $x_t$, we also identify the ambiguous label-set $\mathcal{L}_{ambig,t} = \{l_i, l_j | l \in L\}$. This set contains the top-2 labels, $l_i$ and $l_j$, that the model is most confused about, where both labels belong to the set $L$ of all output labels.

**+GOLD** Select those examples from $R(x_t)$ as demonstrations where the ground truth label of each demonstration belongs to the ambiguous label set of $x_t$ denoted by:

$$\text{ICL}(x_t) = \left\{ \begin{array}{l} (x_i, y_i) \text{ if } y_i \in \mathcal{L}_{ambig,t} \\ \text{for } (x_i, y_i) \in R(x_t) \end{array} \right\}$$

---

[4]We chose $k = 4, 8$ for two reasons: a) to limit the sequence length to 1024 tokens for faster inference, and b) in some settings we found $k = 4$ often outperforming $k = 8$ (Table 5.2), which led us to believe that adding more examples will not benefit much.

**+GOLD+MIS** Select those examples from $R(x_t)$ as demonstrations where the ground truth labels fall in $\mathcal{L}_{ambig,t}$ and they are mis-classified, denoted by:

$$\text{ICL}(x_t) = \left\{ \begin{array}{l} (x_i, y_i) \text{ if } y_i \in \mathcal{L}_{ambig,t}, \hat{y}_i \neq y_i \\ \text{for } (x_i, y_i) \in R(x_t) \end{array} \right\}$$

Note that the model predictions ($\hat{y}$) on the $R(x_t)$ are obtained from the ZERO model.

**+GOLD+MIS+PRED** Select those examples from $R(x_t)$ as demonstrations where the ground truth labels fall in $\mathcal{L}_{ambig,t}$. Ensure they are mis-classified and with an additional constraint, that their model predictions also fall within $\mathcal{L}_{ambig,t}$, denoted by:

$$\text{ICL}(x_t) = \left\{ \begin{array}{l} (x_i, y_i) \text{ if } y_i \in \mathcal{L}_{ambig,t}, \hat{y}_i \neq y_i, \\ \hat{y}_i \in \mathcal{L}_{ambig,t} \text{ for } (x_i, y_i) \in R(x_t) \end{array} \right\}$$

Same as above, the model predictions on the training data are obtained from ZERO.

For all our proposed model variants, we select $n$ demonstrations where $n = 4$ and $n = 8$.

## 5.5 Results and Discussion

We report all our results in Table 5.2. Specifically, we use the F1 macro scores to compare the model performance, as all our tasks have unbalanced datasets.[5] First, we note across all three tasks, our proposed methods outperform the baselines.

We also note that the zero-shot model (ZERO), which only uses a task definition but no task demonstrations, is already a strong baseline for both the Flan-PaLM 2 models (M/L). In particular, comparing the average scores of the few-shot baselines and ZERO, we find that ZERO outperforms few-shot baselines by 1.4% on Flan-PaLM 2 (M), but the larger model Flan-PaLM 2 (L) benefits from the addition of ICL demonstrations (+1.4% gain). This is because larger-parameter models make better use of in-context learning [152, 153, 154]. Interestingly, we also observe that for SST and GoEmotions, the Flan-PaLM 2 (L) model achieves higher performance with $n = 4$ over $n = 8$, which highlights that quantity does not necessarily lead to better performance.

---
[5]We report the accuracy, precision and recall in Appendix.

|  |  | EDOS | | SST | | GoEmotions | | Avg. | |
|---|---|---|---|---|---|---|---|---|---|
|  |  | M | L | M | L | M | L | M | L |
| Baselines | FREQ | 15.9 | 15.9 | 7.5 | 7.5 | 0.8 | 0.8 | 8.1 | 8.1 |
|  | ZERO | 50.7 | 60.5 | 49.2 | 54.1 | <u>40.5</u> | 43.4 | <u>46.8</u> | 52.7 |
|  | STATIC-$N$ | <u>51.1</u>$_{\pm0.3}$ | 58.5$_{\pm0.4}$ | 50.3$_{\pm0.4}$ | <u>56.5</u>$_{\pm0.3}$ | 34.3$_{\pm0.5}$ | 44.4$_{\pm0.3}$ | 45.2 | 53.1 |
|  | RETR-4 | 48.5$_{\pm0.3}$ | <u>62.3</u>$_{\pm0.4}$ | 49.9$_{\pm0.3}$ | 55.4$_{\pm0.3}$ | 38.3$_{\pm0.3}$ | 46.2$_{\pm0.4}$ | 45.6 | <u>54.6</u> |
|  | RETR-8 | 47.1$_{\pm0.2}$ | 61.8$_{\pm0.1}$ | <u>51.5</u>$_{\pm0.1}$ | 55.2$_{\pm0.4}$ | 37.5$_{\pm0.2}$ | <u>46.7</u>$_{\pm0.1}$ | 45.4 | 54.6 |
| Ours | AMBIG-4 |  |  |  |  |  |  |  |  |
|  | +GOLD | 49.3$_{\pm0.6}$ | 62.6$_{\pm0.2}$ | 51.5$_{\pm0.4}$ | 56.1$_{\pm0.0}$ | 40.7$_{\pm0.3}$ | **48.2**$_{\pm0.2}$ | 47.2 | 55.6 |
|  | +GOLD+MIS | 52.2$_{\pm0.5}$ | 61.7$_{\pm0.9}$ | 52.3$_{\pm0.1}$ | 57.4$_{\pm0.1}$ | 40.1$_{\pm0.2}$ | 47.6$_{\pm0.1}$ | 48.2 | 55.6 |
|  | +GOLD+MIS+PRED | **53.9**$_{\pm0.5}$ | 62.9$_{\pm0.4}$ | 53.3$_{\pm0.4}$ | **58.0**$_{\pm0.0}$ | 42.3$_{\pm0.5}$ | 47.7$_{\pm0.2}$ | **49.8** | **56.2** |
|  | AMBIG-8 |  |  |  |  |  |  |  |  |
|  | +GOLD | 47.5$_{\pm0.1}$ | **63.2**$_{\pm0.2}$ | 52.9$_{\pm0.1}$ | 56.5$_{\pm0.6}$ | 42.0$_{\pm1.2}$ | 47.7$_{\pm0.1}$ | 47.5 | 55.8 |
|  | +GOLD+MIS | 50.4$_{\pm0.4}$ | 62.0$_{\pm0.4}$ | 53.4$_{\pm0.1}$ | 57.7$_{\pm0.1}$ | **43.9**$_{\pm0.2}$ | 47.6$_{\pm0.4}$ | 49.2 | 55.8 |
|  | +GOLD+MIS+PRED | 50.9$_{\pm0.6}$ | 62.7$_{\pm0.2}$ | **54.3**$_{\pm0.2}$ | 57.2$_{\pm0.3}$ | 41.3$_{\pm0.3}$ | 47.4$_{\pm0.3}$ | 48.8 | 55.8 |

Table 5.2: F1 macro (%) comparison between our baselines (top) and our proposed methods (bottom) with Flan-PaLM 2 (M/L). 4 or 8 refers to the number of ICL demonstrations. The best performance across all methods is **highlighted**, and the best-performing baseline is <u>underlined</u>. The "Avg." column shows the average scores across all datasets. The standard deviations are computed over three random seeds, with the order of demonstrations shuffled.

**Considering output label space is more important than semantic similarity.** Within the few-shot methods, where we use ICL demonstrations along with the task definition, we compute from Table 5.3 that our proposed methods AMBIG-* outperform retrieval-based models (RETR-*) by +3.0% (avg.) for Flan-PaLM 2 (M), and by +1.2% (avg.) for Flan-PaLM 2 (L), suggesting that *considering output label space for selecting demonstrations is as important as considering the input similarity.* In particular, we find that considering mis-classified demonstrations that fall near the test example's decision boundary leads to the overall

|       | ZERO | STATIC-$N$ | AMBIG-ICL |       |       |
|-------|------|------------|-----------|-------|-------|
|       |      |            | +GOLD     | +MIS  | +PRED |
| M     | 1.3  | -0.2       | 1.9       | 3.3   | **3.9** |
| L     | -1.9 | -1.5       | 1.1       | 1.1   | **1.4** |
| all   | -0.3 | -0.9       | 1.5       | 2.2   | **2.6** |

We omitted RETR in the table, which are inherently zero as we compare against RETR.

For both RETR and AMBIG-ICL, we average results on both 4 and 8 shots before computing differences.

Table 5.3: F1 macro (%) differences compared to RETR, averaged across all datasets as detailed in Table 5.2. M and L refers to Flan-PaLM 2 sizes, and "all" is averaged on results of size M and L. "+MIS" and "+PRED" refer to "+GOLD+MIS" and "+GOLD+MIS+PRED", respectively.

best performance. In Table 5.4, we show the demonstrations selected for the $n = 4$ setting for one example of the GoEmotions task. We see that for the test input "Ok! I like making friends", the RETR method retrieved similar examples from $\mathcal{D}_{train}$ (all examples refer to *friends*). Now from the ZERO model, we calculated the model prediction scores and found that *Love* and *Joy* are the two labels the model is most confused about. However, because we do not consider any test example ambiguity in RETR, only one of the retrieved examples represents the labels *Love* or *Joy*, which are the two labels the model is most confused about for this test example. On the other hand, in the AMBIG-ICL setting, because of our constraints, all the examples chosen for ICL belong to the ambiguous label set. This allows all our proposed methods to better understand this fine-grained nuance across label space and make the correct model prediction of *Love*. Below, we conduct some analysis to further explain the way our proposed methods work.

**Considering output label space compensates for the sacrifice in semantic similarity.** As we introduce more constraints (i.e., +GOLD, +MIS, and +PRED), we find that we need to sacrifice the semantic similarity to the test input. For example, consider the 4-shot AMBIG-ICL experiment on EDOS (Task-B). To satisfy the constraints for the +GOLD setting, we need to select up to top-16 retrieved examples in order to obtain the 4 ICL demonstrations; for +GOLD+MIS we need top-55 retrieved examples and more than top-250 retrieved examples for +GOLD+MIS+PRED.[6] Clearly, by selecting lower ranked examples from the retrieved set $R(x_t)$ we are sacrificing the semantic similarity to the test input. While previous studies, such as [29, 31, 33], have indicated that greater semantic similarity can enhance model performance, we can see that our methods can still outperform the retrieval-based baselines which prioritize it.

| **Test Example: Ok! I like making friends** | $\mathcal{L}_{\text{ambig},t}$: **Love, Joy** | **Gold label: Love** |
|---|---|---|
| RETR | 1. Disappointment: I want to *make friends* too :( but I feel like I have nothing good to offer<br>2. Joy: I, too, am a lot of fun at parties. We can stand together in the corner!<br>3. Gratitude: Thanks. I am. I *make some new friends*.<br>4. Disapproval: Not really. My group of *friends* are awesome in every way possible except they are homophobic | Predicted: Joy |
| AMBIG-ICL | | |
| +GOLD | 1. Joy: I, too, am a lot of fun at parties. We can stand together in the corner!<br>2. Love: I ... I like you<br>3. Love: Married to the love of my life. LOL<br>4. Love: I do. but some people love it | Predicted: Love |
| +GOLD+MIS | 1. Joy: I, too, am a lot of fun at parties. We can stand together in the corner!<br>2. Love: Too cute for me. Why cant i have a boyfriend *[NAME]*<br>3. Joy: FaceTime with wifey!! Happy anniversary!<br>4. Love: Stick around! Would love your input POV! | Predicted: Love |
| +GOLD+MIS+PRED | 1. Joy: FaceTime with wifey!! Happy anniversary!<br>2. Joy: She want to take it slow, I can see that... I deal with those girls all the time, they my favorite<br>3. Love: Ha! I like that one.<br>4. Love: Ooh I like that one :) | Predicted: Love |

Table 5.4: Example demonstrations selected by the RETR and our proposed method AMBIG-ICL for the GoEmotions task, for $n = 4$. Each demonstration comprises the input text and the ground truth label, as selected from the training data. On Flan-PaLM 2 (L), where RETR mis-classified it as "Joy", AMBIG-ICL predicted correctly under all three settings.

---

[6]We set a strict constraint on our selection (top-250 retrieved example for +GOLD, and top-250 misclassified retrieved examples for the other two). If there aren't sufficient examples for +GOLD+MIS+PRED within the top-250 misclassified retrieved example, we fall back on the previous setting (+GOLD+MIS).

|  | EDOS | | SST | | GoEmotions | |
|---|---|---|---|---|---|---|
|  | M | L | M | L | M | L |
| 4-shot | 42.6 | | 29.6 | | 21.6 | |
| 8-shot | 42.5 | | 28.6 | | 20.5 | |
| AMBIG-4 | | | | | | |
| +GOLD | 49.5 | **50.3** | **46.5** | **47.1** | **41.3** | **41.9** |
| +GOLD+MIS | 46.4 | 44.3 | 46.1 | 44.3 | 38.7 | 38.8 |
| +GOLD+MIS+PRED | 48.3 | 42.3 | 46.1 | 44.6 | 37.8 | 40.7 |
| AMBIG-8 | | | | | | |
| +GOLD | **50.3** | **50.3** | 46.0 | 46.8 | 41.2 | 41.7 |
| +GOLD+MIS | 46.9 | 43.8 | 46.4 | 44.7 | 38.7 | 38.6 |
| +GOLD+MIS+PRED | 48.8 | 42.9 | **46.5** | 44.9 | 37.5 | 40.3 |

Table 5.5: Average percentage (%) of examples in the top $4, 8$ retrieved demonstrations that share the same gold labels with the test example.

**The ambiguous label set is a good proxy for the test gold label.** While Min et al. [141] find that using pseudo-demonstrations, i.e., demonstrations with random labels instead of the ground truth labels, does not affect the downstream performance much, Lyu et al. [35] find that for demonstrations that are similar to the test input, such as those from a retriever, pseudo-demonstrations hurt the performance.

They refer to this as the copying-effect hypothesis, which says that the "model prediction is biased towards the labels paired with the inputs in the demonstrations, especially when the inputs are similar to the test inputs." This, in turn, suggests that the best performance could be achieved if the labels paired with the inputs are the same as the gold label of the test example. Given that we do not know the gold label of the test example apriori, the question then becomes *how do we approximate the gold label?*. We find that our *ambiguous label set* acts as a close proxy. In Table 5.5, we compute how many times the label paired with ICL demonstrations is the same as the test example gold label. We find that 44.2% of our proposed methods' (AMBIG) demonstrations have the same gold label as the test example on average, compared to 30.9% from the RETR method. This is why including the ambiguous label set in the demonstration selection process leads to higher performance. This analysis also sheds light on the effectiveness of retrieval-based ICL. From Table 5.5 we can see that the demonstrations selected solely based on input text similarity is only 13.3% points (avg.) behind our proposed methods. This confirms that finding demonstrations similar to the input text also leads to selecting demonstrations that have the 'likely' gold label.

|  | EDOS | | SST | | GoEmotions | |
| --- | --- | --- | --- | --- | --- | --- |
|  | M | L | M | L | M | L |
| uniform | 2.00 | | 2.32 | | 4.75 | |
| ZERO | 0.98 | 1.08 | 1.58 | 1.19 | 2.44 | 1.92 |
| STATIC-$N$ | 0.87 | 1.07 | 1.41 | 1.11 | 1.76 | 1.77 |
| RETR-4 | 0.78 | 0.97 | 1.40 | 1.06 | 1.89 | 1.70 |
| RETR-8 | 0.82 | 0.96 | 1.38 | 1.04 | 1.79 | 1.69 |
| AMBIG-4 | | | | | | |
| +GOLD | **0.77** | 0.93 | 1.39 | 1.02 | 1.86 | 1.43 |
| +GOLD+MIS | 0.85 | 0.98 | 1.41 | 1.06 | 1.92 | 1.48 |
| +GOLD+MIS+PRED | 0.86 | 1.00 | 1.42 | 1.07 | 1.92 | 1.46 |
| AMBIG-8 | | | | | | |
| +GOLD | 0.81 | **0.91** | **1.36** | **0.98** | **1.68** | **1.33** |
| +GOLD+MIS | 0.89 | 0.97 | 1.39 | 1.03 | 1.74 | 1.39 |
| +GOLD+MIS+PRED | 0.90 | 1.00 | 1.40 | 1.04 | 1.76 | 1.37 |

Table 5.6: Average entropy of predicted probability distribution. "uniform" refers to the entropy computed for an uniform probability distribution over the labels. Lower entropy is better.

**AMBIG-ICL helps reduce the model confusion.** To understand whether including test label ambiguity indeed helps decrease the model confusion, we calculate the model entropy over the predicted probability distribution of the output labels in Table 5.6.[7] Overall, we observe that our AMBIG-* methods achieve the lowest entropy across all three datasets and models. This suggests that by explicitly identifying the point of model confusion (in this case the confusion across fine-grained labels) and selecting demonstrations that help resolve this confusion is indeed effective in reducing the confusion across labels, and thereby resulting in higher downstream performance (Table 5.2). In particular, we find that for the Flan-PaLM 2 (L), the gap between the few-shot baselines and the AMBIG-* methods is larger, perhaps because larger models are better able to use the ICL demonstrations [152, 153, 154].

---

[7]We compute entropy with a base of 2.

|  | | EDOS (F1) | SST (acc.) |
|---|---|---|---|
| DistilBERT [140] | | 55.3 | - |
| DeBERTa-v3-base [140] | | 47.9 | - |
| BERT-large [155] | | - | 56.2 |
| RoBERTa-large [155] | | - | 57.9 |
| RoBERTa-large + Self-Explaining [155] | | - | 59.1 |
| Heinsen Routing + RoBERTa Large [156] | | - | 59.8 |
| Ambig-ICL | Flan-PaLM 2 (M) | 53.9 | 54.2 |
| | Flan-PaLM 2 (L) | **63.2** | **60.2** |

Table 5.7: F1 macro/accuracy (%) comparison between our proposed method to finetuned classifiers on EDOS and SST.

We also compute the Pearson correlation coefficient between F1 macro scores and average entropy of the predicted probability distribution (shown in Table 5.2 and Table 5.6, respectively), for all the three datasets. We find that for the Flan-PaLM 2 (L) model, there is a negative correlation for all three datasets, i.e., $r = -0.78$ for EDOS, $-0.48$ for SST and $-0.92$ for GoEmotions, which suggests that lower entropy translates to higher task performance. However, for the Flan-PaLM 2 (M), we have mixed results, as $r$ is positive for EDOS ($0.47$), negative for SST ($-0.55$), and close to zero for GoEmotions ($0.03$).

## 5.6 Comparative Studies

**Comparison with finetuned classifiers.** We compare our strongest Ambig-ICL results to those of finetuned classifiers from the literature, as shown in Table 5.7.[8]

For EDOS (Task B), both DistilBERT [157] and DeBERTa-v3-base [158] are finetuned on the training set [140]. While DistilBERT outperforms Ambig-ICL on Flan-PaLM 2 (M), it's 7.9% behind Ambig-ICL on Flan-PaLM 2 (L) in F1 macro scores. For SST, both Sun et al.

---

[8]We use accuracy for SST for convenience of comparison. We do not compare on GoEmotions as we only use its single-label subset for our experiments.

|              | Macro-F1 |       |           |         |       |         |
|              | GLUE | Ethos | TweetEval | HateS18 | Poem | Average |
|---|---|---|---|---|---|---|
| Uniform      | 75.5 | 65.5 | 63.7 | 63.7 | 68.7 | 67.4 |
| Best-of-10 [134] | 75.8 | 69.1 | **68.8** | 70.6 | 72.2 | **71.3** |
| Topic [160]  | 76.2 | 62.4 | -    | -    | 75.2 | -    |
| KATE [29]    | 72.3 | 71.2 | 66.3 | 66.8 | 73.2 | 69.9 |
| Ambig-ICL    | **76.6** | **71.3** | 68.3 | **72.4** | 67.7 | **71.3** |
| ICR [159]    | **78.7** | **76.5** | **71.0** | **74.4** | **76.5** | **75.4** |

Table 5.8: F1 macro (%) reported by Xu and Zhang.

[155] and Heinsen [156] employ additional techniques beyond simple finetuning, and they perform close to but still worse than Ambig-ICL on Flan-PaLM 2 (L).

When the LLMs are larger and better at following instructions, they can be expected to outperform smaller finetuned models, especially on difficult datasets that are fine-grained and have fewer gold data. Additionally, ICL with LLMs is easily adapted to different classification tasks with only one model, whereas smaller models require finetuning for each specific task.

**Additional Ambig-ICL results from literature.** We present additional results of Ambig-ICL in Table 5.8, as reported in a later study by Xu and Zhang [159]. It provides further context and presents both the strengths and limitations of the Ambig-ICL approach.

Xu and Zhang [159] proposed In-Context Reflection (ICR), a method designed to identify demonstrations that are most likely to challenge the understanding of LLMs. They tested their approach on several benchmark datasets, including GLUE [161], Ethos [162], TweetEval [163], HateS18 [164], and Poem [165]. Their comparison results are shown in Table 5.8,[9] and Ambig-ICL outperforms all other methods, except ICR, on 3 out of the 5 datasets.

For TweetEval, they adopted three subtasks: "Hate Speech Detection", "Emotion Recognition", and "Irony Detection", and reported the average score. Given that "Hate" and "Irony" are binary classification tasks, while "Emotion" is a 4-way classification task, Tweet-Eval may be considered a less fine-grained dataset, which could make Ambig-ICL less preferable in this context.[10] Poem is a 4-way sentiment classification task for poems of different verses. As we lack access to label-wise results, it's difficult to determine the reason for failure. However, Topic [160], which learns latent concepts, achieves the best performance, offering potential insights into achieving higher performance on the Poem dataset. We note that the authors didn't define the labels (negative/positive/neutral/mixed) in their prompt as we did in our experiments. This may have added to the model's difficulty in understanding the classes, leading to a flawed ambiguous label set selection.

---

[9]Please refer to Xu and Zhang [159] for more details, and they use SBERT [166] for semantic similarity ranking for Ambig-ICL.

While our method has demonstrated promising results across various datasets, potential pitfalls remain. If the actual gold label of test example often deviates from the LLM's top two label choices in a particular dataset or model, this can indicate subpar zero-shot performance or flawed ambiguous label set selection. In these scenarios, our method may lead to unsatisfying performance, necessitating further enhancements. To mitigate these issues and improve overall performance, we encourage specifying task instructions and label definitions in the prompt.

## 5.7 Conclusion

In this work, we find that using LLM's existing knowledge (e.g., the model prediction) regarding the output label space of both the test example and the ICL demonstration pool is as important as considering the semantic similarity of the input text alone. We find that our proposed methods consistently outperform the baselines for all three tasks. Although we only consider the top-2 most ambiguous labels in selecting the ICL demonstrations, it would be interesting to expand the ambiguous label set to more than two labels. This would especially be more important for datasets like GoEmotions where the label space is large and much more fine-grained. We leave this effort for future work. Furthermore, in this work, we focus on sentence classification tasks, thus paving the way for others to use our proven techniques to also explore label ambiguity for other token/span-level tasks such as Named Entity Recognition (NER), and Part-Of-Speech (POS) tagging.

---

[10]Still, Ambig-ICL ranks as the second-best method, close to the performance of Best-of-10.

# CHAPTER 6

# Conclusion

## 6.1 Summary of Contributions

This thesis has explored the advancements in text classification by harnessing Pretrained Language Models (PLMs) to address three challenging settings: distractor selection for multiple-choice cloze questions, prompt-based zero-shot text classification, and demonstration selection for retrieval-based in-context learning. The main contributions are:

**Distractor Analysis and Selection for Multiple-Choice Cloze Questions.**

- Experimented on two datasets specifically tailored for evaluating distractor selection in multiple-choice cloze questions. One dataset consists of standalone cloze sentences, while the other includes the cloze sentence within a paragraph. Given the usual lack of direct supervision in past works, the analysis and results provide valuable insights for this domain.

- Designed context-free features as well as context-sensitive features with contextualized word representations derived from PLMs. We find models tend to favor syntactically correct distractors while differing sufficiently in semantics.

- Demonstrated that the use of PLMs significantly enhances the performance of distractor selection models, achieving results comparable to or exceeding human annotators.

Detailed quantitative analysis provided insights into the relative importance of different model components and features.

**Label-Description Training for Zero-Shot Text Classification.**

- Proposed to craft small datasets that describe task labels for zero-shot text classification, applicable for both finetuning and in-context learning.

- Showed that this label-description training method improves accuracy by 17-19% over traditional zero-shot approaches across multiple topic and sentiment classification datasets. The method also demonstrated robustness to variations in patterns and verbalizers, and LABELDESC data can partially compensate when the quality of the verbalizers is unknown or poor.

- Highlighted the domain-independent nature of the approach. In several cases, this method outperformed few-shot learning on out-of-domain datasets.

**Ambiguity-Aware In-Context Learning with Large Language Models.**

- Developed a method for in-context learning that selects demonstrations based on model predictions of both demonstrations and test examples, thereby resolving the model ambiguity about test example labels.

- Conducted extensive experiments on fine-grained text classification tasks, demonstrating that the proposed method improves F1 macro scores by up to 2.6% over traditional retrieval-based in-context learning approaches.

- Found that the set of the top two most likely labels is a good proxy for the gold label of a test example, shedding light on the effectiveness of retrieval-based ICL.

## 6.2 Future Work

While distractor selection is a highly practical application, designing distractors to better assist student learning remains an area for further exploration, and this domain still lacks sufficient datasets. Future work can use our models to collect more training data and delve deeper into the different decisions students make. Additionally, exploring the usage of question answering PLMs could be beneficial. For example, can question answering PLMs simulate students' answers, and how accurately can they capture them? Are the current annotation rules the best practice? Aside from language learning, could we expand to other disciplines?

In Chapter 4 and Chapter 5, both LABELDESCTRAINING and Ambig-ICL were tested on topic and sentiment classification tasks. A future direction is to apply these methods to other span-level tasks, such as Named Entity Recognition (NER) and Part-Of-Speech (POS) tagging, as well as to natural language generation tasks, such as generating text in different styles. Another research avenue would be to explore ways to improve the consistency of models regarding patterns and verbalizers, such as debiasing through calibration. Additionally, the mechanisms underlying retrieval-based ICL are still not fully understood, and the interaction of multiple demonstrations within a single prompt requires further study.

Although all tasks in this thesis are conducted in English, it would be interesting to harness PLMs' intrinsic knowledge for cross-linguistic knowledge transfer. Different languages encode and represent knowledge in different ways. For instance, in French, the number eighty is expressed as "quatre-vingts," where "quatre" means "four" and "vingts" means "twenties." This linguistic construction involves a mathematical operation, even though it is not explicitly stated. Investigating how well PLMs can transfer this type of numerical and linguistic knowledge from one language to another could open new possibilities for multilingual applications, including text classification.

# Appendix A

# Appendix to Chapter 3

## A.1 Dataset

There are some problematic words in the dataset, such as 'testing, test', 'find s', 'find ed' in MCDSENT/MCDPARA candidate words. There are also some extra spaces (or non-breaking spaces) at the start or end of words. To keep the words the same as what the annotators saw, we only remove leading/trailing white space, and replace non-breaking spaces with ordinary spaces. By comparing the percentages of the circumstances where spaces are included in the string before/after tokenization, we find the percentage of extra spaces presented in Table A.1. The vocabulary size after tokenization is presented in Table A.2.

| %        | headword($c$) | $c$    | headword($d$) | $d$    |
|----------|---------------|--------|---------------|--------|
| MCDSENT  | 0             | 0      | 0.0168        | 0.0332 |
| MCDPARA  | 0.0160        | 0.0307 | 0.0364        | 0.0622 |

Table A.1: Percentage of extra spaces (excluding those that are in the middle of words), where headword($c$) denotes headword of correct answer, and $d$ denotes distractor candidates of inflected forms.

|          | headword($c$) | $c$   | headword($d$) | $d$    |
|----------|---------------|-------|---------------|--------|
| MCDSENT  | 2571          | 2731  | 3514          | 11423  |
| MCDPARA  | 2683          | 4174  | 3582          | 13749  |

Table A.2: Vocabulary sizes.

| %        | sent start | sent end | para start | para end |
|----------|------------|----------|------------|----------|
| MCDSENT  | 3.058      | 0.005    | -          | -        |
| MCDPARA  | 2.640      | 0.342    | 18.272     | 22.165   |

Table A.3: Position of the candidates, where "sent" denotes sentence and "para" denotes paragraph. "para start" means that the sentence containing the blank is at the beginning of the paragraph.

### A.1.1 Distractor Annotation

The software tool suggested distractor candidates based on the following priority ranking:

1. It is in a proprietary dictionary.

2. It has the same part-of-speech (POS) as the correct answer (if POS data is available) and satisfies 1.

3. It is part of a proprietary learnable word list for the language learning course under consideration, and satisfies 2.

4. It is in the same course as the correct answer and satisfies 3.

5. It is in the same proprietary study material bundle as the correct answer and satisfies 4.

6. It is in the previous or same study material as the correct answer and satisfies 5.

7. It is in the same study material as the correct answer and satisfies 6.

8. It is in the same NGSL frequency word list band as the correct answer and satisfies 7.

9. It is not used as a distractor for another word with the same task type in the same material at the time that the distractor list for quality assurance (QA) is loaded, and satisfies 8.

### A.1.2 Context Position

Sometimes the blank resides at the start or end of the context (a sentence for MCDSENT, and a paragraph for MCDPARA), counts of which are shown in Table A.3. The percentage when there is only one sentence as context in MCDPARA is 0.894%.

| model variant | | development set | | | | test set | | | | BERT | features | best epoch | threshold |
|---|---|---|---|---|---|---|---|---|---|---|---|---|---|
| | | precision | recall | F1 | AUPR | precision | recall | F1 | AUPR | | | | |
| baseline | | 15.4 | 100 | 26.7 | - | 13.3 | 100 | 23.5 | - | - | - | - | - |
| $M_{feat}$ | | 33.3 | 64.8 | 44.0 | 35.1 | 23.2 | 59.1 | 33.3 | 25.0 | none | yes | 26 | 0.2 |
| | | 42.1 | 67.0 | **51.7** | 45.4 | 31.5 | 57.4 | 40.7 | 32.3 | base | yes | 26 | 0.2 |
| | | 41.3 | 67.1 | 51.1 | **46.7** | 32.4 | 56.6 | **41.2** | **33.9** | large | yes | 25 | 0.3 |
| $M_{ELMo}$ | none | 49.0 | 79.1 | 60.5 | 58.5 | 46.5 | 75.7 | 57.6 | 53.9 | - | no | 6 | 0.3 |
| | gru+c | 49.7 | 77.6 | 60.6 | 54.1 | 46.1 | 73.3 | 56.7 | 53.5 | - | no | 3 | 0.4 |
| | all | 52.9 | 75.8 | 62.3 | 60.4 | 48.0 | 75.9 | 58.8 | 57.6 | - | no | 2 | 0.4 |
| | none | 51.0 | 84.0 | 63.4 | 63.1 | 47.7 | 81.4 | 60.1 | 60.6 | large | yes | 3 | 0.3 |
| | gru+c | 56.9 | 72.3 | 63.7 | 59.1 | 50.6 | 75.9 | **60.8** | 58.6 | large | yes | 5 | 0.4 |
| | all | 53.5 | 80.8 | **64.4** | **63.4** | 50.8 | 75.5 | **60.8** | **59.6** | large | yes | 3 | 0.4 |
| $M_{BERT}$ | | 48.8 | 85.5 | 62.1 | 56.6 | 43.8 | 82.8 | 57.3 | 51.5 | base | no | 4 | 0.2 |
| | | 49.6 | 80.8 | 61.5 | 59.1 | 45.2 | 79.7 | 57.7 | 54.9 | large | no | 3 | 0.3 |
| | | 51.5 | 84.2 | **63.9** | 61.7 | 46.0 | 78.6 | 58.0 | 55.0 | base | yes | 6 | 0.2 |
| | | 51.4 | 81.1 | 62.9 | **64.7** | 46.4 | 79.8 | **58.7** | **57.5** | large | yes | 6 | 0.2 |

Table A.4: Results for MCDSENT tuned based on F1.

## A.2 Results Tuned Based on F1

We report our results tuned based on F1 in Table A.4 and A.5.

## A.3 Supplement for Analysis

The example for MCDPARA is as below, and two sets of its distractors are shown in Figure A.1.

- MCDPARA: A few years have passed since the Great Tohoku Earthquake occurred. It has been extremely costly to rebuild the damaged areas from scratch, with well over $200 billion dollars provided for reconstruction. However, the **availability** of these funds has been limited. However, a large portion of the money has been kept away from the victims due to a system which favors construction companies....

| model | development set | | | | test set | | | | BERT | features | best epoch | threshold |
|---|---|---|---|---|---|---|---|---|---|---|---|---|
| | precision | recall | F1 | AUPR | precision | recall | F1 | AUPR | | | | |
| baseline | 7.3 | 100 | 13.5 | - | 6.6 | 100 | 12.4 | - | - | - | - | - |
| $M_{feat}$ | 17.1 | 53.1 | 25.9 | 15.9 | 15.6 | 51.3 | 23.9 | 15.0 | - | yes | 14 | 0.1 |
| | 19.5 | 63.0 | 29.8 | 20.4 | 17.6 | 61.0 | 22.3 | **18.6** | base | yes | 22 | 0.1 |
| | 20.4 | 63.1 | **30.8** | **22.3** | 16.7 | 62.7 | **26.4** | 18.6 | large | yes | 25 | 0.1 |
| $M_{ELMo}$ | 35.2 | 55.4 | 43.1 | 37.0 | 31.2 | 54.6 | 39.8 | 33.9 | - | no | 5 | 0.3 |
| | 40.2 | 61.3 | **48.5** | **43.8** | 34.1 | 59.4 | **43.3** | 35.2 | large | yes | 5 | 0.3 |
| $M_{ELMo}(\ell)$ | 28.7 | 72.9 | 41.2 | 33.8 | 25.7 | 71.3 | 37.7 | 30.3 | - | no | 2 | 0.2 |
| | 36.2 | 67.3 | 47.1 | 40.8 | 31.0 | 65.6 | 42.1 | **37.3** | large | yes | 7 | 0.3 |
| $M_{BERT}$ | 35.8 | 64.2 | 46.0 | 39.3 | 28.9 | 64.3 | 39.9 | 34.5 | base | no | 5 | 0.2 |
| | 35.2 | 62.1 | 45.0 | 38.3 | 26.9 | 60.5 | 37.3 | 29.3 | large | no | 6 | 0.1 |
| | 44.3 | 55.4 | 49.3 | **47.3** | 34.6 | 56.2 | 42.8 | 36.7 | base | yes | 2 | 0.3 |
| | 37.8 | 63.3 | 47.4 | 44.0 | 32.7 | 66.1 | **43.7** | **38.1** | large | yes | 3 | 0.2 |
| $M_{BERT}(\ell)$ | 34.0 | 64.3 | 44.5 | 36.7 | 29.6 | 62.1 | 40.1 | 32.1 | base | no | 5 | 0.2 |
| | 43.3 | 57.5 | **49.4** | 45.4 | 33.3 | 60.9 | 43.1 | 35.8 | base | yes | 3 | 0.3 |

Table A.5: Results for MCDPARA tuned based on F1.

Figure A.1: Ranks for distractor candidates of MCDPARA question "However, the **availability** of these funds has been limited." along with annotations.

# Appendix B

# Appendix to Chapter 4

## B.1 Domain Transfer

All results on RoBERTa-base/large are shown in Figure B.1 and B.2.

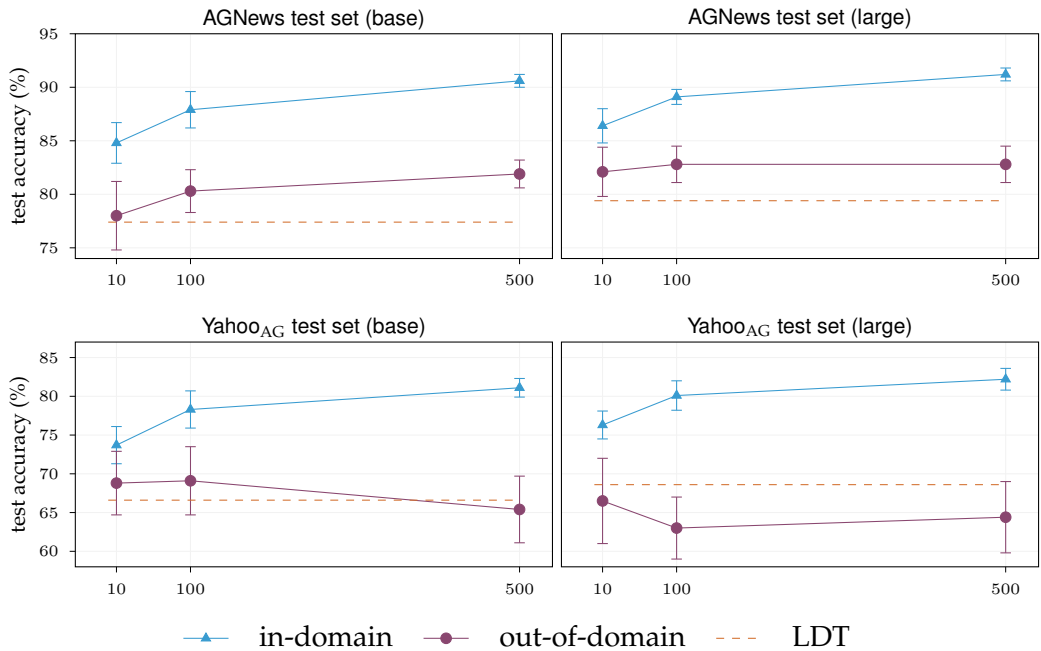

Figure B.1: Domain transfer results of topic classification, where X-axis depicts the number of training examples per label. "base/large" in parenthesis denotes RoBERTa-base/large.

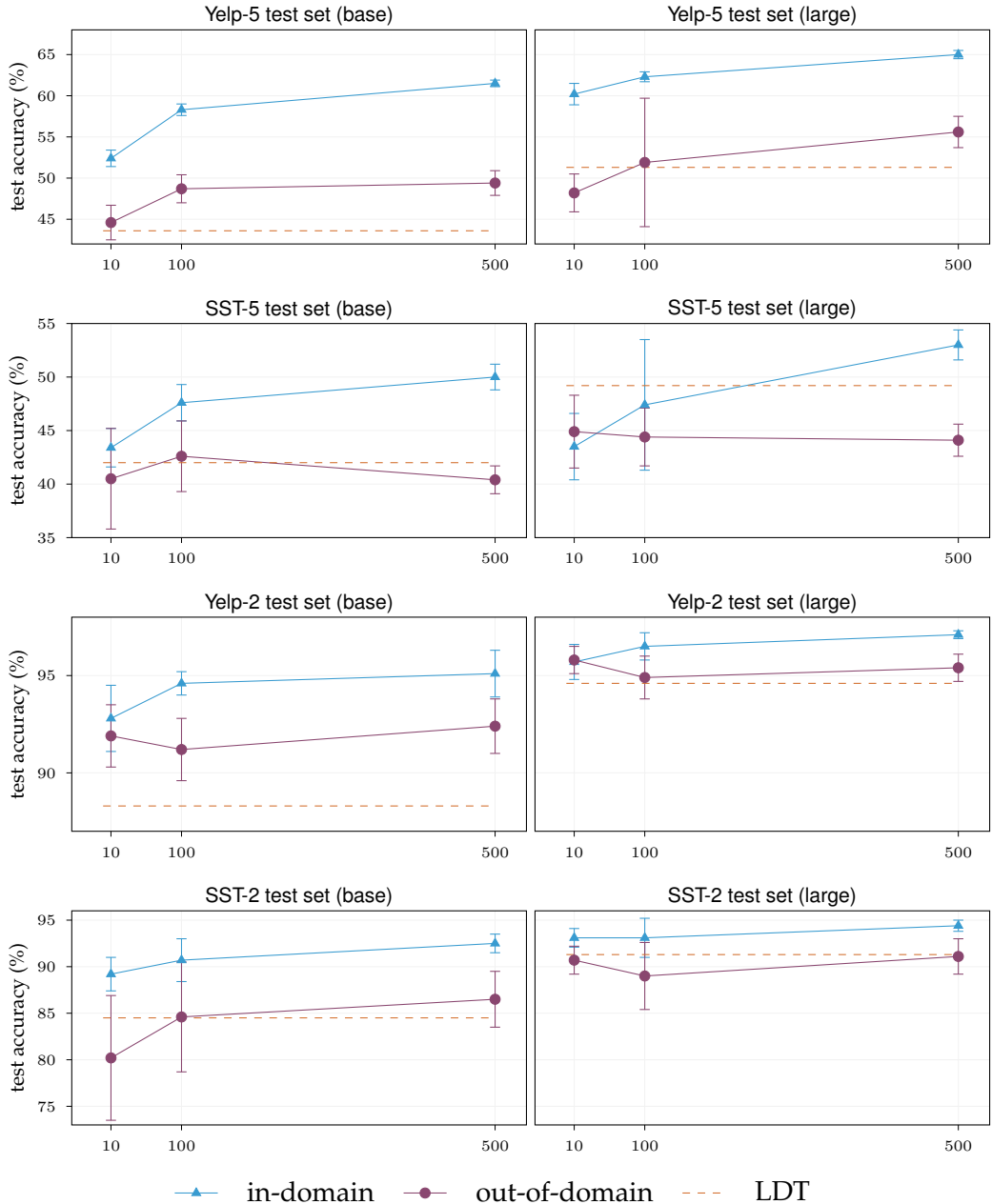

Figure B.2: Domain transfer results of sentiment classification, where X-axis depicts the number of training examples per label. "base/large" in parenthesis denotes RoBERTa-base/large.

## B.2 LABELDESC Data

The statistics of LABELDESC data are shown in Table 4.2. We use the same set of LABELDESC data for AGNews and Yahoo$_{\text{AG}}$, Yelp-5 and SST-5, Yelp-2 and SST-2, respectively. The data is listed in Table B.1 - Table B.4. Each term/sentence that is separated by "|" in tables is an independent LABELDESC example during training. For brevity, we list all hand-crafted templates instead of listing all data for sentiment classification.

| Label | Type | Training Data |
|---|---|---|
| talk.religion.misc | terms | religion | Christian | Buddhist | Jewish |
| | Wiki. | Religion is usually defined as a social-cultural system of designated behaviors and practices, morals, beliefs, worldviews, texts, sanctified places, prophecies, ethics, or organizations, that generally relates humanity to supernatural, transcendental, and spiritual elements; however, there is no scholarly consensus over what precisely constitutes a religion. |
| | dict. | a set of beliefs concerning the cause, nature, and purpose of the universe, especially when considered as the creation of a superhuman agency or agencies, usually involving devotional and ritual observances, and often containing a moral code governing the conduct of human affairs. |
| rec.autos | terms | automobile | truck | car | vehicle |
| | Wiki. | A car (or automobile) is a wheeled motor vehicle that is used for transportation. |
| | dict. | a passenger vehicle designed for operation on ordinary roads and typically having four wheels and a gasoline or diesel internal-combustion engine. |
| sci.med | terms | medicine | hospital | symptom | cure |
| | Wiki. | Medicine is the science and practice of caring for a patient, managing the diagnosis, prognosis, prevention, treatment, palliation of their injury or disease, and promoting their health. |
| | dict. | any substance or substances used in treating disease or illness; medicament; remedy. |
| talk.politics.guns | terms | gun | firearm | weapon | handgun |
| | Wiki. | A gun is a ranged weapon designed to use a shooting tube (gun barrel) to launch projectiles. |
| | dict. | a weapon consisting of a metal tube, with mechanical attachments, from which projectiles are shot by the force of an explosive; a piece of ordnance. |

Table B.1: LABELDESC data for 20NG. "Wiki." and "dict." refers to the data source, i.e., Wikipedia or dictionary definition.

| Label | Type | Training Data |
|---|---|---|
| Negative | terms | awful | terrible | horrendous | horrible | dreadful | bad | unpleasant | unsatisfying | lousy | subpar |
| | sent. | It was $t$. | A(n) $t$ experience. | Just $t$. | Overall, it was $t$. |
| Positive | terms | good | nice | fine | pleasant | neat | great | amazing | excellent | fantastic | outstanding |
| | sent. | It was $t$. | A(n) $t$ experience. | Just $t$. | Overall, it was $t$. |

Table B.2: LABELDESC data for Yelp-2, SST-2, Amz-2 and IMDB.

| Label | Type | Training Data |
|---|---|---|
| Society & Culture | terms | society \| culture |
| | Wiki. | A society is a group of individuals involved in persistent social interaction, or a large social group sharing the same spatial or social territory, typically subject to the same political authority and dominant cultural expectations. \| Culture is an umbrella term which encompasses the social behavior, institutions, and norms found in human societies, as well as the knowledge, beliefs, arts, laws, customs, capabilities, and habits of the individuals in these groups. |
| | dict. | an organized group of persons associated together for religious, benevolent, cultural, scientific, political, patriotic, or other purposes. \| the behaviors and beliefs characteristic of a particular group of people, as a social, ethnic, professional, or age group (usually used in combination) |
| Science & Mathematics | terms | science \| mathematics |
| | Wiki. | Science is a systematic endeavor that builds and organizes knowledge in the form of testable explanations and predictions about the universe. \| Mathematics is an area of knowledge that includes such topics as numbers, formulas and related structures, shapes and the spaces in which they are contained, and quantities and their changes. |
| | dict. | a branch of knowledge or study dealing with a body of facts or truths systematically arranged and showing the operation of general laws \| the systematic treatment of magnitude, relationships between figures and forms, and relations between quantities expressed symbolically. |
| Health | terms | health \| fitness \| medical \| diet |
| | Wiki. | Health, according to the World Health Organization, is "a state of complete physical, mental and social well-being and not merely the absence of disease and infirmity". |
| | dict. | the general condition of the body or mind with reference to soundness and vigor |
| Education & Reference | terms | education \| reference |
| | Wiki. | Education is a purposeful activity directed at achieving certain aims, such as transmitting knowledge or fostering skills and character traits. \| Reference is a relationship between objects in which one object designates, or acts as a means by which to connect to or link to, another object. |
| | dict. | the act or process of imparting or acquiring general knowledge, developing the powers of reasoning and judgment, and generally of preparing oneself or others intellectually for mature life. \| a book or other source of useful facts or information, such as an encyclopedia, dictionary, etc. |
| Computers & Internet | terms | computer \| internet |
| | Wiki. | A computer is a digital electronic machine that can be programmed to carry out sequences of arithmetic or logical operations (computation) automatically. \| The Internet (or internet) is the global system of interconnected computer networks that uses the Internet protocol suite (TCP/IP) to communicate between networks and devices. |
| | dict. | a programmable electronic device designed to accept data, perform prescribed mathematical and logical operations at high speed, and display the results of these operations. Mainframes, desktop and laptop computers, tablets, and smartphones are some of the different types of computers. \| Usually the internet (except when used before a noun). a vast computer network linking smaller computer networks worldwide. The internet includes commercial, educational, governmental, and other networks, all of which use the same set of communications protocols |
| Sports | terms | sport \| sports \| racing \| baseball |
| | Wiki. | Sport pertains to any form of competitive physical activity or game that aims to use, maintain or improve physical ability and skills while providing enjoyment to participants and, in some cases, entertainment to spectators. |
| | dict. | an athletic activity requiring skill or physical prowess and often of a competitive nature, as racing, baseball, tennis, golf, bowling, wrestling, boxing, hunting, fishing, etc. |
| Business & Finance | terms | business \| finance |
| | Wiki. | Business is the activity of making one's living or making money by producing or buying and selling products (such as goods and services). \| Finance is the study and discipline of money, currency and capital assets. |
| | dict. | the purchase and sale of goods in an attempt to make a profit. \| the management of revenues; the conduct or transaction of money matters generally, especially those affecting the public, as in the fields of banking and investment. |
| Entertainment & Music | terms | entertainment \| music |
| | Wiki. | Entertainment is a form of activity that holds the attention and interest of an audience or gives pleasure and delight. \| Music is generally defined as the art of arranging sound to create some combination of form, harmony, melody, rhythm or otherwise expressive content. |
| | dict. | the act of entertaining; agreeable occupation for the mind; diversion; amusement \| an art of sound in time that expresses ideas and emotions in significant forms through the elements of rhythm, melody, harmony, and color. |
| Family & Relationships | terms | family \| relationship |
| | Wiki. | Family is a group of people related either by consanguinity (by recognized birth) or affinity (by marriage or other relationship). \| The concept of interpersonal relationship involves social associations, connections, or affiliations between two or more people. |
| | dict. | a basic social unit consisting of parents and their children, considered as a group, whether dwelling together or not; a social unit consisting of one or more adults together with the children they care for. \| an emotional or other connection between people |
| Politics & Government | terms | politics \| government |
| | Wiki. | Politics is the set of activities that are associated with making decisions in groups, or other forms of power relations among individuals, such as the distribution of resources or status. \| A government is the system or group of people governing an organized community, generally a state. |
| | dict. | the science or art of political government. \| the political direction and control exercised over the actions of the members, citizens, or inhabitants of communities, societies, and states; direction of the affairs of a state, community, etc.; political administration |

Table B.3: LABELDESC data for Yahoo Answers.

| Label | Type | Training Data |
|---|---|---|
| Company | terms | company \| firm \| corporation \| business |
| | Wiki. | A company, abbreviated as co., is a legal entity representing an association of people, whether natural, legal or a mixture of both, with a specific objective. |
| | dict. | a number of persons united or incorporated for joint action, especially for business |
| Educational Institution | terms | educational institution \| university \| college \| school |
| | Wiki. | An educational institution is a place where people of different ages gain an education, including preschools, childcare, primary-elementary schools, secondary-high schools, and universities. |
| | dict. | an institution for instruction in a particular skill or field. |
| Artist | terms | artist \| writer \| actor \| singer |
| | Wiki. | An artist is a person engaged in an activity related to creating art, practicing the arts, or demonstrating an art. |
| | dict. | a person who produces works in any of the arts that are primarily subject to aesthetic criteria. |
| Athlete | terms | athlete \| sports \| footballer \| weightlifter |
| | Wiki. | An athlete (also sportsman or sportswoman) is a person who competes in one or more sports that involve physical strength, speed, or endurance. |
| | dict. | a person trained or gifted in exercises or contests involving physical agility, stamina, or strength; a participant in a sport, exercise, or game requiring physical skill. |
| Office Holder | terms | office-holder \| politics \| mayor \| president |
| | Wiki. | A person who's been appointed to a position by a company or organisation but doesn't have a contract or receive regular payment may be an office-holder. |
| | dict. | a person filling a governmental position; public official. |
| Mean of Transportation | terms | mean of transportation \| car \| bus \| train |
| | Wiki. | Transport (in British English), or transportation (in American English), is the intentional movement of humans, animals, and goods from one location to another. |
| | dict. | a means of transporting or conveying, as a truck or bus. |
| Building | terms | building \| apartment \| skyscraper \| tower |
| | Wiki. | A building or edifice, is an enclosed structure with a roof and walls standing more or less permanently in one place, such as a house or factory (although there's also portable buildings). |
| | dict. | a relatively permanent enclosed construction over a plot of land, having a roof and usually windows and often more than one level, used for any of a wide variety of activities, as living, entertaining, or manufacturing. |
| Natural Place | terms | natural place \| forest \| mountain \| river |
| | Wiki. | The natural environment or natural world encompasses all living and non-living things occurring naturally, meaning in this case not artificial. |
| | dict. | existing in or formed by nature (opposed to artificial) |
| Village | terms | village \| town \| countryside \| rural |
| | Wiki. | A village is a clustered human settlement or community, larger than a hamlet but smaller than a town (although the word is often used to describe both hamlets and smaller towns), with a population typically ranging from a few hundred to a few thousand. |
| | dict. | a small community or group of houses in a rural area, larger than a hamlet and usually smaller than a town, and sometimes (as in parts of the U.S.) incorporated as a municipality. |
| Animal | terms | animal \| insect \| bird \| fish |
| | Wiki. | Animals are multicellular, eukaryotic organisms in the biological kingdom Animalia. |
| | dict. | any member of the kingdom Animalia, comprising multicellular organisms that have a well-defined shape and usually limited growth, can move voluntarily, actively acquire food and digest it internally, and have sensory and nervous systems that allow them to respond rapidly to stimuli: some classification schemes also include protozoa and certain other single-celled eukaryotes that have motility and animallike nutritional modes. |
| Plant | terms | plant \| flower \| tree \| grass |
| | Wiki. | Plants are predominantly photosynthetic eukaryotes, forming the kingdom Plantae. |
| | dict. | Botany. any member of the kingdom Plantae, comprising multicellular organisms that typically produce their own food from inorganic matter by the process of photosynthesis and that have more or less rigid cell walls containing cellulose, including vascular plants, mosses, liverworts, and hornworts: some classification schemes may include fungi, algae, bacteria, and certain single-celled eukaryotes that have plantlike qualities, as rigid cell walls or the use of photosynthesis. |
| Album | terms | album \| soundtrack \| mixtape \| CD |
| | Wiki. | An album is a collection of audio recordings issued on compact disc (CD), vinyl, audio tape, or another medium such as digital distribution. |
| | dict. | a record or set of records containing several musical selections, a complete play or opera, etc. |
| Film | terms | film \| movie \| comedy \| drama |
| | Wiki. | A film – also called a movie, motion picture, moving picture, picture, photoplay or (slang) flick – is a work of visual art that simulates experiences and otherwise communicates ideas, stories, perceptions, feelings, beauty, or atmosphere through the use of moving images. |
| | dict. | a sequence of consecutive still images recorded in a series to be viewed on a screen in such rapid succession as to give the illusion of natural movement; motion picture. |
| Written Work | terms | written work \| novel \| newspaper \| book |
| | Wiki. | A book is a medium for recording information in the form of writing or images, typically composed of many pages (made of papyrus, parchment, vellum, or paper) bound together and protected by a cover. |
| | dict. | a handwritten or printed work of fiction or nonfiction, usually on sheets of paper fastened or bound together within covers. |

Table B.4: LABELDESC data for DBPedia.

# Appendix C

# Appendix to Chapter 5

## C.1 Prompt Construction Template

```
x_defn

Thus given the following input:
input:    x_t
answer:
```

```
x_defn

Some examples are:
input:    x_{1,t}
answer:   y_{1,t}

input:    x_{2,t}
answer:   y_{2,t}

input:    x_{3,t}
answer:   y_{3,t}

input:    x_{4,t}
answer:   y_{4,t}

Thus given the following input:
input:    x_t
answer:
```

Table C.1: Prompt templates for zero-shot and few-shot ICL. $x_t$ refers to the test example, and $x_{i,t}, y_{i,t}$ refers to the text inputs and gold labels of ICL demonstrations selected for $x_t$, respectively.

|  |  | EDOS | | SST | | GoEmotions | |
|---|---|---|---|---|---|---|---|
|  |  | M | L | M | L | M | L |
| Baselines | FREQ | 11.7 | 11.7 | 4.6 | 4.6 | 0.4 | 0.4 |
|  | ZERO | 65 | 60.7 | 54 | 56.2 | 42.6 | 46.3 |
|  | STATIC-$N$ | 65.2$_{\pm0.6}$ | 58.1$_{\pm0.4}$ | 54.5$_{\pm0.6}$ | 58.2$_{\pm0.3}$ | 42.6$_{\pm1.2}$ | 46.2$_{\pm0.3}$ |
|  | RETR-4 | 67.1$_{\pm1.1}$ | 63.6$_{\pm0.5}$ | 53.4$_{\pm0.3}$ | 57.4$_{\pm0.4}$ | 43.7$_{\pm0.4}$ | 47.6$_{\pm0.4}$ |
|  | RETR-8 | 65.0$_{\pm0.2}$ | 63.9$_{\pm0.3}$ | 54.4$_{\pm0.1}$ | 57.6$_{\pm0.5}$ | 43.7$_{\pm0.4}$ | 48.3$_{\pm0.1}$ |
| Ours | AMBIG-4 |  |  |  |  |  |  |
|  | +GOLD | 65.9$_{\pm0.8}$ | 63.6$_{\pm0.4}$ | 54.1$_{\pm0.3}$ | 57.7$_{\pm0.1}$ | 45.7$_{\pm0.3}$ | 50.5$_{\pm0.2}$ |
|  | +GOLD+MIS | 66.6$_{\pm1.1}$ | 63.6$_{\pm1.0}$ | 54.1$_{\pm0.2}$ | 58.8$_{\pm0.1}$ | 44.8$_{\pm0.4}$ | 49.2$_{\pm0.1}$ |
|  | +GOLD+MIS+PRED | 67.4$_{\pm0.4}$ | 65.0$_{\pm0.5}$ | 54.8$_{\pm0.5}$ | 59.4$_{\pm0.0}$ | 46.9$_{\pm1.3}$ | 47.9$_{\pm0.2}$ |
|  | AMBIG-8 |  |  |  |  |  |  |
|  | +GOLD | 66.4$_{\pm1.1}$ | 64.8$_{\pm0.1}$ | 54.7$_{\pm0.2}$ | 58.5$_{\pm0.7}$ | 48.0$_{\pm1.8}$ | 49.9$_{\pm0.1}$ |
|  | +GOLD+MIS | 68.4$_{\pm0.8}$ | 64.4$_{\pm0.6}$ | 54.5$_{\pm0.1}$ | 59.6$_{\pm0.1}$ | 48.7$_{\pm0.5}$ | 48.8$_{\pm0.5}$ |
|  | +GOLD+MIS+PRED | 66.6$_{\pm1.2}$ | 66.4$_{\pm0.3}$ | 54.9$_{\pm0.2}$ | 59.1$_{\pm0.4}$ | 43.7$_{\pm0.5}$ | 47.4$_{\pm0.3}$ |

Table C.2: Precision (%) comparison between our proposed methods and baselines with Flan-PaLM 2 (M, L).

|  |  | EDOS | | SST | | GoEmotions | |
|---|---|---|---|---|---|---|---|
|  |  | M | L | M | L | M | L |
| Baselines | FREQ | 25 | 25 | 20 | 20 | 3.7 | 3.7 |
|  | ZERO | 46 | 62.8 | 53.8 | 55.2 | 42.4 | 47.2 |
|  | STATIC-$N$ | 46.2$_{\pm0.3}$ | 63.0$_{\pm0.3}$ | 54.0$_{\pm0.4}$ | 56.5$_{\pm0.2}$ | 34.8$_{\pm0.5}$ | 49.5$_{\pm0.4}$ |
|  | RETR-4 | 44.8$_{\pm0.3}$ | 63.4$_{\pm0.2}$ | 53.4$_{\pm0.3}$ | 55.7$_{\pm0.3}$ | 38.5$_{\pm0.2}$ | 49.7$_{\pm0.3}$ |
|  | RETR-8 | 44.0$_{\pm0.1}$ | 62.1$_{\pm0.2}$ | 54.2$_{\pm0.1}$ | 55.3$_{\pm0.4}$ | 37.8$_{\pm0.3}$ | 50.1$_{\pm0.3}$ |
| Ours | AMBIG-4 |  |  |  |  |  |  |
|  | +GOLD | 45.1$_{\pm0.6}$ | 64.1$_{\pm0.2}$ | 54.6$_{\pm0.4}$ | 56.4$_{\pm0.1}$ | 41.4$_{\pm0.3}$ | 51.3$_{\pm0.2}$ |
|  | +GOLD+MIS | 48.0$_{\pm0.4}$ | 62.1$_{\pm0.9}$ | 54.9$_{\pm0.1}$ | 57.3$_{\pm0.1}$ | 40.9$_{\pm0.1}$ | 51.0$_{\pm0.4}$ |
|  | +GOLD+MIS+PRED | 49.5$_{\pm0.4}$ | 63.1$_{\pm0.3}$ | 55.6$_{\pm0.4}$ | 57.7$_{\pm0.0}$ | 42.7$_{\pm0.2}$ | 51.7$_{\pm0.4}$ |
|  | AMBIG-8 |  |  |  |  |  |  |
|  | +GOLD | 43.6$_{\pm0.1}$ | 64.0$_{\pm0.2}$ | 55.0$_{\pm0.1}$ | 56.5$_{\pm0.6}$ | 41.9$_{\pm0.9}$ | 50.8$_{\pm0.5}$ |
|  | +GOLD+MIS | 47.3$_{\pm0.4}$ | 61.8$_{\pm0.3}$ | 54.9$_{\pm0.1}$ | 57.3$_{\pm0.1}$ | 44.4$_{\pm0.2}$ | 51.4$_{\pm0.3}$ |
|  | +GOLD+MIS+PRED | 48.0$_{\pm0.4}$ | 61.7$_{\pm0.2}$ | 55.6$_{\pm0.2}$ | 56.7$_{\pm0.2}$ | 43.2$_{\pm0.3}$ | 51.3$_{\pm0.1}$ |

Table C.3: Recall (%) comparison between our proposed methods and baselines with Flan-PaLM 2 (M, L).

## C.2 Accuracy, Precision, Recall

Please refer to Table C.2, C.3 and C.4.

|  |  | EDOS | | SST | | GoEmotions | |
|---|---|---|---|---|---|---|---|
|  |  | M | L | M | L | M | L |
| Baselines | FREQ | 46.8 | 46.8 | 23.1 | 23.1 | 11.7 | 11.7 |
|  | ZERO | 55.4 | 59.2 | 49.9 | 57.1 | 47.1 | 46.2 |
|  | STATIC-$N$ | 54.3$_{\pm0.3}$ | 57.6$_{\pm0.1}$ | 50.5$_{\pm0.4}$ | 59.0$_{\pm0.2}$ | 39.8$_{\pm0.2}$ | 46.7$_{\pm0.2}$ |
|  | RETR-4 | 53.6$_{\pm0.3}$ | 61.0$_{\pm0.5}$ | 50.0$_{\pm0.3}$ | 58.5$_{\pm0.3}$ | 45.9$_{\pm0.0}$ | 50.1$_{\pm0.2}$ |
|  | RETR-8 | 53.8$_{\pm0.3}$ | 61.1$_{\pm0.2}$ | 51.8$_{\pm0.1}$ | 58.6$_{\pm0.4}$ | 45.3$_{\pm0.0}$ | 51.0$_{\pm0.2}$ |
| Ours | AMBIG-4 | | | | | | |
|  | +GOLD | 54.3$_{\pm0.4}$ | 61.3$_{\pm0.4}$ | 51.5$_{\pm0.4}$ | 58.9$_{\pm0.1}$ | 46.6$_{\pm0.2}$ | 50.3$_{\pm0.1}$ |
|  | +GOLD+MIS | 56.1$_{\pm0.2}$ | 60.9$_{\pm0.6}$ | 52.1$_{\pm0.1}$ | 59.7$_{\pm0.1}$ | 45.6$_{\pm0.1}$ | 49.5$_{\pm0.1}$ |
|  | +GOLD+MIS+PRED | 56.5$_{\pm0.1}$ | 61.4$_{\pm0.4}$ | 53.0$_{\pm0.4}$ | 60.1$_{\pm0.1}$ | 45.6$_{\pm0.1}$ | 50.0$_{\pm0.2}$ |
|  | AMBIG-8 | | | | | | |
|  | +GOLD | 53.6$_{\pm0.2}$ | 61.8$_{\pm0.0}$ | 52.9$_{\pm0.1}$ | 59.5$_{\pm0.6}$ | 46.8$_{\pm0.1}$ | 50.4$_{\pm0.2}$ |
|  | +GOLD+MIS | 55.4$_{\pm0.6}$ | 61.1$_{\pm0.3}$ | 53.2$_{\pm0.1}$ | 60.2$_{\pm0.1}$ | 45.8$_{\pm0.2}$ | 50.0$_{\pm0.3}$ |
|  | +GOLD+MIS+PRED | 55.1$_{\pm0.6}$ | 61.5$_{\pm0.3}$ | 54.2$_{\pm0.2}$ | 59.6$_{\pm0.3}$ | 44.9$_{\pm0.2}$ | 50.2$_{\pm0.2}$ |

Table C.4: Accuracy (%) comparison between our proposed methods and baselines with Flan-PaLM 2 (M, L).

|  | EDOS | SST | GoEmotions |
|---|---|---|---|
| Flan-PaLM 2 (M) | 91.2 | 85.8 | 61.2 |
| Flan-PaLM 2 (L) | 88.2 | 87.6 | 61.6 |

Table C.5: Percentage of times the test example's gold label is in $\mathcal{L}_{ambig,t}$ (as obtained from ZERO model).

## C.3 Label-wise Percentage Analysis of Gold Label Inclusion in $\mathcal{L}_{\text{ambig,t}}$

We compute the percentage of times that the test example's gold label is in $\mathcal{L}_{ambig,t}$ (as obtained with ZERO) in Table C.5, and we present label-wise results in Table C.6.

| | | |
|---|---|---|
| EDOS | M | Animosity 99.1, Derogation 97.4, Prejudiced 52.1, Threat 71.9 |
| | L | Animosity 90.1, Derogation 90.1, Prejudiced 68.1, Threat 93.3 |
| SST | M | Bad 78.4, Good 98.0, Great 88.0, Okay 73.8, Terrible 93.9 |
| | L | Bad 89.3, Good 99.2, Great 99.2, Okay 59.1, Terrible 85.3 |
| GoEmotions | M | Admiration 72.7, Amusement 90.3, Anger 58.8, Annoyance 44.3, Approval 24.6, Caring 52.9, Confusion 73.2, Curiosity 64.2, Desire 35.7, Disappointment 58.0, Disapproval 52.3, Disgust 55.3, Embarrassment 30.4, Excitement 56.4, Fear 75.4, Gratitude 75.7, Grief 100.0, Joy 80.4, Love 86.8, Nervousness 83.3, Optimism 51.4, Pride 42.9, Realization 27.0, Relief 28.6, Remorse 38.6, Sadness 71.6, Surprise 62.1 |
| | L | Admiration 40.2, Amusement 84.9, Anger 52.7, Annoyance 40.2, Approval 36.0, Caring 36.5, Confusion 74.2, Curiosity 64.8, Desire 58.9, Disappointment 59.1, Disapproval 81.0, Disgust 38.2, Embarrassment 30.4, Excitement 61.8, Fear 73.8, Gratitude 88.4, Grief 100.0, Joy 84.8, Love 86.2, Nervousness 83.3, Optimism 65.4, Pride 71.4, Realization 41.6, Relief 42.9, Remorse 70.5, Sadness 67.6, Surprise 64.4 |

Table C.6: Label-wise percentage of times the test example's gold label is in $\mathcal{L}_{ambig,t}$ (as obtained from ZERO), where M and L refers to Flan-PaLM 2 sizes.

## C.4 Sorting Order with Predicted Probability Distribution Entropy

Since we have the predicted probability distribution of ICL demonstrations, we tried to sort the ICL demonstrations by increasing entropy order. However, it doesn't consistently improve model performance, which is shown in Table C.7 and C.8.

## C.5 Example on SST for Comparison between RETR and AMBIG-ICL

We list example demonstrations on SST where the model correctly predict the test example in our proposed method, but wrongly classify it in RETR in Table C.9.

|  | EDOS | | SST | |
|---|---|---|---|---|
|  | M | L | M | L |
| AMBIG-4 | | | | |
| +GOLD | 50.2 | 62.9 | 51.6 | 55.8 |
| +GOLD+MIS | 51.2 | 62.7 | 53.0 | 57.0 |
| +GOLD+MIS+PRED | 53.4 | 63.8 | 52.7 | 57.7 |
| AMBIG-8 | | | | |
| +GOLD | 48.1 | 63.3 | 53.2 | 56.5 |
| +GOLD+MIS | 50.4 | 62.9 | 53.6 | 57.4 |
| +GOLD+MIS+PRED | 50.3 | 62.9 | 54.3 | 57.1 |

Table C.7: F1 macro scores (%) of our method. M and L refers to size of Flan-PaLM 2. The ICL demonstrations are sorted by increased entropy order.

|  | EDOS | | SST | |
|---|---|---|---|---|
|  | M | L | M | L |
| AMBIG-4 | | | | |
| +GOLD | 0.9 | 0.3 | 0.1 | -0.3 |
| +GOLD+MIS | -1 | 1 | 0.7 | -0.4 |
| +GOLD+MIS+PRED | -0.5 | 0.9 | -0.6 | -0.3 |
| AMBIG-8 | | | | |
| +GOLD | 0.6 | 0.1 | 0.3 | 0 |
| +GOLD+MIS | 0 | 0.9 | 0.2 | -0.3 |
| +GOLD+MIS+PRED | -0.6 | 0.2 | 0 | -0.1 |

Table C.8: The difference of F1 macro scores (%) between the "increased entropy order" and the "averaged over 3 random seeds".

| Test Example: A hip ride into hyper-time, Clockstoppers is a lively and enjoyable adventure for all ages at any time. $\mathcal{L}_{\text{ambig},t}$: Great, Good | | Gold label: Great |
|---|---|---|
| RETR | 1. Bad: See *Clockstoppers* if you have nothing better to do with 94 minutes.<br>2. Bad: Time stands still in more ways that one in *Clockstoppers*, a sci-fi thriller as lazy as it is interminable.<br>3. Bad: *Clockstoppers* is one of those crazy, mixed-up films that doesn't know what it wants to be when it grows up.<br>4. Good: Even with all its botches, Enigma offers all the pleasure of a handsome and well-made entertainment. | Predicted: Good |
| AMBIG-ICL | | |
| +GOLD | 1. Good: Even with all its botches, Enigma offers all the pleasure of a handsome and well-made entertainment.<br>2. Great: A breathtaking adventure *for all ages*, Spirit tells its poignant and uplifting story in a stunning fusion of music and images.<br>3. Great: A rollicking ride, with jaw-dropping action sequences, striking villains, a gorgeous color palette, astounding technology, stirring music and a boffo last hour that leads up to a strangely sinister happy ending.<br>4. Great: This gorgeous epic is guaranteed to lift the spirits of the whole family. | Predicted: Great |
| +GOLD+MIS | 1. Good: As action-adventure, this space-based homage to Robert Louis Stevenson's Treasure Island fires on all plasma conduits.<br>2. Good: Horns and Halos benefits from serendipity but also reminds us of our own responsibility to question what is told as the truth.<br>3. Great: Return to Never Land is reliable, standard Disney animated fare, with enough creative energy and wit to entertain *all ages*.<br>4. Great: It's a smart, solid, kinetically-charged spy flick worthy of a couple hours of summertime and a bucket of popcorn. | Predicted: Great |
| +GOLD+MIS+PRED | 1. Good: As action-adventure, this space-based homage to Robert Louis Stevenson's Treasure Island fires on all plasma conduits.<br>2. Good: Horns and Halos benefits from serendipity but also reminds us of our own responsibility to question what is told as the truth.<br>3. Great: Return to Never Land is reliable, standard Disney animated fare, with enough creative energy and wit to entertain all ages.<br>4. Great: It's a smart, solid, kinetically-charged spy flick worthy of a couple hours of summertime and a bucket of popcorn. | Predicted: Great |

Table C.9: Example demonstrations selected by the RETR and our proposed method AMBIG-ICL for the SST task, for $n = 4$. The model used here for prediction is Flan-PaLM 2 (L).